\documentclass[11pt]{article}

\usepackage[final]{acl}

\usepackage{times}
\usepackage{latexsym}

\usepackage[T1]{fontenc}

\usepackage[utf8]{inputenc}

\usepackage{microtype}

\usepackage{inconsolata}

\usepackage{graphicx}

\usepackage{xspace}
\usepackage{siunitx}
\usepackage{hyperref}
\usepackage{graphicx}
\usepackage{url}
\usepackage{booktabs} 
\usepackage{multirow}
\usepackage{arydshln} 
\usepackage{graphicx}
\usepackage{wrapfig}
\usepackage{subcaption}
\usepackage{paralist}
\usepackage{enumitem}
\usepackage[table]{xcolor}   
\definecolor{Highlight}{HTML}{FFF176}  
\usepackage{amsmath}
\usepackage{amssymb}
\usepackage{algorithm}
\usepackage{algorithmic}
\usepackage{tcolorbox}
\usepackage{mathtools}
\usepackage[capitalise,noabbrev]{cleveref}
\usepackage{placeins}

\def\method{Probabilistic Confidence Selection And Ranking\xspace}
\def\methodtwo{\textbf{P}robab\textbf{i}listic \textbf{C}onfidence \textbf{S}election \textbf{A}nd \textbf{R}anking\xspace}
\def\methodabbrev{PiCSAR\xspace}

\def\eg{{\em e.g.,}\xspace}
\def\ie{{\em i.e.,}\xspace}

\definecolor{mygreencolor}{HTML}{3FAE63}
\definecolor{myredcolor}{HTML}{D85C5C}

\title{PiCSAR: Probabilistic Confidence Selection \\ and Ranking for Reasoning Chains}

\author{%
Joshua Ong Jun Leang$^{1,2}$ \quad Zheng Zhao$^{2}$ \quad Aryo Pradipta Gema$^{2}$ \quad Sohee Yang$^{3}$ \\
\textbf{Wai-Chung Kwan}$^{2}$ \quad \textbf{Xuanli He}$^{3}$ \quad \textbf{Wenda Li}$^{2}$ \quad \textbf{Pasquale Minervini}$^{2,4}$ \\
\textbf{Eleonora Giunchiglia}$^{1}$ \quad \textbf{Shay B. Cohen}$^{2}$ \\
$^{1}$Imperial College London \quad
$^{2}$University of Edinburgh \quad
$^{3}$UCL \quad
$^{4}$Miniml.AI
\\
\texttt{\{j.ong25,e.giunchiglia\}@imperial.ac.uk} \quad \texttt{scohen@inf.ed.ac.uk}
}

\begin{document}
\maketitle
\begin{abstract}
%
Best-of-$n$ sampling improves the accuracy of large language models (LLMs) and large reasoning models (LRMs) by generating multiple candidate solutions and selecting the one with the highest reward.
%
A key challenge for reasoning tasks is designing a scoring function that can identify correct reasoning chains without access to ground-truth answers.
We propose \methodtwo (\methodabbrev): a simple, training-free method that scores each candidate generation using the joint log-likelihood of the reasoning and final answer.
%
This method uses both the scores of the reasoning path (\emph{reasoning confidence}) and the final answer (\emph{answer confidence}).
%
%
\methodabbrev achieves substantial gains across several benchmarks
($+11.7$ on AIME2024, $+9.81$ on AIME2025), outperforming baselines with at least 2x fewer samples in 20 out of 25 comparisons.
%
Our analysis reveals that correct reasoning chains exhibit higher reasoning and answer confidence levels, justifying the effectiveness of \methodabbrev.

\let\thefootnote\relax\footnotetext{\emph{In Findings of the Association for Computational Linguistics: ACL 2026.}}


%
%
%
%
\end{abstract}

\section{Introduction}

Recent studies have shown that LLMs achieve strong performance on complex reasoning tasks~\citep{grattafiori2024llama, team2024gemma, hurst2024gpt}.
Techniques such as Chain-of-Thought~\citep[CoT;][]{wei2022chain, kojima2022large} aim
to enhance the reasoning process by generating explicit intermediate reasoning steps.
%
%
Building on these advances, large reasoning models (LRMs), LLMs that receive intensive reasoning‑focused post‑training, such as DeepSeek‑R1~\citep{guo2025deepseek} and Qwen3~\citep{yang2025qwen3}, solve complex problems by generating long CoT reasoning traces.
These traces are often extended via test‑time scaling~\citep{muennighoff2025s1} and can include reflective self‑checking~\citep{yang2025well}.

\begin{figure}[t]
  \centering
  \includegraphics[width=0.48\textwidth]
{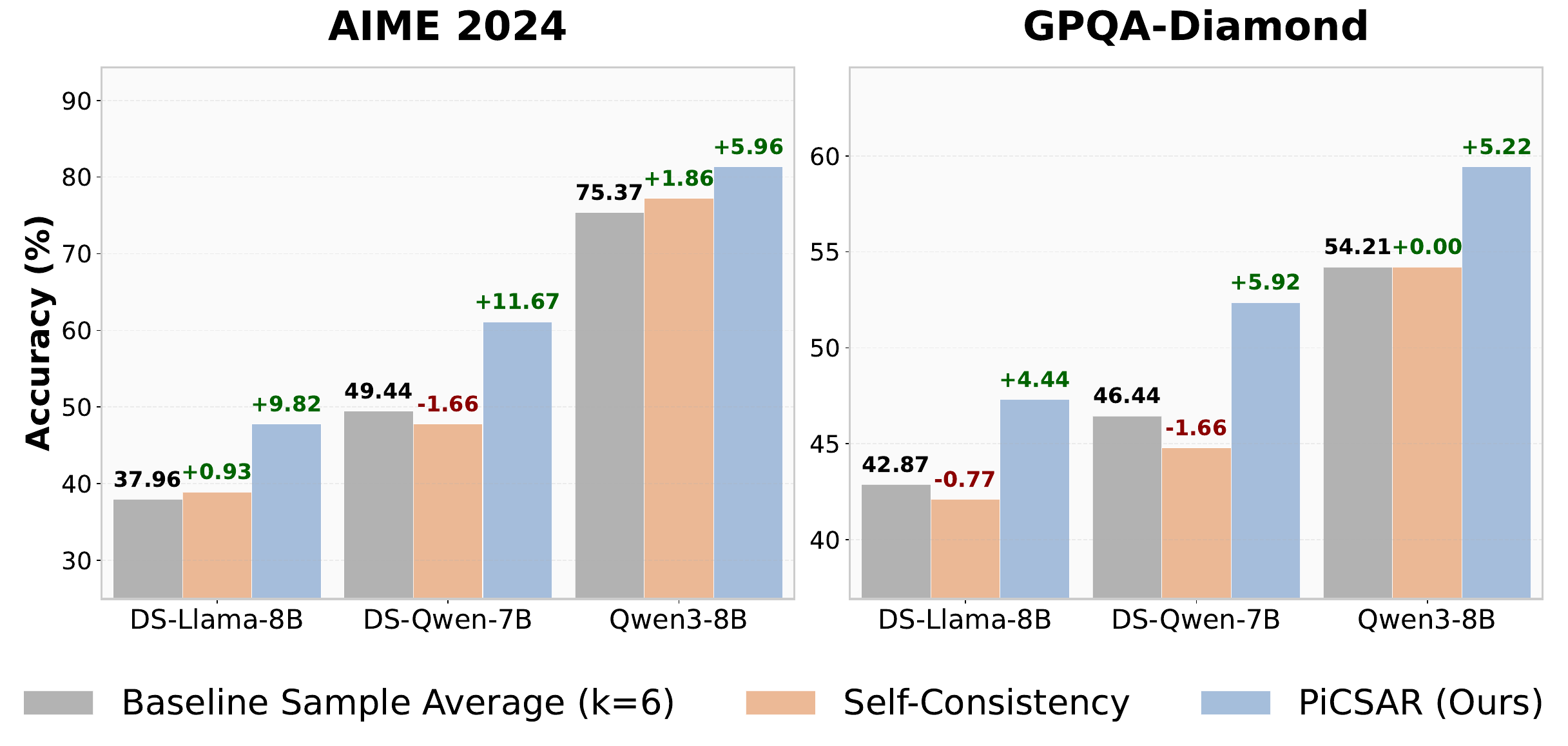}
  \caption{Performance of PiCSAR and Self-Consistency on two reasoning datasets and three models.}
  \label{fig:picsar_acc_fig1}
  \vspace{-3ex}
\end{figure}

Despite these advances, classic decoding approaches such as greedy decoding often fall short of state-of-the-art performance on complex benchmarks~\citep{team2025kimi, balunovic2025matharena}, emphasising the need for more sophisticated inference-time strategies.
\begin{figure*}[t]
  \centering
  \includegraphics[width=0.85\textwidth] {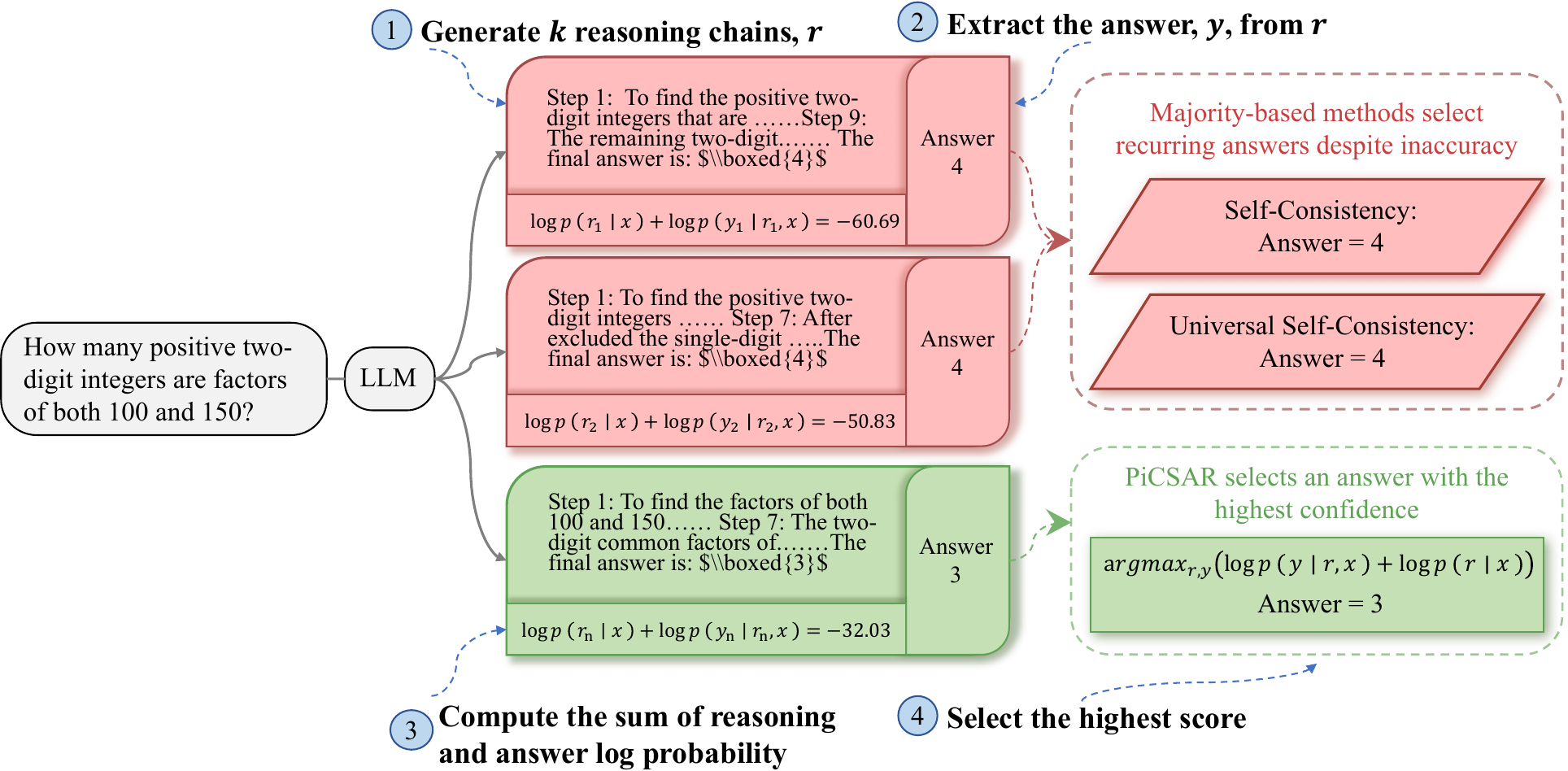}
  \caption{Example with \emph{Llama-3.1-8B} on \emph{MATH500}, where \methodabbrev selects the most likely reasoning trace $r$ and answer $y$ by jointly maximising their log-likelihoods $\log p(r \mid x)$ and $\log p(y \mid r,x)$.
  }
  \label{fig:confidence-iclr-main}
  \vspace{-3ex}
\end{figure*}
\emph{Best-of-N} (BoN) sampling~\citep{stiennon2020learning} emerged as an important technique, where $n$ candidate responses are generated, and the highest-scoring one is selected via a reward model~\citep{mudgal2023controlled, huang2025best}.
However, training external reward models can be computationally expensive~\citep{wang2023math} and vulnerable to distribution shifts~\citep{eisenstein2023helping}. 
%

%
This led to the adoption of simpler, training-free BoN variants, such as Self-Consistency~\citep{wang2022self}, which selects the most frequent answer among multiple generated outputs.
%
However, a key limitation of Self-Consistency is its exclusive reliance on the final answer while ignoring the reasoning that leads to it.
Extensions such as Universal Self-Consistency~\citep[USC;][]{chen2023universal} prompt the model to identify the most consistent response from a set of candidates.
%
However, USC focuses on majority agreement over full responses, overlooking reasoning-level signals critical to answer quality, such as coherence and plausibility. 
%
USC is further constrained by context-window size and reasoning ability~\citep{chen2023universal}, proving particularly ineffective with smaller models~\citep{kang2025scalable}. 
Attempts to overcome this by prompting the model to self-evaluate are often ineffective, as explicit confidence is often poorly calibrated~\citep{miao2023selfcheck,taubenfeld2025confidence}.


%
%
To address these challenges, we introduce~\method(\methodabbrev), a probabilistic confidence method for selecting a reasoning chain $r$ together with its corresponding answer $y$ without requiring any additional training or fine-tuning.
Our approach is straightforward to implement and can be used with any LLM or LRM as an inference-time tool.
It is based on a new scoring function that, given a prompt $x$, selects a reasoning chain $r$ and the answer $y$ by maximising their joint conditional likelihood $\log p(y, r \mid x)$.
This objective naturally separates into two complementary components.
The \emph{reasoning confidence} term $\log p(r \mid x)$ promotes high-probability reasoning sequences by implicitly evaluating the likelihood of the chain given the prompt.
The \emph{answer confidence} term $\log p(y \mid r, x)$ quantifies the model’s certainty in its final prediction, conditioned on the generated reasoning chain.
%
%
\cref{fig:confidence-iclr-main} shows a high-level outline of \methodabbrev, and how it can solve instances that Self-Consistency and USC cannot solve correctly.
%

%
We evaluate \methodabbrev on reasoning tasks across five LLMs and three LRMs, outperforming Self-Consistency and USC in most cases.
\methodabbrev achieves these gains with far fewer samples, often requiring only $k=6$ samples to beat baselines using $k=16 \text{ or } 32$ samples.
PiCSAR substantially improves LRM performance, with Deepseek-R1-distilled-Llama-3 gaining \emph{+13.33\% and +7.58\%} over Self-Consistency on AIME2024 and GPQA-Diamond, respectively (\cref{fig:picsar_acc_fig1}). 
%
%
Unlike USC, which is bounded by the model's reasoning abilities, \methodabbrev decouples confidence estimation, allowing smaller models to effectively capture stable reasoning process properties rather than model artefacts (\S\ref{sec:decoupled_confidence}).
%
%

%
Beyond empirical results, we provide a comprehensive analysis of LLM confidence behaviour.
At finer granularity, we analyse answer confidence at the sentence level using \emph{information density}, defined as the ratio of peak-confidence instances to sentence count (peak-to-sentence ratio), which measures how frequently a reasoning chain attains high confidence relative to its length. 
We find that higher accuracy correlates with high information density 
within model families~(\S\ref{sec:section3}).
%
In addition, we show that answer confidence positively correlates with downstream accuracy 
~(\S\ref{intra-inter-confidence}). 



\section{A Joint Probabilistic Method for Reasoning Chain Selection}
We propose a training-free method for selecting a reasoning chain from a set of candidates, grounded in a probabilistic framework that leverages the model's confidence as its scoring signal. We frame the selection problem as an approximation of maximum a posteriori (MAP) decoding over the joint space of reasoning chains and final answers.

\subsection{Scoring Function and Log-likelihood Decomposition}
We denote by $\mathcal{X}$ a set of possible prompts, $\mathcal{R}$ a set of reasoning chains, and $\mathcal{Y}$ the set of possible final answers.
For a given input prompt $x \in \mathcal{X}$, our goal is to find the high-confidence reasoning chain $r \in \mathcal{R}$ and its corresponding answer $y \in \mathcal{Y}$.
%
%
%
Consider a selection criterion that aims to identify the pair $(r, y)$ with the highest joint conditional probability, $p(r, y \mid x)$. By the chain rule of probability, this decomposes into two distinct components:
%
\vspace{-1ex}
\begin{equation}
\label{eq:joint_prob_decomposition}
p(r, y \mid x) = p(y \mid r, x) \cdot p(r \mid x).
\vspace{-1ex}
\end{equation}
%
In log-space, the joint probability becomes the sum of two log-likelihood terms as follows:
%
\vspace{-1ex}
\begin{equation}
\label{eq:main_score}
\text{Score}(r, y) = \underbrace{\log p(r \mid x)}_{\mathclap{\text{Reasoning Confidence}}} \ + \ \underbrace{\log p(y \mid r, x)}_{\mathclap{\text{Answer Confidence}}}.
\vspace{-1ex}
\end{equation}
These two terms provide complementary signals regarding the quality of a candidate generation:
\begin{itemize}[topsep=0pt,noitemsep,parsep=0pt,leftmargin=*]
    \item \textbf{Reasoning Confidence ($\log p(r \mid x)$):} This term quantifies the model's  confidence in generating $r$ given the prompt $x$. It quantifies the plausibility of the reasoning path itself.
    \item \textbf{Answer Confidence ($\log p(y \mid r, x)$):} measures the model's certainty in the answer $y$, \emph{conditioned on the reasoning chain it has produced}.
\end{itemize}

\subsection{\method (\methodabbrev) }

Directly selecting $r \in \mathcal{R}$, $y \in Y$, where the joint log likelihood \text{Score}$(r, y)$ 
is maximised over the space of possible pairs, is intractable.
We therefore approximate this optimisation with our \methodabbrev sampling-based approach, as outlined in~Algorithm~\ref{alg:prob_confidence}. We first generate $k$ candidate reasoning chains $\{ r_1, r_2, \ldots, r_{k}\}$ from the model's posterior $p(r \mid x)$. Each chain $r_i$ implies a corresponding final answer $y_i$. We then re-rank these candidates using the \methodabbrev scoring function. 

The \emph{reasoning confidence} term is obtained by summing the token-level log-probabilities from the model during the generation of $r_i$. By not applying length normalisation, this term naturally favours more concise and direct reasoning paths as it involves a cumulative sum of individual token log-probabilities. We also consider the length-normalised variant, \methodabbrev-N, which focuses more on the impact of log probability per token rather than favouring concise reasoning paths, leading to similar results (details in Appendix~\ref{app:picsar-n}).


The \emph{answer confidence} term, $\log p(y \mid r, x)$, however, presents a practical challenge. As the model's distribution is over all possible text continuations, the probability of a final answer is confounded by the likelihood of whatever text might follow it. This makes the raw log-probabilities of different answers fundamentally incomparable.
%
%
To address this and ensure we can reliably extract a final answer for answer confidence computation, we condition the model on an explicit instruction prompt, denoted as $\langle a \rangle$, which is appended after the reasoning chain. This prompt explicitly asks the model to provide the final answer based on the preceding context (\ie ``\textit{When you see a potential reasoning followed by} $\langle\text{sep}\rangle$\textit{, output the final answer.}''), with details of the prompt provided in~\cref{appendix:implementation}.
While we extract the answer $y$ directly from the reasoning chain $r$, we use this augmented prompt to compute the answer confidence. 

\begin{algorithm}[t!]
\caption{\method(\methodabbrev)}
\label{alg:prob_confidence}
\begin{algorithmic}[1]
    \STATE \textbf{Input:} Prompt $x$, number of samples $k$, instruction prompt $\langle a \rangle$.
    \STATE \textbf{Output:} Reasoning chain $r^*$ and answer $y^*$.
    \STATE \textbf{Generate Candidates:} Independently sample $k$ reasoning chains $\{r_1, r_2, \ldots, r_k\}$ from the model, where each ${r_i \sim p(r | x), i = 1, \ldots, k}$.
    \STATE \textbf{Score Candidates:}
    \FOR{each $i \in \{1, \ldots, k\}$}
        \STATE \textbf{Extract Reasoning Confidence:} Retrieve $C_{\text{reason}}(i) = \log p(r_i \mid x)$ from $r_i$.
        \STATE \textbf{Extract Answer:} Extract answer, $y_{i}$, from reasoning chain, $r_i$.
        \STATE \textbf{Compute Answer Confidence:} \\ $C_{\text{answer}}(i) = \log p(y_i \mid \langle a \rangle, r_i, x)$.
        \STATE \textbf{Compute Final Score:} \\
        $\text{Score}(r_i, y_i) = C_{\text{reason}}(i) + C_{\text{answer}}(i)$.
    \ENDFOR
    \STATE \textbf{Select Best:} Identify the highest-scoring candidate: $i^* = \arg\max_{i} \text{Score}(r_i, y_i)$.
    \STATE \textbf{Return:} $(r_{i^*}, y_{i^*})$.
\end{algorithmic}
\end{algorithm}

Our modified objective is thus:
\vspace{-1ex}
\begin{equation}
\label{eq:modified_objective}
\arg\max_{r,y} \left[ \log p(r \mid x) + \log p(y \mid \langle a \rangle, r, x) \right].
\vspace{-1ex}
\end{equation}

\textbf{Methodological Departure from Standard MAP Decoding.} 
While the decomposition in \cref{eq:modified_objective} relies on the foundational chain rule, \methodabbrev fundamentally differs from standard Maximum A Posteriori (MAP) decoding or beam search. In standard continuous decoding, the joint probability of a CoT sequence is disproportionately dominated by the arbitrary length and local perplexity of the reasoning steps, effectively drowning out the signal of the final deductive answer. This limitation has historically driven the field away from likelihood-based scoring for reasoning tasks, favouring majority voting or externally trained reward models. By introducing the instructional intervention $\langle a \rangle$, \methodabbrev breaks this continuous autoregressive evaluation. It explicitly forces the model to evaluate the logical entailment of the answer independently of the generative probability of the preceding text. This isolates the conditional \emph{answer confidence}, turning \cref{eq:main_score} into \cref{eq:modified_objective} and thereby yielding a robust, training-free ranking mechanism.

%

The final step is to select the candidate pair with the highest score. As illustrated in Figure~\ref{fig:confidence-iclr-main}, the two components of our scoring function play complementary roles. 
The \emph{reasoning confidence} is the sum of log-probabilities for every token in the reasoning chain. 
Since these log-probabilities are negative, longer sequences tend to accumulate more negative values (i.e., larger magnitude), and can therefore dominate the overall score
(see Appendix~\ref{sec:example_picsar_scoring}).
The \emph{answer confidence} in turn serves as a discriminator, often proving decisive when multiple candidate chains exhibit similar reasoning plausibility. 




\subsection{Confidence Information Plane}
\label{sec:empirical_justification}

\begin{figure}[t]
  \centering
  \includegraphics[width=0.48\textwidth]
  {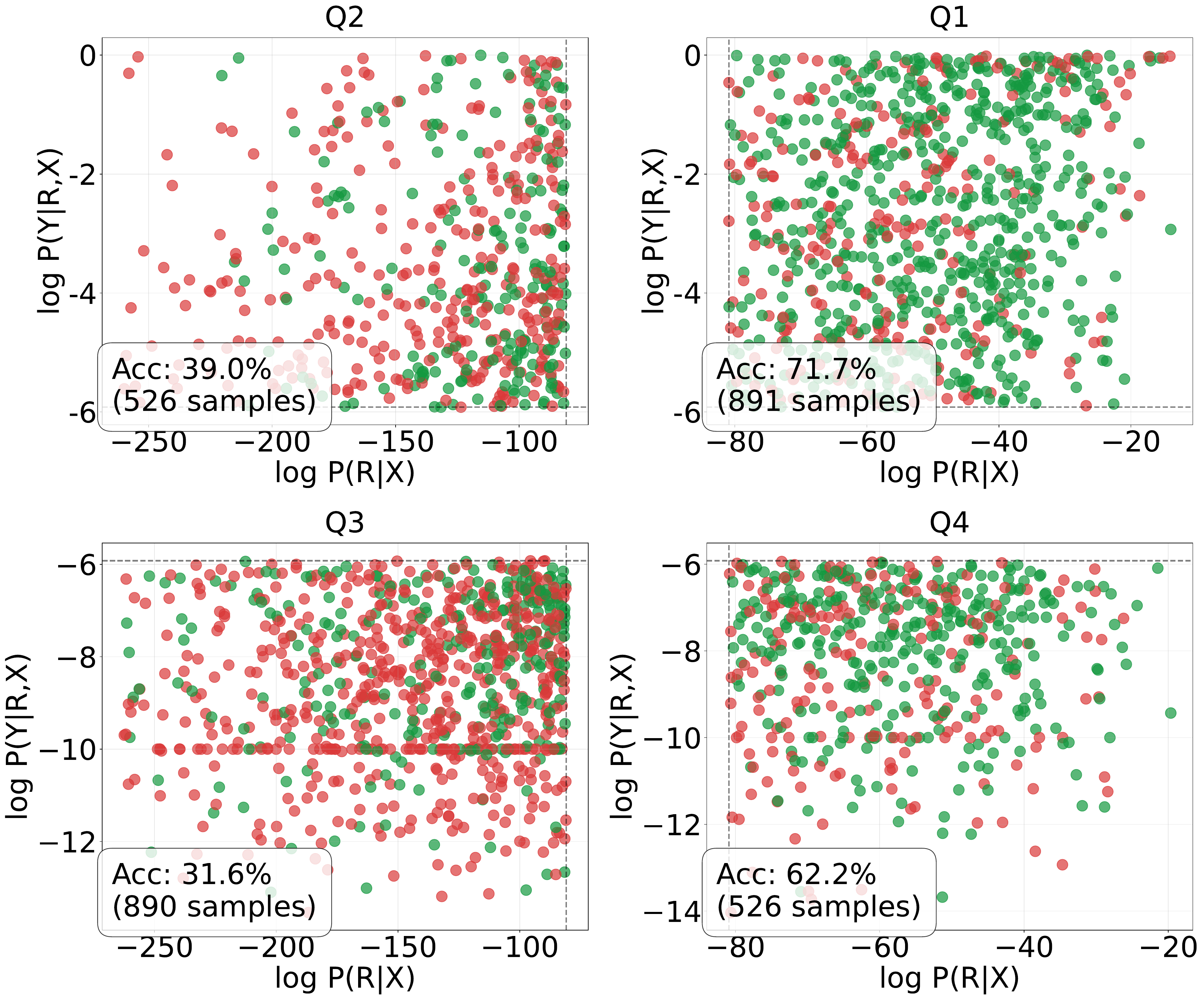}
  \caption{Information plane of MATH500 questions that Llama-3.1-8B predicts \textcolor{mygreencolor}{correctly} and \textcolor{myredcolor}{incorrectly} ($k=6$). Quadrants show combinations of answer and reasoning confidence. This pattern is consistent across LLMs, LRMs, and datasets (Appendix~\ref{sec:information_plane_selection}).}
  \label{fig:info_plane_llama_math500}
  \vspace{-2ex}
\end{figure}

To motivate \methodabbrev design, we analyse the distribution of model-generated samples on a 2D ``Information Plane'', with respect to our two confidence terms (\cref{fig:info_plane_llama_math500}).
We partition the plane into four quadrants using the median value of each axis. $\log p(y\mid r,x)=-10$ is used when the model fails to answer (\ie when no answer token is generated and the answer-confidence term cannot be computed). We compared this fallback value ($-10$) with various other values, and the results are in Appendix~\ref{sec:fallback}.
For Llama-3.1-8B on the MATH500 dataset, we see that correct answers (green) are concentrated in the upper-right quadrant (Q1), corresponding to high scores on \emph{both} confidence terms.

The quadrant-wise accuracy breakdown is stark: the upper-right quadrant (Q1) achieves 71.7\% accuracy, outperforming other quadrants (Q2: 39.0\%, Q3: 31.6\%, Q4: 62.2\%).
High reasoning confidence (Q1 and Q4) leads to a higher performance than a high answer confidence (Q2 and Q3).
This is reinforced by a statistical t-test that, while both terms are highly significant predictors of correctness, reasoning confidence is a significantly stronger predictor ($\text{t-statistics} \approx\!9.111$) than answer confidence ($\text{t-statistics} \approx\!4.753$). For more details on the statistical tests, see \cref{sec:statsmodel}.
Nevertheless, both 
confidence measures remain essential components for reasoning chain selection. 

This principle can be used as a practical filter; tightening the thresholds to the 75th percentile, for instance, isolates a subset of samples with near-perfect accuracy (\ie 100\% on DS-Distilled-Qwen-2.5-7B with AIME2025), providing a mechanism to identify reliable instances (further examples in \cref{sec:information_plane_selection}). Overall, our analysis reveals that correct reasoning exhibits higher reasoning and answer confidence, \emph{with reasoning confidence being a substantially stronger predictor of correctness.}

\section{Experimental Setup}
\begin{table*}[t] \centering
\setlength{\tabcolsep}{3pt} 
\renewcommand{\arraystretch}{0.85} 
\resizebox{1.01\textwidth}{!}{
\begin{tabular}{l c c c c c c c c c r}
\toprule
\textbf{Method} & \multicolumn{2}{c}{\textbf{SVAMP}} & \multicolumn{2}{c}{\textbf{GSM8K}} & \multicolumn{2}{c}{\textbf{MATH500}} & \multicolumn{2}{c}{\textbf{GPQA-Diamond}} &
\multicolumn{2}{c}{\textbf{TheoremQA}} \\
& $k=6$ & $k=16/32$ & $k=6$ & $k=16/32$ & $k=6$ & $k=16/32$ & $k=6$ & $k=16/32$ & $k=6$ & $k=16/32$\\
\midrule

\multicolumn{11}{c}{\emph{Gemma-2-9B-Instruct}} \\
\midrule
Greedy Decoding
  & \multicolumn{2}{c}{$87.33$}
  & \multicolumn{2}{c}{$86.64$}
  & \multicolumn{2}{c}{$41.40$}
  & \multicolumn{2}{c}{$29.80$} 
  & \multicolumn{2}{c}{$17.14$}\\
Self-Consistency & $88.15{\scriptstyle\pm 0.22}$ & $88.89{\scriptstyle\pm 0.22}$ & $87.04{\scriptstyle\pm 0.24}$ & $88.10{\scriptstyle\pm 0.05}$ & $41.60{\scriptstyle\pm 0.40}$ & $43.27{\scriptstyle\pm 0.23}$ & $27.27{\scriptstyle\pm 0.58}$ & $23.91{\scriptstyle\pm 1.38}$ &
$15.44{\scriptstyle\pm 0.12}$ &
$14.10{\scriptstyle\pm 0.00}$ \\
USC & $88.63 {\scriptstyle\pm 0.13}$ & - & $85.74 {\scriptstyle\pm 0.27}$& - & $42.54 {\scriptstyle\pm 0.37}$ & - & $ 24.33{\scriptstyle\pm 1.21}$ & - & $ 17.24{\scriptstyle\pm 0.33}$ & - \\
\emph{p(True)}  & $88.56{\scriptstyle\pm 0.44}$ & $87.89{\scriptstyle\pm 0.22}$ & $88.36{\scriptstyle\pm 0.22}$ & $88.38{\scriptstyle\pm 0.08}$ & $\bf 46.87$${\scriptstyle\pm 0.07}$ & $46.80{\scriptstyle\pm 0.70}$ & $30.30{\scriptstyle\pm 1.54}$ & $33.50{\scriptstyle\pm 0.17}$ & $15.62{\scriptstyle\pm 0.37}$ & $15.98{\scriptstyle\pm 0.44}$ \\
Self-Certainty  & $88.48{\scriptstyle\pm 0.04}$ & $88.33{\scriptstyle\pm 0.06}$ & $87.18{\scriptstyle\pm 0.08}$ & $87.32{\scriptstyle\pm 0.03}$ & $43.93{\scriptstyle\pm 0.13}$ & $43.93{\scriptstyle\pm 0.08}$ & $26.77{\scriptstyle\pm 0.42}$ & $27.41{\scriptstyle\pm 0.83}$ & 
$14.73{\scriptstyle\pm 0.28}$ & 
$14.77{\scriptstyle\pm 0.04}$ \\
\rowcolor{blue!10} \methodabbrev &$ \bf  89.00{\scriptstyle\pm 0.38}^{*}$ & $\bf 91.02$${\scriptstyle\pm 0.59}$ & $ \bf 88.66$${\scriptstyle\pm 0.11}^{*}$ & $ \bf 88.99$${\scriptstyle\pm 0.20}$ & $ 46.53$${\scriptstyle\pm 0.29}^{*}$ & $\bf 47.13$${\scriptstyle\pm 0.13}$ & $ \bf 32.32$${\scriptstyle\pm 0.51}^{*}$ & $ \bf 34.01$${\scriptstyle\pm 1.94}$ & $ \bf 18.62$${\scriptstyle\pm 0.39}^{*}$ & $\bf 18.88$${\scriptstyle\pm 0.54}$\\

Upper Bound & $93.44{\scriptstyle\pm 0.22}$ & $95.67{\scriptstyle\pm 0.38}$ & $93.44{\scriptstyle\pm 0.09}$ & $95.60{\scriptstyle\pm 0.04}$ & $58.47{\scriptstyle\pm 0.27}$ & $66.67{\scriptstyle\pm 0.47}$ & $55.22{\scriptstyle\pm 1.10}$ & $82.49{\scriptstyle\pm 1.02}$ &
$24.32{\scriptstyle\pm 0.49}$ & 
${32.40\scriptstyle\pm 0.20}$ \\

\midrule
\multicolumn{11}{c}{\emph{Llama-3.1-8B-Instruct}} \\
\midrule
Greedy Decoding
  & \multicolumn{2}{c}{$89.67$}
  & \multicolumn{2}{c}{$87.47$}
  & \multicolumn{2}{c}{$50.40$}
  & \multicolumn{2}{c}{$27.27$} 
  & \multicolumn{2}{c}{$17.80$}\\
Self-Consistency & $88.33{\scriptstyle\pm 0.67}$ & $89.89{\scriptstyle\pm 0.11}$ & $86.67{\scriptstyle\pm 0.38}$ & $89.52{\scriptstyle\pm 0.16}$ & $46.33{\scriptstyle\pm 0.13}$ & $50.13{\scriptstyle\pm 0.48}$ & $26.09{\scriptstyle\pm 0.45}$ & $26.67{\scriptstyle\pm 1.34}$ &
$15.62{\scriptstyle\pm 0.18}$ &
$12.72{\scriptstyle\pm 0.48}$ \\
USC & $89.87 {\scriptstyle\pm 0.23}$ & - & $88.22 {\scriptstyle\pm 0.23}$ & - & $51.80 {\scriptstyle\pm 1.25}$ & - & $25.67{\scriptstyle\pm 1.54}$ & - & $18.88{\scriptstyle\pm 0.31}$ & -\\
\emph{p(True)}  & $85.33{\scriptstyle\pm 0.00}$ & $83.22{\scriptstyle\pm 0.91}$ & $87.40{\scriptstyle\pm 0.44}$ & $86.59{\scriptstyle\pm 0.03}$ & $47.73{\scriptstyle\pm 0.66}$ & $47.80{\scriptstyle\pm 0.72}$ & $27.27{\scriptstyle\pm 1.75}$ & $26.09{\scriptstyle\pm 2.07}$ & $14.41{\scriptstyle\pm 0.59}$ & $14.10{\scriptstyle\pm 0.51}$ \\
Self-Certainty  & $89.44{\scriptstyle\pm 0.06}$ & $89.49{\scriptstyle\pm 0.26}$ & $87.43{\scriptstyle\pm 0.24}$ & $87.35{\scriptstyle\pm 0.02}$ & $51.04{\scriptstyle\pm 0.20}$ & $51.09{\scriptstyle\pm 0.16}$ & $26.54{\scriptstyle\pm 0.49}$ & $26.30{\scriptstyle\pm 0.49}$  &
$14.91{\scriptstyle\pm 0.13}$ & 
$14.62{\scriptstyle\pm 0.14}$ \\
\rowcolor{blue!10} \methodabbrev & $\bf 91.78$${\scriptstyle\pm 0.11}^{*}$ & $\bf 93.44$${\scriptstyle\pm 0.89}$ & $\bf 89.09$${\scriptstyle\pm 0.13}^{*}$ & $\bf 89.98$${\scriptstyle\pm 0.23}$ & $\bf 53.33$${\scriptstyle\pm 0.73}^{*}$ & $\bf 53.87$${\scriptstyle\pm 0.70}$ & $ \bf 29.80$${\scriptstyle\pm 1.34}^{*}$ & $\bf  33.67$${\scriptstyle\pm 3.06}$ & $\bf 20.08$${\scriptstyle\pm 0.43}^{*}$ & $\bf 19.72$${\scriptstyle\pm 0.39}$ \\
Upper Bound & $ 96.78$${\scriptstyle\pm 0.11}$ & $ 99.11$${\scriptstyle\pm 0.11}$ & $96.15$${\scriptstyle\pm 0.07}$ & $ 98.18$${\scriptstyle\pm 0.04}$ & $72.80$${\scriptstyle\pm 0.23}$ & $ 82.20$${\scriptstyle\pm 0.60}$ & $65.82{\scriptstyle\pm 1.50}$ & $92.76{\scriptstyle\pm 0.73}$ & $28.20{\scriptstyle\pm 0.32}$ & $37.84{\scriptstyle\pm 1.13}$ \\
\midrule
\multicolumn{11}{c}{\emph{Qwen3-8B (Non-thinking)}} \\
\midrule
Greedy Decoding
  & \multicolumn{2}{c}{$93.33$}
  & \multicolumn{2}{c}{$92.48$}
  & \multicolumn{2}{c}{$\underline{\mathbf{73.40}}$}
  & \multicolumn{2}{c}{$42.23$} 
  & \multicolumn{2}{c}{27.71}\\
Self-Consistency & $92.52{\scriptstyle\pm 0.33}$ & $93.11{\scriptstyle\pm 0.11}$ & $92.29{\scriptstyle\pm 0.13}$ & $91.69{\scriptstyle\pm 0.11}$ & $73.00{\scriptstyle\pm 0.23}$ & $72.27{\scriptstyle\pm 0.00}$ & $ 47.47$${\scriptstyle\pm 0.29}$ & $40.74{\scriptstyle\pm 1.61}$ 
& $28.33{\scriptstyle\pm 0.31}$
& $28.51{\scriptstyle\pm 0.33}$
\\
USC & $93.11{\scriptstyle\pm 0.22}$ & - & $ \bf 93.24$${\scriptstyle\pm 0.13}$ & - & $73.60{\scriptstyle\pm 0.12}$ & - & $\bf 48.38$${\scriptstyle\pm 2.06}$ & 
- & $27.88$${\scriptstyle\pm 0.55}$ & - \\
\emph{p(True)}  & $92.44{\scriptstyle\pm 0.56}$ & $91.78{\scriptstyle\pm 0.44}$ & $92.10{\scriptstyle\pm 0.00}$ & $91.22{\scriptstyle\pm 0.18}$ & $72.67{\scriptstyle\pm 0.24}$ & $71.20{\scriptstyle\pm 0.60}$ & $41.25{\scriptstyle\pm 1.71}$ & $36.20{\scriptstyle\pm 1.44}$ & $27.84{\scriptstyle\pm 0.18}$ & $28.28{\scriptstyle\pm 0.13}$ \\
Self-Certainty  & $92.63{\scriptstyle\pm 0.21}$ & $92.83{\scriptstyle\pm 0.04}$ & $92.29{\scriptstyle\pm 0.07}$ & $92.25{\scriptstyle\pm 0.04}$ & $71.94{\scriptstyle\pm 0.16}$ & $71.82{\scriptstyle\pm 0.14}$ &  $44.33{\scriptstyle\pm 0.54}$ & $42.29{\scriptstyle\pm 0.81}$ & 
$27.97{\scriptstyle\pm 0.66}$ & 
$27.92{\scriptstyle\pm 0.77}$\\
\rowcolor{blue!10} \methodabbrev & $ \bf 93.56$${\scriptstyle\pm 0.22}^{*}$ & $\bf 95.13$${\scriptstyle\pm 0.22}$ & $92.33$${\scriptstyle\pm 0.13}^{*}$ & $ \bf 93.22$${\scriptstyle\pm 0.08}$ & $ \bf 73.67$${\scriptstyle\pm 0.24}^{*}$ & $\bf \underline{ 73.40}$${\scriptstyle\pm 0.13}$ & $46.98{\scriptstyle\pm 1.01}^{*}$ & 
$\bf 43.69$${\scriptstyle\pm 1.26}$ & $\bf 29.76{\scriptstyle\pm 0.58}^{*}$ & 
$\bf 29.17$${\scriptstyle\pm 0.64}$ 
\\
Upper Bound & $96.33{\scriptstyle\pm 0.67}$ & $97.89{\scriptstyle\pm 0.11}$ & $95.52{\scriptstyle\pm 0.00}$ & $96.84{\scriptstyle\pm 0.03}$ & $81.13{\scriptstyle\pm 0.44}$ & $83.53{\scriptstyle\pm 0.24}$ & $76.26{\scriptstyle\pm 1.62}$ & $86.36{\scriptstyle\pm 0.29}$ &
$34.94{\scriptstyle\pm 0.00}$ &
$40.03{\scriptstyle\pm 0.35}$ \\

\midrule
\multicolumn{11}{c}{\emph{Llama-3.1-70B-Instruct}} \\
\midrule
Greedy Decoding
  & \multicolumn{2}{c}{$\bf 94.33$}
  & \multicolumn{2}{c}{$93.93$}
  & \multicolumn{2}{c}{$60.20$}
  & \multicolumn{2}{c}{$40.44$} 
  & \multicolumn{2}{c}{$\bf 30.79$}\\

Self-Consistency & $92.78{\scriptstyle\pm 0.56}$ & $93.45{\scriptstyle\pm 0.11}$ & $94.00{\scriptstyle\pm 0.10}$ & $93.98{\scriptstyle\pm 0.13}$ & $58.60{\scriptstyle\pm 0.46}$ & $60.80{\scriptstyle\pm 0.87}$ & $42.59{\scriptstyle\pm 1.02}$ & $37.54{\scriptstyle\pm 0.67}$ &
$26.55{\scriptstyle\pm 0.47}$ &
$25.61{\scriptstyle\pm 0.00}$ \\
USC & $92.78{\scriptstyle\pm 0.11}$ & - & $93.29{\scriptstyle\pm 0.20}$ & - & $60.60{\scriptstyle\pm 0.95}$ & - & $41.25{\scriptstyle\pm 1.76}$ & - & 
$27.44{\scriptstyle\pm 0.67}$ & - \\
\emph{p(True)}  & $93.11{\scriptstyle\pm 0.78}$ & $93.11{\scriptstyle\pm 0.40}$ & $94.51{\scriptstyle\pm 0.13}$ & $94.08{\scriptstyle\pm 0.23}$ & $61.47{\scriptstyle\pm 1.14}$ & $62.33{\scriptstyle\pm 1.16}$ & $41.25{\scriptstyle\pm 1.61}$ & $42.09{\scriptstyle\pm 2.21}$ & $24.45{\scriptstyle\pm 0.31}$ & $24.23{\scriptstyle\pm 0.61}$ \\
Self-Certainty  & $93.02{\scriptstyle\pm 0.30}$ & $93.84{\scriptstyle\pm 0.01}$ & $94.01{\scriptstyle\pm 0.13}$ & $94.04{\scriptstyle\pm 0.05}$ & $61.82{\scriptstyle\pm 0.08}$ & $61.70{\scriptstyle\pm 0.14}$ &  $39.84{\scriptstyle\pm 0.88}$ & $38.87{\scriptstyle\pm 0.67}$ &  $24.43{\scriptstyle\pm 0.18}$ & $24.56{\scriptstyle\pm 0.11}$ \\
\rowcolor{blue!10} \methodabbrev & $ 94.10{\scriptstyle\pm 0.11}^{*}$ & $\bf 95.58$${\scriptstyle\pm 0.22}$ & $\bf 94.58$${\scriptstyle\pm 0.03}^{*}$ & $\bf 94.81$${\scriptstyle\pm 0.13}$ & $ \bf 63.67$${\scriptstyle\pm 1.51}^{*}$ & $ \bf 64.07$${\scriptstyle\pm 0.87}$ & $ \bf 46.91$${\scriptstyle\pm 2.65}^{*}$ & $\bf 46.46$${\scriptstyle\pm 2.59}$ & $ 27.84$${\scriptstyle\pm 0.19}^{*}$ & \bf $26.73$${\scriptstyle\pm 0.27}$ \\

Upper Bound & $97.22{\scriptstyle\pm 0.22}$ & $97.78{\scriptstyle\pm 0.22}$ & $96.91{\scriptstyle\pm 0.03}$ & $97.44{\scriptstyle\pm 0.03}$ & $77.07{\scriptstyle\pm 0.47}$ & $81.67{\scriptstyle\pm 0.18}$ & $75.59{\scriptstyle\pm 0.61}$ & $87.71{\scriptstyle\pm 0.45}$ &
$40.70{\scriptstyle\pm 0.20}$ &
$43.47{\scriptstyle\pm 0.18}$ \\

\midrule
\multicolumn{11}{c}{\emph{Qwen3-32B (Non-thinking)}} \\
\midrule
Greedy decoding
  & \multicolumn{2}{c}{$92.33$}
  & \multicolumn{2}{c}{$93.24$}
  & \multicolumn{2}{c}{$75.00$}
  & \multicolumn{2}{c}{$\bf 48.48$} 
  & \multicolumn{2}{c}{$29.99$}\\

Self-consistency & $92.67{\scriptstyle\pm 0.33}$ & $93.11{\scriptstyle\pm 0.33}$ & $93.62{\scriptstyle\pm 0.00}$ & $93.75{\scriptstyle\pm 0.08}$ & $75.93{\scriptstyle\pm 0.33}$ & $\bf 76.27$${\scriptstyle\pm 0.12}$ & $ 47.31{\scriptstyle\pm 1.98}$ & $44.44$${\scriptstyle\pm 0.51}$ &
$30.79{\scriptstyle\pm 0.00}$ & $30.92{\scriptstyle\pm 0.28}$ \\

USC & $92.44{\scriptstyle\pm 0.78}$ & - & $93.69{\scriptstyle\pm 0.13}$ & - & $76.16{\scriptstyle\pm 0.64}$ & - & $44.90{\scriptstyle\pm 0.55}$ & - &
$30.07{\scriptstyle\pm 0.51}$ & - \\
\emph{p(True)}  & $\bf \underline{93.22}$${\scriptstyle\pm 0.11}$ & $93.00{\scriptstyle\pm 0.69}$ & $92.79{\scriptstyle\pm 0.53}$ & $92.91{\scriptstyle\pm 0.25}$ & $74.07{\scriptstyle\pm 1.07}$ & $74.00{\scriptstyle\pm 0.35}$ & $39.90{\scriptstyle\pm 2.81}$ & $38.05{\scriptstyle\pm 0.94}$ & $30.79{\scriptstyle\pm 0.00}$ & $30.08{\scriptstyle\pm 0.12}$ \\
Self-certainty  & $92.63{\scriptstyle\pm 0.18}$ & $92.92{\scriptstyle\pm 0.16}$ & $92.29{\scriptstyle\pm 0.03}$ & $93.45{\scriptstyle\pm 0.02}$ & $71.94{\scriptstyle\pm 0.09}$ & $75.68{\scriptstyle\pm 0.10}$ &  $43.07{\scriptstyle\pm 1.16}$ & $43.39{\scriptstyle\pm 0.73}$ & $30.23{\scriptstyle\pm 0.00}$ & $30.61{\scriptstyle\pm 0.13}$ \\
\rowcolor{blue!10} \methodabbrev & $ \bf \underline{93.22}$${\scriptstyle\pm 0.22}^{*}$ & $ \bf 93.55$${\scriptstyle\pm 0.33}$ & $ \bf 93.90$${\scriptstyle\pm 0.28}^{*}$ & $ \bf 93.88$${\scriptstyle\pm 0.22}$ & $\bf 77.00$${\scriptstyle\pm 0.18}^{*}$ & $ 75.93{\scriptstyle\pm 0.13}$ & $ 46.91$${\scriptstyle\pm 1.02}^{*}$ & $44.44$${\scriptstyle\pm 2.28}$ & $\bf 31.46$${\scriptstyle\pm 0.04}^{*}$ &  $\bf  31.42$${\scriptstyle\pm 0.27}$
\\
Upper Bound & $96.78{\scriptstyle\pm 0.11}$ & $98.00{\scriptstyle\pm 0.00}$ & $96.28{\scriptstyle\pm 0.13}$ & $96.99{\scriptstyle\pm 0.07}$ & $82.27{\scriptstyle\pm 0.13}$ & $83.73{\scriptstyle\pm 0.07}$ & $72.56{\scriptstyle\pm 1.87}$ & $86.20{\scriptstyle\pm 1.02}$ & $39.76{\scriptstyle\pm 0.00}$ & $42.93{\scriptstyle\pm 0.12}$ \\

\bottomrule
\end{tabular}
}
\caption{
\textbf{Comparison of model accuracies 
on LLMs.} 
\(k=\{6,32\}\) for Gemma-2-9B, Llama-3.1-8B, Qwen3-8B; \(k=\{6,16\}\) for Llama-3.1-70B, Qwen3-32B due to computational constraints. \textbf{Bold}: highest, \textbf{\underline{underline}}: equal highest, $^{*}$: \(k=6\) outperforms \(k=16,32\) baselines. \emph{\methodabbrev with \(k=6\) outperforms larger \(k\) in 20/25 cases.
}
}
\label{tab:model-comparison}
\vspace{-2ex}
\end{table*}
\label{section:experimental setup}

\paragraph{Models.}
We evaluate PiCSAR across a diverse set of recent LLMs and LRMs. Our experiments include LLMs from three major families: Llama-3.1-Instruct (8B and 70B;~\citealt{dubey2024llama}), Gemma-2-Instruct (9B;~\citealt{team2024gemma}), and Qwen3 (8B and 32B;~\citealt{yang2025qwen3}). For the Qwen3 models, we disable the \emph{thinking mode}. 
For LRMs, we include two distilled models from the DeepSeek-R1 series (DS-distill-Llama-3.1-8B and DS-distill-Qwen-2.5b;~\citealt{guo2025deepseek}), and the {Qwen-3-8B} model with \emph{thinking mode} enabled. We exclude larger LRMs due to computational cost.

\paragraph{Baselines.}
We compare against six baselines:  \emph{Greedy Decoding} (1); \emph{Self-Consistency}~\citep{wang2022self} (2);  \emph{USC}~\citep{chen2023universal} (3);  \emph{p(True)}~\citep{kadavath2022language} (4);  \emph{Self-Certainty}~\citep{kang2025scalable} (5).  
Confidence-Interval Self-Consistency (CISC; \citealt{taubenfeld2025confidence}) is discussed in \cref{sec:cisc-comparison}, as it involves weight voting. While CISC was originally proposed using \emph{p(True)}, we also report CISC(\methodabbrev) for a fair comparison.
Due to context length limits and computational constraints, we exclude (3), (4), and (5) in LRMs and set \(k=16,32\) in LLMs. 

To isolate each component’s contribution in \methodabbrev, we include three ablations
in~\cref{sec:breakdown} and \ref{app:picsar-n}: \emph{Reasoning Confidence} ($ \max_r(\log p(r \mid x))$), with (6) and without (7) length normalisation, and \emph{Answer Confidence} ($\max_y(\log p(y \mid r, x))$) (8). For LRMs,
we compare against (1), (2), (6), (7), (8).
We also include the $pass@k$ upper bound, representing the maximum achievable accuracy when at least one of the $k$ candidates is correct.
Implementation details can be found in~\cref{appendix:implementation}.

\begin{table*}[t]
\centering
\resizebox{0.95\textwidth}{!}{
\renewcommand{\arraystretch}{0.85}
\begin{tabular}{llccccccc}
\toprule
\textbf{Method} & \textbf{SVAMP} & \textbf{GSM8K} & \textbf{MATH500} & \textbf{GPQA-Diamond} & \textbf{TheoremQA} & \textbf{AIME\,2024} & \textbf{AIME\,2025} \\
\midrule
\multicolumn{7}{c}{\emph{DS-Distill-llama-3-8B}} \\
\midrule
Average & $82.11{\scriptstyle\pm 0.13}$ & $73.67{\scriptstyle\pm 0.32}$ & $65.55{\scriptstyle\pm 0.25}$ & $42.87{\scriptstyle\pm 1.07}$ & $26.58{\scriptstyle\pm 0.06}$ & $37.96{\scriptstyle\pm 1.52}$ & $29.63{\scriptstyle\pm 0.37}$ \\
Self-Consistency & $\bf 86.17$${\scriptstyle\pm 0.27}$ & $74.01{\scriptstyle\pm 0.70}$ & $66.25{\scriptstyle\pm 0.40}$ & $42.10{\scriptstyle\pm 1.77}$ & $27.98{\scriptstyle\pm 0.87}$ & $38.89{\scriptstyle\pm 1.67}$ & $25.00{\scriptstyle\pm 0.37}$ \\
\rowcolor{blue!10}
\methodabbrev & $85.67{\scriptstyle\pm 0.07}$ & $\bf76.42$${\scriptstyle\pm 0.16}$ & $\bf 67.20$${\scriptstyle\pm 0.60}$ & $\bf 47.31$${\scriptstyle\pm 0.17}$ & $\bf 28.02$${\scriptstyle\pm 0.78}$ & $\bf 47.78$${\scriptstyle\pm 4.01}$ & $\bf 33.33$${\scriptstyle\pm 1.11}$ \\
Upper Bound & $95.67{\scriptstyle\pm 0.00}$ & $92.91{\scriptstyle\pm 0.35}$ & $82.00{\scriptstyle\pm 0.13}$ & $77.27{\scriptstyle\pm 0.77}$ & $36.37{\scriptstyle\pm 2.83}$ & $66.67{\scriptstyle\pm 5.09}$ & $51.11{\scriptstyle\pm 1.11}$ \\
\midrule
\multicolumn{7}{c}{\emph{DS-Distill-Qwen-2.5-7B}} \\
\midrule
Average & $89.26{\scriptstyle\pm 0.13}$ & $87.29{\scriptstyle\pm 0.14}$ & $72.79{\scriptstyle\pm 0.16}$ & $46.44{\scriptstyle\pm 1.63}$ & $33.11{\scriptstyle\pm 0.14}$ & $49.44{\scriptstyle\pm 3.06}$ & $41.30{\scriptstyle\pm 1.30}$ \\
Self-Consistency & $90.39{\scriptstyle\pm 0.20}$ & $\bf 89.50$${\scriptstyle\pm 0.37}$ & $73.87{\scriptstyle\pm 0.25}$ & $44.78{\scriptstyle\pm 1.83}$ & $35.88{\scriptstyle\pm 0.35}$ & $47.78{\scriptstyle\pm 3.40}$ & $38.33{\scriptstyle\pm 3.34}$ \\
\rowcolor{blue!10}
\methodabbrev & $\bf 91.78$${\scriptstyle\pm 0.48}$ & $88.18{\scriptstyle\pm 0.07}$ & $\bf 74.00$${\scriptstyle\pm 0.70}$ & $\bf 52.36$${\scriptstyle\pm 2.88}$ & $\bf 36.76$${\scriptstyle\pm 0.44}$ & $\bf 61.11$${\scriptstyle\pm 1.11}$ & $\bf 51.11$${\scriptstyle\pm 1.11}$ \\
Upper Bound & $96.33{\scriptstyle\pm 0.38}$ & $96.79{\scriptstyle\pm 0.13}$ & $83.33{\scriptstyle\pm 0.18}$ & $79.12{\scriptstyle\pm 2.07}$ & $48.59{\scriptstyle\pm 0.08}$ & $72.22{\scriptstyle\pm 1.11}$ & $70.00{\scriptstyle\pm 0.00}$ \\
\midrule
\multicolumn{7}{c}{\emph{Qwen3-8B}} \\
\midrule
Average & $91.43{\scriptstyle\pm 0.07}$ & $95.43{\scriptstyle\pm 0.01}$ & $80.44{\scriptstyle\pm 0.10}$ & $54.21{\scriptstyle\pm 0.83}$ & $40.83{\scriptstyle\pm 0.13}$ & $75.37{\scriptstyle\pm 0.19}$ & $67.04{\scriptstyle\pm 2.06}$ \\
Self-Consistency & $91.83{\scriptstyle\pm 0.33}$ & $95.68{\scriptstyle\pm 0.03}$ & $80.40{\scriptstyle\pm 0.18}$ & $54.21{\scriptstyle\pm 1.68}$ & $41.81{\scriptstyle\pm 0.11}$ & $77.23{\scriptstyle\pm 1.11}$ & $65.56{\scriptstyle\pm 2.58}$ \\
\rowcolor{blue!10}
\methodabbrev & $\bf 94.33$${\scriptstyle\pm 0.33}$ & $\bf 95.94$${\scriptstyle\pm 0.04}$ & $\bf 80.60$${\scriptstyle\pm 0.13}$ & $\bf 59.43$${\scriptstyle\pm 1.61}$ & $\bf 42.57$${\scriptstyle\pm 0.27}$ & $\bf 81.33$${\scriptstyle\pm 1.34}$ & $\bf 68.89$${\scriptstyle\pm 2.22}$ \\
Upper Bound & $97.56{\scriptstyle\pm 0.11}$ & $97.54{\scriptstyle\pm 0.03}$ & $84.00{\scriptstyle\pm 0.12}$ & $80.13{\scriptstyle\pm 0.45}$ & $44.71{\scriptstyle\pm 1.34}$ & $87.78{\scriptstyle\pm 1.11}$ & $82.22{\scriptstyle\pm 1.11}$ \\
\bottomrule
\end{tabular}
}
\caption{
\textbf{Comparison of model accuracies 
on LRMs} ($k=6$). \emph{PiCSAR outperforms the baselines in 19/21 cases.}}
\label{tab:deepseek_qwen}
\vspace{-2ex}
\end{table*}

\paragraph{Datasets.}
We evaluate LLMs on three maths benchmarks: GSM8K~\citep{cobbe2021training}, SVAMP~\citep{patel2021nlp}, MATH500~\citep{hendrycks2021measuring}, and two  scientific reasoning benchmarks GPQA-Diamond~\citep{rein2024gpqa}, and TheoremQA~\citep{chen2023theoremqa}. We additionally evaluate LRMs on AIME 2024 and 2025, omitted for LLMs due to difficulty.
Results are averaged over three runs and reported with standard errors.


\section{Experimental Results} 
\label{sec:LLM results}

\paragraph{Performance on LLMs.}
In Table~\ref{tab:model-comparison}, we see that when using \methodabbrev, Llama models show consistent improvements across all baselines. With $k=6$ on Llama-3.1-8B, \methodabbrev outperforms the best-performing baseline (\ie Self-Certainty) by 3.26\% 
(26.54\% $\rightarrow$ 29.80\%) on GPQA-Diamond. On Llama-3.1-70B \methodabbrev shows similar gains: 7.07\% improvement over Self-Certainty and 5.66\% over USC. 
We observe a similar trend on Gemma-2-9B; at $k{=}6$, \methodabbrev outperforms Self-Consistency by 4.93\%. This outcome aligns with our information-plane analysis (see \cref{fig:info_plane_llama_math500}); \methodabbrev selects candidates in the top-right, high-accuracy quadrant by maximising the joint score of reasoning and answer confidence.
For the Qwen family, \methodabbrev generally leads across benchmarks and sample counts ($k$). 
While there are a few exceptions,
\methodabbrev maintains the strongest overall profile. For instance, on MATH500 with $k=6$, it improves the accuracy of Qwen3-32B from 75.93\% (Self-Consistency) to 77.00\%. 

Our results show that \methodabbrev outperforms most existing baselines and datasets, demonstrating consistent improvements across various reasoning tasks. As shown in~\cref{sec:cisc-comparison}, CISC (\methodabbrev) consistently outperforms CISC (\emph{p(True)}), indicating its potential for weighting augmentation, but detailed voting strategy analysis remains future work. To verify the statistical significance of our results, we perform the Friedman test~\cite{demvsar2006statistical}, returning a p-value of $\sim\!6e^{-17}$, followed by the post-hoc Nemenyi test, which confirms that PiCSAR significantly outperforms all baselines (more in Appendix~\ref{sec:nemenyi-test}).
\emph{ These findings validate our hypothesis that the model's confidence provides more informative clues than frequency-based selection.}

\emph{\methodabbrev is also sample efficient.} \methodabbrev with a small sampling budget ($k=6$) frequently outperforms both Self-Consistency and Self-Certainty at higher sampling budgets ($k=16, 32$), narrowing the gap to the upper bound by detecting correct reasoning even within a small sample. For instance, Gemma-2-9B Instruct with $k=6$ ($46.53\%)$ outperforms $k=32$ ($43.27\%$). This indicates that correct reasoning chains are often present in small candidate sets, and that better selection is more important than increased sampling. See Appendix~\ref{sec:upperbound} for details of the upper bound analysis.



Overall, the joint score acts as a paired scoring function: the \emph{reasoning confidence} 
provides an assessment of plausibility towards its own reasoning, while the \emph{answer confidence},
focused on the final answer, serves as a fine-grained discriminator. 
This approach yields consistent improvements across evaluated models.

\paragraph{Performance on LRMs. }
Table~\ref{tab:deepseek_qwen} reports results from the LRMs. 
Across 19 out of 21 comparisons, \methodabbrev outperforms all baselines.
Relative to Self-Consistency, DS-Distill-Llama-3-8B demonstrates substantial improvements on AIME2024 (8.89\%) and AIME2025 (8.33\%). 
DS-Distill-Qwen-2.5-7B shows greater improvements compared to Self-Consistency,  
with an improvement of 12.33\% on AIME2024 and of 12.78\% on AIME2025.
When applied on a relatively more capable model such as Qwen3-8B, \methodabbrev increases accuracy by 4.1\% and 3.33\% on AIME\,2024 and AIME\,2025, respectively. 
While improvements on previously evaluated benchmarks (MATH500, SVAMP, GSM8K) yield smaller gains, we observe substantial improvements on GPQA-Diamond: 5.21\%, 7.58\%, and 5.22\% for DS-Distill-Llama-3-8B, DS-Distill-Qwen-2.5-7B, and Qwen3-8B, respectively.
These trends mirror those observed with LLMs: gains are most pronounced on challenging datasets where the models' initial baseline accuracies are relatively lower. 
The Friedman and post-hoc Nemenyi testing additionally confirm that PiCSAR significantly outperforms all baselines (see Appendix~\ref{sec:nemenyi-test}).

\emph{
\methodabbrev, 
validates the information plane principle in~\S\ref{sec:empirical_justification} and provides a scoring method that improves accuracy both for LLMs and LRMs.}

\paragraph{Comparison with Trained Reward Models.} 
While our primary baselines consist of training-free BoN methods, a critical question is how \methodabbrev compares to explicitly trained verifiers. To establish this, we benchmarked \methodabbrev against top-performing reward models on the RewardBench~\citep{lambert-etal-2025-rewardbench} leaderboard, specifically \emph{Skywork-Reward-V2-Llama-3.1-8B} and \emph{LMUnit-qwen2.5-72B}. Despite being a completely zero-shot, training-free method, \methodabbrev achieves parity with, and in several cases, outperforms these heavily trained reward models across both MATH500 and GSM8K. This confirms that \methodabbrev's probabilistic formulation extracts a signal as reliable as explicit preference tuning, but at zero training cost. Detailed empirical results and analysis for this comparison are provided in Appendix~\ref{sec:rm_comparison}.

\section{Further Analysis}
In our analysis, we study (1) how information density correlates with accuracy; (2) the confidence-accuracy relationship within each model; (3) the robustness of our confidence metric when generation and evaluation are decoupled.

\begin{table}[t]
\centering
\small
\setlength{\tabcolsep}{5.5pt} 
\renewcommand{\arraystretch}{0.85} 

\begin{tabular}{l c c c r}
\toprule
\textbf{Rank} & \textbf{Peaks} & \textbf{Sent.} & \textbf{Ratio} (\%) & \textbf{Acc.} (\%) \\
\midrule

\multicolumn{5}{c}{\textit{\textbf{Llama-3.1-8B}}} \\
\midrule
Highest & 1.88 & 16.4 & 14.8 & 53.3 \\
Middle  & 2.00 & 22.9 & 12.8 & 48.8 \\
Lowest  & 2.47 & 64.7 & \phantom{0}8.6  & 44.2 \\
\midrule

\multicolumn{5}{c}{\textit{\textbf{Llama-3.1-70B}}} \\
\midrule
Highest & 1.80 & 14.1 & 15.5 & 63.7 \\
Middle  & 1.83 & 19.9 & 13.0 & 60.4 \\
Lowest  & 3.08 & 38.4 & 10.8 & 59.4 \\
\midrule

\multicolumn{5}{c}{\textit{\textbf{Qwen3-8B}}} \\
\midrule
Highest & 1.99 & 15.8 & 17.6 & 73.7 \\
Middle  & 1.91 & 17.6 & 17.0 & 72.8 \\
Lowest  & 2.18 & 26.4 & 14.2 & 69.4 \\
\midrule

\multicolumn{5}{c}{\textit{\textbf{Qwen3-32B}}} \\
\midrule
Highest & 1.48 & 11.6 & 22.4 & 77.0 \\
Middle  & 1.57 & 12.0 & 19.4 & 76.8 \\
Lowest  & 1.76 & 25.1 & 16.1 & 72.6 \\
\midrule

\multicolumn{5}{c}{\textit{\textbf{Gemma-2-9B}}} \\
\midrule
Highest & 1.46 & \phantom{0}8.5  & 24.5 & 46.5 \\
Middle  & 1.38 & 10.0 & 19.0 & 44.0 \\
Lowest  & 1.20 & 11.6 & 14.3 & 41.6 \\

\bottomrule
\end{tabular}
\caption{Peak count analysis across different PiCSAR confidence rankings (highest, middle (3rd),
and lowest) for different models. \textbf{Peaks}: Average number of peaks; \textbf{Sent.}: Average sentence count; \textbf{Ratio}: Average Peak-to-Sentence ratio; \textbf{Acc.}: Model accuracy. Chains in the ``Highest'' rank consistently show a higher peak density (Ratio) compared to the ``Lowest'' rank.}
\label{tab:model-stats}
\vspace{-3ex}
\end{table}

\subsection{Sentence-Level Confidence Dynamics as a Proxy for Reasoning Quality}
\label{sec:section3}

To understand the dynamics of \methodabbrev, we analyse the evolution of answer confidence across reasoning chains.
For a given reasoning chain $r$ composed of sentences $(r^{1}, r^{2}, \ldots, r^{m})$ and its corresponding final answer $y$, we measure how the model's confidence in $y$ changes as it processes more of the reasoning. We compute a sequence of scores, $\log p(y \mid r^{1:j}, x)$, for each partial reasoning prefix $r^{1:j}$, where $j$ ranges from 1 to $m$. 
To capture the characteristics of these confidence sequences, we rank the responses by \methodabbrev scoring function into three groups (highest, middle, lowest), and analyse the ``peakiness'' of the confidence trajectory within each group. 
We define a \emph{peak} as a sentence where the confidence $\log p(y \mid r^{1:j},x)$ exceeds the 95th percentile of all sentence-level scores observed across reasoning chains with the correct answer for that particular problem. The \emph{peak-to-sentence ratio} is the peak count divided by the total sentences. We term this \emph{information density}: the proportion of reasoning sentences contributing meaningfully to answer confidence.

Table~\ref{tab:model-stats} shows:
(1) Higher peak-to-sentence ratio aligns with higher accuracy across different models, showing that \textit{reasoning chains that lead to the correct answer tend to have higher information density}.
For instance, Llama-3.1-8B achieves 53.33\% accuracy with a 14.75\% ratio in the highest-scoring group, compared to 44.20\% with only 8.58\% in the lowest;
(2) \textit{Longer reasoning chains do not necessarily improve accuracy}. 
The lowest-ranked responses are substantially longer yet less accurate. For example, Llama-3.1-8B averages 64.72 sentences with 44.20\% accuracy in the lowest group, versus 16.43 sentences with 53.33\% accuracy in the highest group.
This observation aligns with recent findings of inverse scaling in test-time compute~\citep{chen2024not, wu2025more, hassid2025don, ghosal2025does, gema2025inverse}, showing that solely extended reasoning length does not guarantee improved performance. 

As unnormalised \methodabbrev naturally rewards these high-density, convergent trajectories, it serves as our recommended default. Length normalisation (\methodabbrev-N) is typically only necessary when evaluating weaker models that are highly prone to ``verbose hallucinations'', where the model accumulates massive negative log-probabilities through unproductive, circular generation rather than meaningful reasoning. We provide a comprehensive sentence-level trajectory analysis detailing the exact criteria and decision boundary for enabling length normalisation in Appendix~\ref{sec: picsarn_length} and Appendix~\ref{appendix:sentence_level_analysis}.

\subsection{Intra-model Confidence Duality}
\label{intra-inter-confidence}

\begin{figure}[t]
  \centering
  \includegraphics[width=0.48\textwidth]{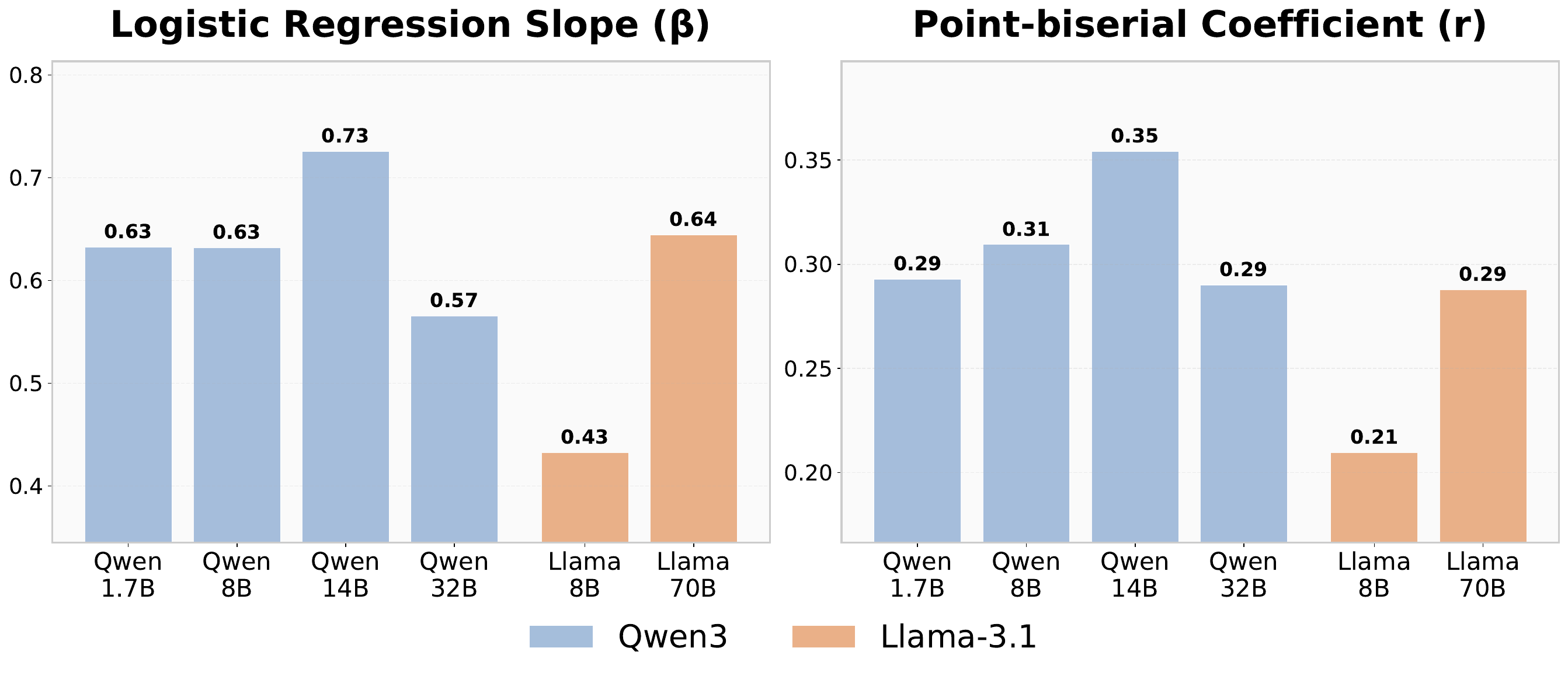}
  \caption{Calibration summary for Qwen3 and Llama-3.1-8B models. We show that the $\beta$ and r coefficients are consistently positive across all models.}
  \label{fig:qwen-calibration-wrap}
  \vspace{-3ex}
\end{figure}

In this section, we investigate the reliability of \methodabbrev for predicting correctness within individual models (\emph{intra-model reliability analysis}). We further examine whether these confidence scores remain comparable across different models (\emph{inter-model variance analysis}) in ~\cref{sec:intermodel}. 
We fit regressions for the Qwen and Llama families (\cref{fig:qwen-calibration-wrap}), with correctness (correct/incorrect) as the dependent variable and the answer confidence score as the independent variable. This approach allows us to interpret the regression slope ($\beta$), which represents the incremental change in log-odds of correctness per unit increase in confidence score.

We find that the $\beta$ is consistently positive across all model sizes, consistent with prior findings
~\citet{huh2024platonic, goel2025great} of a strong positive relationship between confidence scores and their likelihood of being correct. For example, Qwen3-14B shows a $\beta$ of 0.7255, implying that each unit increase in log-probability more than doubles the odds of correctness ($e^{0.7255} \approx 2.07$). The Point-Biserial Correlation Coefficient further confirms the positive relationship by measuring the linear association between binary correctness and continuous confidence.
%
\emph{These findings show that \methodabbrev serves as a reliable predictor of correctness within each model.}
See \cref{sec:detailed-intra-model} for more details.

\subsection{Confidence Portability: Decoupling Generation from Evaluation}
\label{sec:decoupled_confidence}
Having established the properties of the confidence signal within a single model, we extend our analysis to multi-model scenarios, evaluating confidence signal robustness when generation and evaluation are decoupled. This decoupling is motivated by practical system design, where one might use a costly API model for reasoning confidence, while relying on a smaller local model for answer confidence estimation.
In this \emph{decoupled} setting,
the model that generates the reasoning chain ($M_{\text{gen}}$) differs from the model that evaluates the answer confidence ($M_{\text{eval}}$). The scoring function for a chain $r_i$ generated by $M_{\text{gen}}$ becomes:
\vspace{-0.5ex}
\begin{equation}
\begin{split}
\text{Score}(r_i, y_i)& = \underbrace{\log p(r_i \mid x; M_{\text{gen}})}_{\text{Generated by } M_{\text{gen}}} \\
&+ \underbrace{\log p(y_i \mid \langle a \rangle, r_i, x; M_{\text{eval}})}_{\text{Evaluated by } M_{\text{eval}}}.
\end{split}
\end{equation}
\begin{figure}[t]
  \centering
  \includegraphics[width=0.48\textwidth]{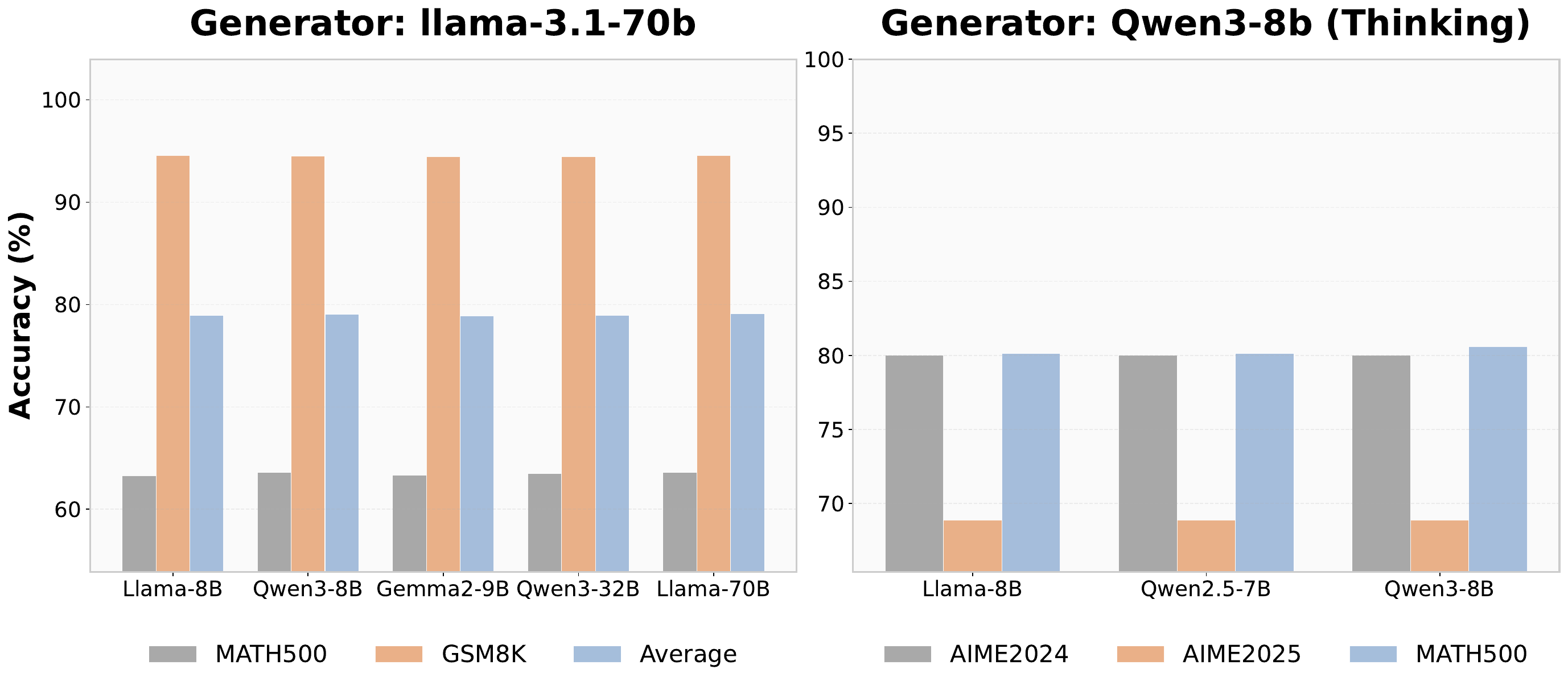}
  \vspace{-3ex}
  \caption{Decoupling analysis for Llama-3.1-70B and Qwen3-8B (Thinking Enabled) as $M_{\text{gen}}$, with various $M_{\text{eval}}$, showing performance remains similar when different models are used to estimate $\log p(y \mid r, x)$.}
  \label{fig:cross-pollination_single}
  \vspace{-3ex}
\end{figure}
We test this by having $M_{\text{gen}}$ generate reasoning chains, and various models acting as $M_{\text{eval}}$. For LRMs, the base instruction tuned model is used as $M_{\text{eval}}$. Results in Figure~\ref{fig:cross-pollination_single} and Appendix~\ref{sec:additional_results_decoupled} show that overall accuracy remains
largely unaffected under this decoupling, with only minor degradation even when $M_{\text{eval}}$ is a significantly smaller model than $M_{\text{gen}}$. For instance, accuracy remains similar when $M_{\text{gen}}$ is generated by Llama-3.1-70B, while $M_{\text{eval}}$ is estimated with 
other smaller models.
This suggests that the answer confidence term, $\log p(y \mid r, x)$, is not merely a model-specific artefact but functions as a more portable measure of the logical entailment between a given reasoning chain and its conclusion, enabling flexible and computationally efficient answer confidence prediction.

\label{sec:section1}

\section{Related Work}

\textbf{LLM Reasoning} LLM reasoning abilities has gained significantly on complex tasks~\citep{li2025system, muennighoff2025s1}. While CoT reasoning improves performance~\citep{wei2022chain, leang2024comat}, subsequent work introduced hierarchical reasoning phases: multi-path exploration~\citep{yao2023tree, guan2025rstar}, step verification~\citep{lightman2023let, leang2025theorem}, iterative refinement~\citep{madaan2023self}, and analysis and implementation of repeated inference~\citep{levi2025a}. These techniques are computationally prohibitive for LRMs~\citep{team2025kimi, yang2025qwen3}, which produce long, unstructured outputs.

\paragraph{BoN.} BoN is an alignment-via-inference method that optimises outputs with a scoring function~\citep{charniak2005coarse, stiennon2020learning, amini2024variational}. With scale-time inference, LLMs benefit from generating multiple samples and selecting the best via reward models~\citep{snell2024scaling}. Due to their training cost, reward models are often replaced by training-free methods such as Self-Consistency and its variants~\citep{wan2024reasoning, lyu2025calibrating}. 


\paragraph{Sampling and Reranking.}
Reranking improves generation quality~\citep{adiwardana2020towards,shen2021generate}, often via trained verifiers to re-rank candidates, outperforming fine-tuning~\citep{cobbe2021training,guan2025rstar}. Confidence estimation for re-ranking has been explored via sample agreement~\citep{kuhn2023semantic,manakul2023selfcheckgpt, tian2023fine,simhi2025hack}, or prompting models to verbalise confidence~\citep{tian2023just,kadavath2022language}.
\section{Conclusion}
We introduced \methodabbrev, a sample-efficient, training-free scoring function for BoN sampling that selects a reasoning chain by maximising a score decomposed into reasoning and answer confidence. \methodabbrev yields consistent gains across models and datasets, 
narrowing the gap to oracle performance while requiring only $k=6$ samples to outperform baselines using $k=32$. The answer confidence component can be estimated by different models than the one used for generation, enabling flexible and efficient deployment. At the trajectory level, peak-count-to-sentence ratios correlate with accuracy, showing that reasoning chains leading to correct answers are more information-dense. 
Overall, \methodabbrev offers a promising probabilistic confidence route to reasoning selection.

\section*{Limitations}
\methodabbrev{} targets domains with well-defined reasoning structures and definitive answers, such as \emph{mathematical and scientific problem-solving}. We view this scope as both deliberate and essential: these domains represent a substantial class of high-value reasoning tasks where precision is important. Furthermore, restricting our analysis to these settings enables a rigorous evaluation of confidence calibration, a task that remains difficult in open-ended domains -- could be characterised by ambiguity and multiple valid solutions. This controlled environment allows us to validate the efficacy of model confidence as a selection metric without the confounding factors of subjective evaluation.

Extending \methodabbrev to open-ended generation remains an important avenue for future research. To address the lack of definitive answer boundaries in such tasks, a promising direction is to augment the probabilistic framework with learned reward models for answer evaluation. We believe this adaptation could extend the reliability benefits of \methodabbrev beyond fixed-format problems, offering a pathway toward robust reasoning in broader, general-purpose applications.

\section*{Acknowledgements}
We thank the anonymous reviewers and area chairs for their helpful comments and feedback. We also thank Waylon Li and Adi Simhi for their valuable feedback. Lastly,  
we are grateful for the compute resources provided to us by the University of
Edinburgh (Edinburgh International Data Facility), and UKRI
(Isambard AI service, University of Bristol).

\bibliography{custom}

\clearpage
\appendix
\section{Additional Results for Decoupled Confidence Estimation}
\begin{figure*}[!t]
  \centering
  \begin{subfigure}[t]{0.95\textwidth}
    \centering
    \includegraphics[width=0.95\textwidth]{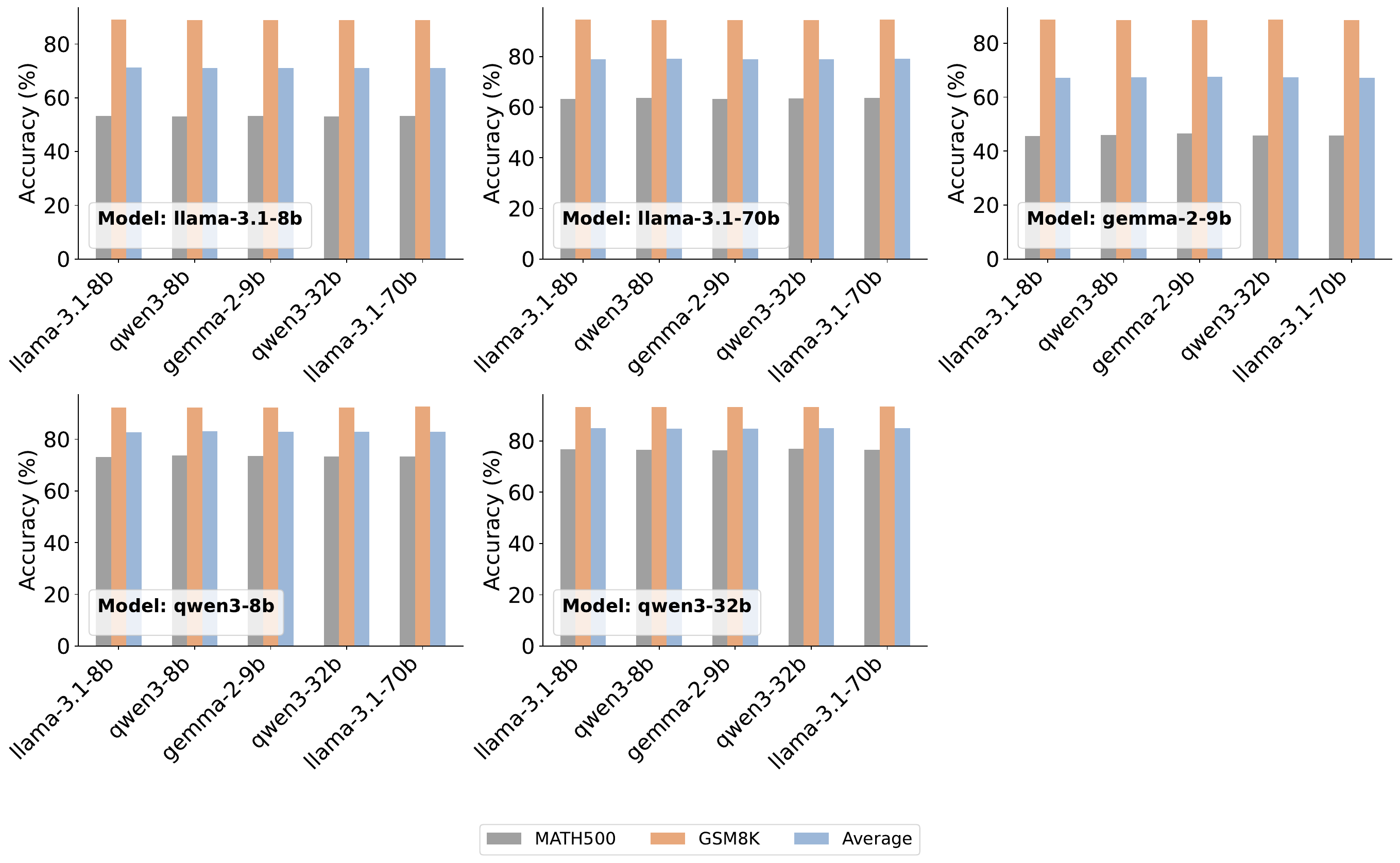}
    \caption{Decoupling plot by using various LLMs to evaluate $p(y \mid r,x)$ across a particular model reasoning chain, $p(r \mid x)$. Each subplot represents a $M_{gen}$, and the \emph{x-axis} represents various $M_{eval}$. The results remain similar when $M_{eval}$ varies, even with smaller models predicting larger $M_{gen}$.}
    \label{fig:cross_pollination}

  \end{subfigure}
  
  \vspace{1em}
  
  \begin{subfigure}[t]{0.95\textwidth}
    \centering
    \includegraphics[width=0.9\textwidth]{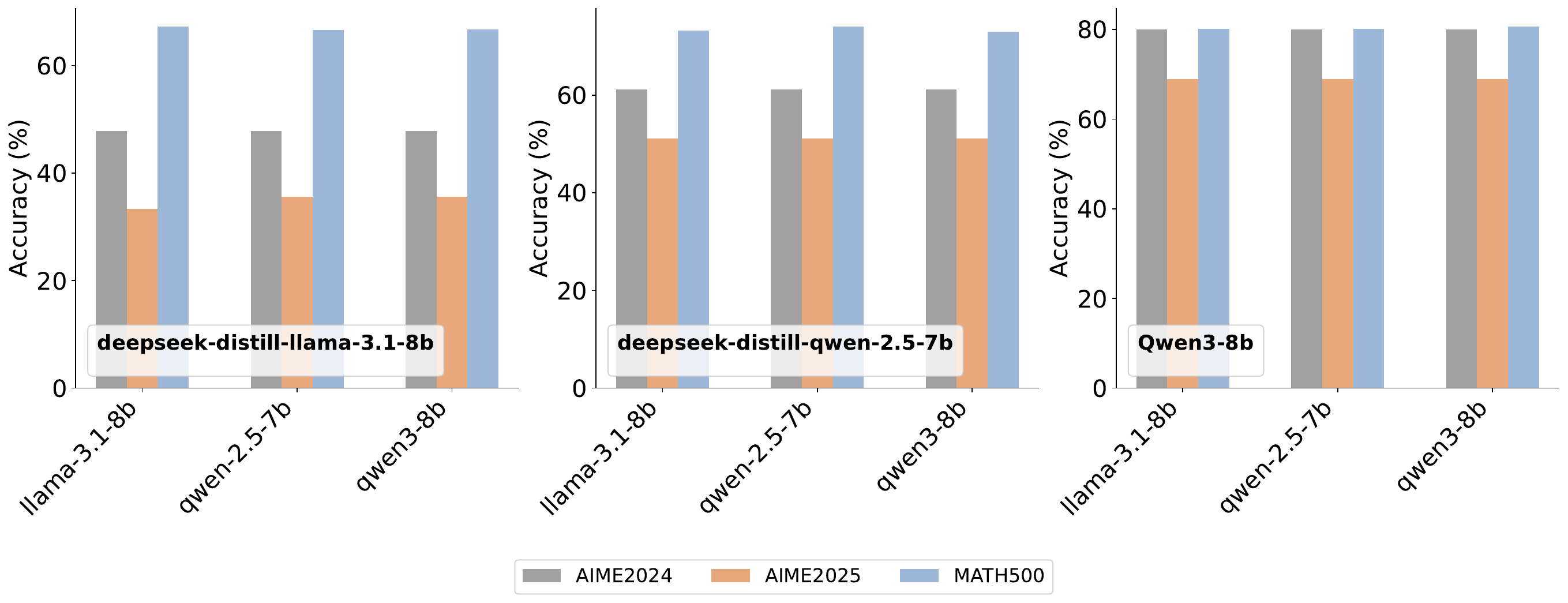}
    \caption{Decoupling plot by using various LRMs to evaluate $p(y \mid r,x)$ across a particular model reasoning chain, $p(r \mid x)$. Each subplot represents a $M_{gen}$, and the \emph{x-axis} represents various $M_{eval}$. The results remain similar when $M_{eval}$ varies, even with smaller models predicting larger $M_{gen}$.}
    \label{fig:cross_pollination_LRM}

  \end{subfigure}
  \caption{Decoupling analysis for LLMs and LRMs (Thinking Enabled) as $M_{\text{gen}}$, with various $M_{\text{eval}}$, showing performance remains similar when different models are used to estimate $\log p(y \mid r, x)$.}
\end{figure*}

\label{sec:additional_results_decoupled}
In this section, we provide supplementary evidence that the decoupled confidence estimation experiments introduced in~\S\ref{sec:decoupled_confidence} are portable across distinct evaluator models. This analysis aims to strengthen the claim that the answer-confidence term, $\log p(y \mid r, x)$, does not depend on the specific evaluator used. 



Based on Figure~\ref{fig:cross_pollination}, switching the evaluator model, $M_{eval}$ while holding the reasoning distribution fixed yields a similar accuracy across datasets. This observation shows that the answer-confidence term, $\log p(y \mid r, x)$, is highly portable, allowing small-scale LLMs to reliably evaluate the reasoning chains of larger models.

When examining LRMs, we observe the same qualitative pattern (shown in Figure~\ref{fig:cross_pollination_LRM}), indicating that the phenomenon generalises across models. This reinforces the hypothesis that decoupled confidence estimation captures a stable property of the reasoning process itself, rather than an artefact of the evaluator model. 

\section{Additional Implementation Details}
\label{appendix:implementation}

\paragraph{Sampling and Decoding.}
For sampling-based methods, we use $k \in \{6, 32\}$ reasoning traces for smaller models and $k \in \{6, 16\}$ for the larger Llama-3.1-70B and Qwen3-32B models, due to computational constraints. For all the models, we apply a hyperparameter of temperature $=0.7$ and top-p $=0.6$. The greedy decoding (temperature $=0$, top-p $=1.0$) baseline corresponds to $k=1$, for which we report $\mathrm{Pass}@1$ accuracy. For specialised LRMs, we use $k=6$ uniformly across all methods due to computational constraints. Since LRMs are not typically evaluated using greedy decoding, we follow the approach of~\citet{yang2025qwen3}, which is a temperature of 0.6, top-k of 20 and top-p $=0.95$, reporting the average accuracy across $k$ samples. For all our baselines except greedy decoding, we evaluate three times with the standard error reported. For LLMs, we cap the maximum token budget at 8,096 tokens. For LRMs, we follow the configuration of \citet{yang2025qwen3}, using a maximum output length of 32,768 tokens, except for AIME’24 and AIME’25, where we extend the budget to 38,912 tokens to ensure sufficient reasoning space.

\paragraph{Baselines and Hyperparameters.}

We compare \methodabbrev against a range of decoding, confidence and re-ranking baselines.  

\begin{itemize}[noitemsep,parsep=0pt,leftmargin=*]
    \item \textbf{Greedy Decoding}  
    As a deterministic decoding strategy, greedy decoding selects at each step the token with the highest conditional probability. Unlike greedy decoding, which selects a single high-probability continuation, PiCSAR evaluates multiple full reasoning trajectories and ranks them using joint reasoning-and-answer log-likelihood, enabling selection of the most globally probable chain.
    \item \textbf{Self-Consistency} (SC; \citealt{wang2022self}).  
    This method samples $k$ reasoning chains and aggregates predictions via majority voting on the final answer. In cases where multiple answers receive equal support, we break ties by selecting one at random. While SC relies purely on majority voting over final answers, PiCSAR incorporates the full reasoning chain’s token-level likelihood along with answer confidence, allowing it to prefer coherent but minority reasoning paths that SC would discard.
    \item \textbf{Universal Self-Consistency} (USC; \citealt{chen2023universal}).  
    We include USC only for LLMs under $k{=}6$ sampling, as prompt and context length restrictions prevent its application in the LRM setting. We use the prompting strategy proposed in \citet{chen2023universal}.  Unlike USC, which asks the model to internally judge “consistency” among samples, PiCSAR uses a probabilistic, model-agnostic scoring function based directly on log-likelihoods of reasoning and answers, avoiding USC’s reliance on model self-evaluation and context-window limits.
    \item \textbf{Self-Certainty}~\citep{kang2025scalable}.
    This method applies KL-divergence-based confidence scores, aggregated via Borda voting with parameter $p{=}0.5$. It provides a probabilistic variant of self-consistency, where each candidate’s confidence distribution informs the re-ranking process. Instead of re-ranking chains with KL-based self-estimated correctness like Self-Certainty, PiCSAR scores each candidate through the true generative probabilities of its entire reasoning path and answer
    \item \textbf{\emph{P(True)}}~\citep{kadavath2022language}. This method prompts the model to evaluate whether the answer or reasoning is \emph{True} or \emph{False}, then parses the probability of the response. While P(True) extracts a scalar correctness probability from a meta-prompt, PiCSAR leverages the actual likelihood structure of the model’s forward pass, combining reasoning and answer probabilities without relying on verbalized or poorly calibrated self-judgments.
    \item \textbf{CISC}~\citep{taubenfeld2025confidence}. This method 
    aggregates multiple sampled reasoning paths by weighting each path’s vote with the model’s own estimated correctness. For a fair comparison, we compare CISC with \methodabbrev as estimated correctness, termed CISC (\methodabbrev), with CISC (\emph{P(True)}, which originally proposed, in Appendix~\ref{sec:cisc-comparison}.
\end{itemize}

We have summarised the novelty of PiCSAR against other baselines in Table~\ref{tab:method_comparison}.

\begin{table*}[t]
\begin{tabular}{lcccc}
\hline
\textbf{Method} & \textbf{SC} & \textbf{USC} & \textbf{Self-Cert.} & \textbf{PiCSAR} \\
\hline
Full Reasoning Chain & \checkmark & \checkmark & & \checkmark \\
Model Confidence & & \checkmark & \checkmark & \checkmark \\
Computationally Efficient & \checkmark & \texttimes$^*$ & \checkmark & \checkmark \\
Smaller Model Capable & \checkmark & & \checkmark & \checkmark \\
\hline
\multicolumn{5}{l}{\small $^*$Due to context length} \\
\end{tabular}
\centering
\caption{Comparison of different baselines with our proposed method across several dimensions. }
\label{tab:method_comparison}
\vspace{-2ex}
\end{table*}

\paragraph{Baseline Restrictions.} Due to context length constraints, USC can only handle a limited number of samples and is therefore evaluated exclusively in the LLM setting with $k{=}6$, and excluded from all LRM experiments.




\paragraph{Ablations.}  
To disentangle the contributions of the two terms in our joint objective, we introduce single-term ablations. \emph{Reasoning Confidence} ranks candidates solely by $\log p(r \mid x)$, favouring plausible reasoning traces. \emph{Answer Confidence} instead ranks by $\log p(y \mid r, x)$, prioritising certainty in the final answer given the reasoning path.  

\paragraph{Framework and Hardware.}
All experiments are conducted using the vLLM framework~\citep{kwon2023efficient}.
All experiments are conducted on 2–4 NVIDIA H100 GPUs (80GB). Results are reported as averages over independent evaluation runs to ensure robustness.  

\paragraph{Prompt.} 

For the reasoning confidence $\log p(r \mid x)$ generation, we utilise the following prompt:

\begin{tcolorbox}[colback=gray!10,colframe=gray!50,boxrule=0.5pt]
\small

\texttt{You are a helpful AI Assistant that provides well-reasoned and detailed responses. Think step by step and provide the final answer in the form of `The final answer is: [answer]'. Decompose and break down your reasoning into smallest possible steps (Do not combine multiple inferences in one step), and do label your steps very clearly with `Step 1... \textbackslash n\textbackslash n Step 2... \textbackslash n\textbackslash n Step 3.... \textbackslash n\textbackslash n..... \textbackslash n\textbackslash n Step N-1..... \textbackslash n\textbackslash n Step N \textbackslash n\textbackslash n The final answer is: [answer]'.}
\end{tcolorbox}

For predicting answer confidence $\log p(y \mid r,x)$, we follow a similar method to~\citep{ton2024understanding} but without training. Specifically, we use the prompt template $\langle a\rangle$ with 5-shot learning:

\begin{tcolorbox}[colback=gray!10,colframe=gray!50,boxrule=0.5pt]
\small
\texttt{You are a helpful assistant. When you see a potential partial reasoning followed by `<sep>', output the final answer.}

\end{tcolorbox}
\subsection{Analyis of Prompts}
To verify that the observed improvements are not attributable to the explicit instruction prompt (see \eqref{eq:modified_objective}), we evaluated several alternative prompt formulations on the Llama-3.1-8B model. Using the \textsc{MATH500} benchmark, we compared the resulting answer-confidence estimates across prompts.

\begin{tcolorbox}[colback=gray!10,colframe=gray!50,boxrule=0.5pt]
\small
\texttt{Prompt 1: "You are a helpful assistant. When you see a potential partial reasoning followed by '<sep>', output the final answer. Here are some examples" + system\_contents + "You are not allowed to provide any redundant symbols at for the final answer, including '\#', '/', '\$', '**' or others. Please only provide numbers as the final answer."}
\end{tcolorbox}

\begin{tcolorbox}[colback=gray!10,colframe=gray!50,boxrule=0.5pt]
\small
\texttt{Prompt 2 (original prompt): "You are a helpful assistant. When you see a potential partial reasoning followed by '<sep>', output the final answer. Here are some examples"}
\end{tcolorbox}

\begin{tcolorbox}[colback=gray!10,colframe=gray!50,boxrule=0.5pt]
\small
\texttt{Prompt 3: "You are a helpful assistant. By providing the partial reasoning, output the final answer directly without any additional texts."}
\end{tcolorbox}

\begin{tcolorbox}[colback=gray!10,colframe=gray!50,boxrule=0.5pt]
\small
\texttt{Prompt 4: "You are a helpful assistant. Based on the reasoning provided, output the final answer directly without any additional texts. Only Provide the final answer."}
\end{tcolorbox}

\begin{tcolorbox}[colback=gray!10,colframe=gray!50,boxrule=0.5pt]
\small
\texttt{Prompt 5: "You are a helpful assistant. Provide the final answer directly without any additional texts (only the final answer) based on the partial reasoning."}
\end{tcolorbox}

\begin{table}[h!]
\centering
\begin{tabular}{l c}
\hline
\textbf{Prompt} & \textbf{Accuracy} \\
\hline
Prompt 1 & 54.60\% \\
Prompt 2 & 54.00\% \\
Prompt 3 & 54.20\% \\
Prompt 4 & 54.40\% \\
Prompt 5 & 54.40\% \\
\hline
\end{tabular}
\caption{Performance of PiCSAR on Llama-3.1-8B on MATH500 with different prompts for answer-confidence extraction.}\label{tab:prompt_performance}
\vspace{-2ex}
\end{table}

\begin{table*}[ht]
  \centering
  \small
  \setlength{\tabcolsep}{5pt}      
  \resizebox{\textwidth}{!}{
  \begin{tabular}{l c c c c c c c c}
    \toprule
    \textbf{Method}
      & \multicolumn{2}{c}{\textbf{SVAMP}}
      & \multicolumn{2}{c}{\textbf{GSM8K}}
      & \multicolumn{2}{c}{\textbf{MATH500}}
      & \multicolumn{2}{c}{\textbf{TheoremQA}}\\
    & $k=6$ & $k=16/32$
    & $k=6$ & $k=16/32$
    & $k=6$ & $k=16/32$ 
    & $k=6$ & $k=16/32$ \\
    \midrule
    \multicolumn{9}{c}{\emph{Gemma-2-9B-Instruct}} \\
    \midrule
    CISC (\emph{p(True)} & $89.22{\scriptstyle\pm 0.22}$ & $88.67{\scriptstyle\pm 0.38}$ & $88.89{\scriptstyle\pm 0.26}$ & $89.14{\scriptstyle\pm 0.15}$ & $46.87{\scriptstyle\pm 0.33}$ & $47.67{\scriptstyle\pm 0.07}$ & $17.09{\scriptstyle\pm 0.43}$ & $17.45{\scriptstyle\pm 0.12}$ \\
    \rowcolor{blue!10} \methodabbrev &$ 89.00{\scriptstyle\pm 0.38}$ & $91.02{\scriptstyle\pm 0.59}$ & $ 88.66$${\scriptstyle\pm 0.11}$ & $ 88.99$${\scriptstyle\pm 0.20}$ & $ 46.53$${\scriptstyle\pm 0.29}$ & $47.13$${\scriptstyle\pm 0.13}$ & $ 18.62$${\scriptstyle\pm 0.39}$ & $ 18.88$${\scriptstyle\pm 0.54}$ \\
    \rowcolor{blue!10} CISC (\methodabbrev) & $\bf 91.89{\scriptstyle\pm 0.22}$ & $\bf 92.33{\scriptstyle\pm 0.19}$ & $\bf 91.85{\scriptstyle\pm 0.20}$ & $\bf 92.43{\scriptstyle\pm 0.22}$ & $\bf 51.33{\scriptstyle\pm 0.07}$ & $\bf 52.13{\scriptstyle\pm 0.29}$ & $\bf 21.02{\scriptstyle\pm 0.58}$ & $\bf 23.16{\scriptstyle\pm 0.39}$ \\
    Upper Bound & $24.32{\scriptstyle\pm 0.49}$ & $32.40{\scriptstyle\pm 0.20}$ & $93.44{\scriptstyle\pm 0.09}$ & $95.60{\scriptstyle\pm 0.04}$ & $58.47{\scriptstyle\pm 0.27}$ & $66.67{\scriptstyle\pm 0.47}$ & $55.22{\scriptstyle\pm 1.10}$ & $82.49{\scriptstyle\pm 1.02}$ \\
    \midrule
    \multicolumn{9}{c}{\emph{Llama-3.1-8B-Instruct}} \\
    \midrule
    CISC (\emph{p(True)}) & $91.44{\scriptstyle\pm 0.48}$ & $92.78{\scriptstyle\pm 0.29}$ & $91.17{\scriptstyle\pm 0.18}$ & $91.91{\scriptstyle\pm 0.49}$ & $54.93{\scriptstyle\pm 0.41}$ & $58.20{\scriptstyle\pm 0.42}$ & $18.03{\scriptstyle\pm 0.73}$ & $39.38{\scriptstyle\pm 18.91}$ \\
    \rowcolor{blue!10} \methodabbrev & $91.78$${\scriptstyle\pm 0.11}$ & $93.44$${\scriptstyle\pm 0.89}$ & $89.09$${\scriptstyle\pm 0.13}$ & $89.98$${\scriptstyle\pm 0.23}$ & $53.33$${\scriptstyle\pm 0.73}$ & $53.87$${\scriptstyle\pm 0.70}$ & $ 20.08$${\scriptstyle\pm 0.43}$ & $ 19.72$${\scriptstyle\pm 0.39}$ \\
    \rowcolor{blue!10} CISC (\methodabbrev) & $\textbf{94.33}{\scriptstyle\pm 0.33}$ & $\textbf{96.22}{\scriptstyle\pm 0.11}$ & $\textbf{93.98}{\scriptstyle\pm 0.14}$ & $\textbf{94.23}{\scriptstyle\pm 0.08}$ & $\textbf{62.47}{\scriptstyle\pm 0.07}$ &  $\textbf{62.40}{\scriptstyle\pm 0.50}$ & $\textbf{22.71}{\scriptstyle\pm 0.25}$ & $\textbf{41.50}{\scriptstyle\pm 17.34}$ \\
    Upper Bound                        & $  96.78$${\scriptstyle\pm 0.11}$ & $ 99.11$${\scriptstyle\pm 0.11}$ & $96.15{\scriptstyle\pm 0.07}$ & $ 98.18$${\scriptstyle\pm 0.04}$ & $72.80{\scriptstyle\pm 0.23}$ & $ 82.20$${\scriptstyle\pm 0.60}$ & $28.20{\scriptstyle\pm 0.32}$ & $37.846{\scriptstyle\pm 1.13}$ \\
    \midrule
    \multicolumn{9}{c}{\emph{Qwen3-8B (Non-thinking)}} \\
    \midrule
    CICS (\emph{p(True)}) & $94.33{\scriptstyle\pm 0.00}$ & $94.56{\scriptstyle\pm 0.11}$ & $93.80{\scriptstyle\pm 0.13}$ & $94.05{\scriptstyle\pm 0.14}$ & $77.20{\scriptstyle\pm 0.20}$ & $77.93{\scriptstyle\pm 0.24}$ & $31.24{\scriptstyle\pm 0.04}$ & $32.75{\scriptstyle\pm 0.45}$ \\
    \rowcolor{blue!10} \methodabbrev & $ 93.56$${\scriptstyle\pm 0.22}$ & $95.13$${\scriptstyle\pm 0.22}$ & $92.33$${\scriptstyle\pm 0.13}$ & $ 93.22$${\scriptstyle\pm 0.08}$ & $ 73.67$${\scriptstyle\pm 0.24}$ & $73.40$${\scriptstyle\pm 0.13}$ & $29.76{\scriptstyle\pm 0.57}$ & $ 29.17$${\scriptstyle\pm 0.64}$ \\
     \rowcolor{blue!10} CICS (\methodabbrev)  & $\bf 95.11{\scriptstyle\pm 0.11}$ & $\bf 95.67{\scriptstyle\pm 0.19}$ & $\bf 94.89{\scriptstyle\pm 0.14}$ & $\bf 95.22{\scriptstyle\pm 0.12}$ & $\bf 79.80{\scriptstyle\pm 0.40}$ & $\bf 79.60{\scriptstyle\pm 0.42}$ & $\bf 36.46{\scriptstyle\pm 0.04}$ & $\bf 36.32{\scriptstyle\pm 0.04}$ \\
    Upper Bound & $96.33{\scriptstyle\pm 0.67}$ & $97.89{\scriptstyle\pm 0.11}$ & $95.52{\scriptstyle\pm 0.00}$ & $96.84{\scriptstyle\pm 0.03}$ & $81.13{\scriptstyle\pm 0.44}$ & $83.53{\scriptstyle\pm 0.24}$ & $34.94{\scriptstyle\pm 0.00}$ & $40.03{\scriptstyle\pm 0.35}$ \\
    \midrule
    \multicolumn{9}{c}{\emph{Llama-3.1-70B-Instruct}} \\
    \midrule
    CISC (\emph{p(True)}) & $94.22{\scriptstyle\pm 0.22}$ & $94.11{\scriptstyle\pm 0.11}$ & $94.68{\scriptstyle\pm 0.00}$ & $95.09{\scriptstyle\pm 0.09}$ & $65.07{\scriptstyle\pm 1.05}$ & $66.27{\scriptstyle\pm 0.29}$ & $28.07{\scriptstyle\pm 0.68}$ & $29.41{\scriptstyle\pm 0.12}$ \\
    \rowcolor{blue!10} \methodabbrev & $94.10{\scriptstyle\pm 0.11}$ & $95.58$${\scriptstyle\pm 0.22}$ & $94.58$${\scriptstyle\pm 0.03}$ & $94.81$${\scriptstyle\pm 0.13}$ & $ 63.67$${\scriptstyle\pm 1.51}$ & $ 64.07$${\scriptstyle\pm 0.87}$ & $ 27.84$${\scriptstyle\pm 0.19}$ & $26.73$${\scriptstyle\pm 0.27}$ \\
     \rowcolor{blue!10} CISC (PiCSAR) & $\textbf{96.78}{\scriptstyle\pm 0.11}$ & $\textbf{96.44}{\scriptstyle\pm 0.11}$ & $\textbf{95.90}{\scriptstyle\pm 0.08}$ & $\textbf{96.03}{\scriptstyle\pm 0.11}$ & $\textbf{69.60}{\scriptstyle\pm 0.31}$ & $\textbf{70.80}{\scriptstyle\pm 0.76}$ & $\textbf{31.91}{\scriptstyle\pm 0.31}$ & $\textbf{31.59}{\scriptstyle\pm 0.27}$ \\
    Upper Bound & $97.22{\scriptstyle\pm 0.22}$ & $97.78{\scriptstyle\pm 0.22}$ & $96.91{\scriptstyle\pm 0.03}$ & $97.44{\scriptstyle\pm 0.03}$ & $77.07{\scriptstyle\pm 0.47}$ & $81.67{\scriptstyle\pm 0.18}$ & $40.70{\scriptstyle\pm 0.20}$ & $43.47{\scriptstyle\pm 0.18}$ \\
    \midrule
    \multicolumn{9}{c}{\emph{Qwen3-32B (Non-thinking)}} \\
    \midrule
    CICS (P-True) & $94.33{\scriptstyle\pm 0.00}$ & $94.56{\scriptstyle\pm 0.11}$ & $93.80{\scriptstyle\pm 0.13}$ & $94.05{\scriptstyle\pm 0.14}$ & $77.20{\scriptstyle\pm 0.20}$ & $77.93{\scriptstyle\pm 0.24}$ & $31.24{\scriptstyle\pm 0.04}$ & $32.75{\scriptstyle\pm 0.45}$ \\
    \rowcolor{blue!10} \methodabbrev & $ 93.22$${\scriptstyle\pm 0.22}$ & $ 93.55$${\scriptstyle\pm 0.33}$ & $ 93.90$${\scriptstyle\pm 0.28}$ & $ 93.88$${\scriptstyle\pm 0.22}$ & $77.00$${\scriptstyle\pm 0.18}$ & $ 75.93{\scriptstyle\pm 0.13}$ & $31.46$${\scriptstyle\pm 0.04}$ & $31.42$${\scriptstyle\pm 0.27}$ \\
     \rowcolor{blue!10} CICS (\methodabbrev) & $\textbf{95.11}{\scriptstyle\pm 0.11}$ & $\textbf{95.67}{\scriptstyle\pm 0.19}$ & $\textbf{94.89}{\scriptstyle\pm 0.14}$ & $\textbf{95.22}{\scriptstyle\pm 0.12}$ & $\textbf{79.80}{\scriptstyle\pm 0.40}$ & $\textbf{79.60}{\scriptstyle\pm 0.42}$ & $\textbf{36.46}{\scriptstyle\pm 0.04}$ & $\textbf{36.32}{\scriptstyle\pm 0.04}$ \\

    Upper Bound & $96.78{\scriptstyle\pm 0.11}$ & $98.00{\scriptstyle\pm 0.00}$ & $96.28{\scriptstyle\pm 0.13}$ & $96.99{\scriptstyle\pm 0.07}$ & $82.27{\scriptstyle\pm 0.13}$ & $83.73{\scriptstyle\pm 0.07}$ & $39.76{\scriptstyle\pm 0.00}$ & $42.93{\scriptstyle\pm 0.12}$ \\
    \bottomrule
  \end{tabular}
  }
  \caption{
  \textbf{Performance comparison on benchmarks across CISC (\emph{p(True)}) and CISC (PiCSAR) on LLMs.}
  Values represent mean accuracy ± standard error over three independent evaluation runs.
  \textbf{Bold} indicates the best-performing method per column based on the mean accuracy. 
  %
  Sampling parameters: $k=\{6,32\}$ for Gemma-2-9B, Llama-3.1-8B, and Qwen3-8B; $k=\{6,16\}$ for Llama-3.1-70B and Qwen3-32B.
  }
  \label{tab:cisc-comparison}
  
\end{table*}

Our results in Table~\ref{tab:prompt_performance} show that changes in prompt phrasing have minimal influence on model performance. This suggests that, although the instructional content of a prompt remains essential for eliciting the final answer, the precise wording plays only a limited role in shaping the model’s behaviour.


\section{Further Experimental Results and Ablation Studies}
\subsection{Comparison between CISC (p(True)) and CISC (PiCSAR)} 
\label{sec:cisc-comparison}
Based on~\cref{tab:cisc-comparison}, \methodabbrev shows a great performance when integrated with weightage voting on CISC~\citep{taubenfeld2025confidence}, consistently improving baseline CICS (\emph{p(True)}) metrics across all evaluated methods. This indicates that \methodabbrev functions effectively both as a standalone selection mechanism and as an augmentation to existing weighting schemes. While these findings suggest promising direction for performance optimisation, this lies beyond the current research scope.

\subsection{Component Analysis and Main Results Breakdown}
\label{sec:breakdown}
\label{sec:appendix_results}
In this section, we first provide a detailed breakdown of the experimental results for all methods, as summarised in Table~\ref{tab:model-comparison_breakdown}, where we also show the performance of \methodabbrev-N, a length-normalised variant of our primary method. Finally, we present ablation studies on LRMs in Table~\ref{tab:deepseek_qwen_ablation}.
We compare three primary approaches: \emph{Reasoning Confidence} ($\log p(r \mid x)$), \emph{Answer Confidence} ($\log p(y \mid r, x)$), and our main method, \emph{\methodabbrev} (the joint probability). 

Across the majority of benchmarks and model families presented in Table~\ref{tab:model-comparison_breakdown}, we generally observe that \methodabbrev outperforms its individual components. This pattern underscores the benefit of jointly considering the likelihood of both the reasoning process and the final answer. However, there are specific instances where relying solely on answer confidence, $\log p(y \mid r, x)$, achieves comparable or slightly better results (e.g., Gemma-2-9B and Qwen3-32B on GPQA-Diamond for $k=32$), highlighting that answer confidence remains a strong and competitive signal on its own.

\subsection{Length-Normalised Variant: \methodabbrev-N}
\label{app:picsar-n}

As introduced in the main paper, we proposed a variant of our method, \methodabbrev-N, which applies length normalisation to the reasoning confidence term. The scoring function for \methodabbrev-N is defined as:

\begin{equation} \label{eq:modified_objective_norm_appendix}
\begin{split}
\text{Score}(r, y) = &\left[ \displaystyle\frac{1}{N}\,\log p(r \mid x) \right] \\
&+ \log p(y \mid \langle a \rangle, r, x),
\end{split}
\end{equation}

%
where $N$ is the number of tokens in the reasoning chain $r$. This normalisation is intended to mitigate any potential length bias, which might unfairly penalise longer reasoning paths.

\begin{table*}[t]
  \centering
  \small
  \setlength{\tabcolsep}{5pt}      
  \resizebox{\textwidth}{!}{
  \begin{tabular}{l c c c c c c c c}
    \toprule
    \textbf{Method}
      & \multicolumn{2}{c}{\textbf{SVAMP}}
      & \multicolumn{2}{c}{\textbf{GSM8K}}
      & \multicolumn{2}{c}{\textbf{MATH500}}
      & \multicolumn{2}{c}{\textbf{GPQA-Diamond}}\\
    & $k=6$ & $k=16/32$
    & $k=6$ & $k=16/32$
    & $k=6$ & $k=16/32$ 
    & $k=6$ & $k=16/32$ \\
    \midrule
    \multicolumn{9}{c}{\emph{Gemma-2-9B-Instruct}} \\
    \midrule
    Reasoning Confidence              & $88.66{\scriptstyle\pm 0.33}$ & $89.67{\scriptstyle\pm 0.49}$ & $88.51{\scriptstyle\pm 0.05}$ & $88.46{\scriptstyle\pm 0.25}$ & $45.87{\scriptstyle\pm 0.47}$ & $45.87{\scriptstyle\pm 0.68}$ & $30.64{\scriptstyle\pm 0.45}$ & $32.32{\scriptstyle\pm 1.52}$ \\
    Answer Confidence                 & $89.66$${\scriptstyle\pm 0.33}$ & $89.02{\scriptstyle\pm 0.59}$ & $88.05{\scriptstyle\pm 0.17}$ & $87.04{\scriptstyle\pm 0.05}$ & $46.47{\scriptstyle\pm 0.66}$ & $46.33{\scriptstyle\pm 0.18}$ & $ 34.01$${\scriptstyle\pm 2.65}$ & $\bf 38.22$${\scriptstyle\pm 1.76}$ \\
    Reasoning confidence (normalised) & $89.56{\scriptstyle\pm 0.44}$ & $90.22{\scriptstyle\pm 0.29}$ & $88.76$${\scriptstyle\pm 0.26}$ & $\bf 89.45$${\scriptstyle\pm 0.20}$ & $46.33{\scriptstyle\pm 0.67}$ & $46.47{\scriptstyle\pm 0.18}$ & $29.80{\scriptstyle\pm 1.91}$ & $27.95{\scriptstyle\pm 2.15}$\\
    \rowcolor{blue!10} \methodabbrev &$ 89.00{\scriptstyle\pm 0.38}$ & $\bf 91.02$${\scriptstyle\pm 0.59}$ & $ 88.66$${\scriptstyle\pm 0.11}$ & $ 88.99$${\scriptstyle\pm 0.20}$ & $ 46.53$${\scriptstyle\pm 0.29}$ & $\bf 47.13$${\scriptstyle\pm 0.13}$ & $ 32.32$${\scriptstyle\pm 0.51}$ & $ 34.01$${\scriptstyle\pm 1.94}$ \\
    \rowcolor{blue!10} \methodabbrev-N & $\bf 89.67$${\scriptstyle\pm 0.19}$ & $ 89.22$${\scriptstyle\pm 0.29}$ & $\bf 88.91$${\scriptstyle\pm 0.12}$ & $ 89.27$${\scriptstyle\pm 0.11}$ & $\bf 46.60$${\scriptstyle\pm 0.92}$ & $46.93{\scriptstyle\pm 0.18}$ & $\bf 35.35$${\scriptstyle\pm 1.62}$ & $ 38.05$${\scriptstyle\pm 1.90}$\\
    Upper Bound & $93.44{\scriptstyle\pm 0.22}$ & $95.67{\scriptstyle\pm 0.38}$ & $93.44{\scriptstyle\pm 0.09}$ & $95.60{\scriptstyle\pm 0.04}$ & $58.47{\scriptstyle\pm 0.27}$ & $66.67{\scriptstyle\pm 0.47}$ & $55.22{\scriptstyle\pm 1.10}$ & $82.49{\scriptstyle\pm 1.02}$ \\
    \midrule
    \multicolumn{9}{c}{\emph{Llama-3.1-8B-Instruct}} \\
    \midrule
    Reasoning Confidence               & $91.56{\scriptstyle\pm 0.11}$ & $92.10{\scriptstyle\pm 0.84}$ & $88.89{\scriptstyle\pm 0.09}$ & $89.67{\scriptstyle\pm 0.27}$ & $53.07{\scriptstyle\pm 0.37}$ & $51.53{\scriptstyle\pm 0.35}$ & $29.12{\scriptstyle\pm 1.02}$ & $32.49{\scriptstyle\pm 2.92}$ \\
    Answer Confidence                  & $89.11{\scriptstyle\pm 0.29}$ & $90.44{\scriptstyle\pm 0.95}$ & $86.84{\scriptstyle\pm 0.20}$ & $86.69{\scriptstyle\pm 0.04}$ & $49.27{\scriptstyle\pm 0.64}$ & $50.20{\scriptstyle\pm 0.35}$ & $28.62{\scriptstyle\pm 0.73}$ & $29.46{\scriptstyle\pm 2.63}$ \\
    Reasoning confidence  (normalised) & $90.22{\scriptstyle\pm 0.11}$ & $90.67{\scriptstyle\pm 0.69}$ & $88.38{\scriptstyle\pm 0.23}$ & $86.10{\scriptstyle\pm 0.08}$ & $50.67{\scriptstyle\pm 0.47}$ & $47.13{\scriptstyle\pm 1.39}$ & $22.05{\scriptstyle\pm 0.89}$ & $18.35{\scriptstyle\pm 0.84}$\\
    \rowcolor{blue!10} \methodabbrev & $\bf 91.78$${\scriptstyle\pm 0.11}$ & $\bf 93.44$${\scriptstyle\pm 0.89}$ & $\bf 89.09$${\scriptstyle\pm 0.13}$ & $\bf 89.98$${\scriptstyle\pm 0.23}$ & $\bf 53.33$${\scriptstyle\pm 0.73}$ & $\bf 53.87$${\scriptstyle\pm 0.70}$ & $ 29.80$${\scriptstyle\pm 1.34}$ & $\bf  33.67$${\scriptstyle\pm 3.06}$ \\
    \rowcolor{blue!10} \methodabbrev-N & $90.22{\scriptstyle\pm 0.48}$ & $92.22{\scriptstyle\pm 0.29}$ & $88.59{\scriptstyle\pm 0.18}$ & $89.33{\scriptstyle\pm 0.42}$ & $51.53{\scriptstyle\pm 0.48}$ & $51.60{\scriptstyle\pm 0.42}$ & $\bf 30.81{\scriptstyle\pm 0.87}$ & 
    $ 30.64{\scriptstyle\pm 1.61}$\\
    Upper Bound                        & $  96.78$${\scriptstyle\pm 0.11}$ & $ 99.11$${\scriptstyle\pm 0.11}$ & $96.15{\scriptstyle\pm 0.07}$ & $ 98.18$${\scriptstyle\pm 0.04}$ & $72.80{\scriptstyle\pm 0.23}$ & $ 82.20$${\scriptstyle\pm 0.60}$ & $65.82{\scriptstyle\pm 1.50}$ & $92.76{\scriptstyle\pm 0.73}$ \\
    \midrule
    \multicolumn{9}{c}{\emph{Qwen3-8B (Non-thinking)}} \\
    \midrule
    Reasoning Confidence               & $92.78{\scriptstyle\pm 0.11}$ & $94.34{\scriptstyle\pm 0.33}$ & $92.26{\scriptstyle\pm 0.13}$ & $92.31{\scriptstyle\pm 0.03}$ & $73.53{\scriptstyle\pm 0.24}$ & $72.53{\scriptstyle\pm 0.48}$ & $45.96{\scriptstyle\pm 1.01}$ & $43.77{\scriptstyle\pm 1.21}$ \\
    Answer Confidence                   & $93.45{\scriptstyle\pm 0.19}$ & $94.02{\scriptstyle\pm 0.40}$ & $ 93.22$${\scriptstyle\pm 0.03}$ & $92.94{\scriptstyle\pm 0.17}$ & $71.07{\scriptstyle\pm 0.41}$ & $71.20{\scriptstyle\pm 0.76}$ & $\bf 51.01$${\scriptstyle\pm 1.52}$ & $43.43{\scriptstyle\pm 2.53}$ \\
    Reasoning Confidence (normalised) & $93.33{\scriptstyle\pm 0.00}$ & $93.67{\scriptstyle\pm 0.69}$ & $92.79{\scriptstyle\pm 0.00}$ & $92.61{\scriptstyle\pm 0.20}$ & $71.93{\scriptstyle\pm 0.71}$ & $69.27{\scriptstyle\pm 0.44}$ & $43.43{\scriptstyle\pm 0.51}$ & $38.05{\scriptstyle\pm 1.78}$\\
    \rowcolor{blue!10} \methodabbrev & $ 93.56$${\scriptstyle\pm 0.22}$ & $\bf 95.13$${\scriptstyle\pm 0.22}$ & $92.33$${\scriptstyle\pm 0.13}$ & $ 93.22$${\scriptstyle\pm 0.08}$ & $ 73.67$${\scriptstyle\pm 0.24}$ & $\bf  73.40$${\scriptstyle\pm 0.13}$ & $46.98{\scriptstyle\pm 1.01}$ & $ 43.69$${\scriptstyle\pm 1.26}$ \\
    \rowcolor{blue!10} \methodabbrev-N & $\bf 94.44{\scriptstyle\pm 0.11}$ & \bf$94.56$${\scriptstyle\pm 0.59}$ & $\bf 93.69$${\scriptstyle\pm 0.00}$ & $\bf 93.77{\scriptstyle\pm 0.13}$ & $\bf 73.80{\scriptstyle\pm 0.20}$ & $72.13{\scriptstyle\pm 0.98}$ & $47.98{\scriptstyle\pm 1.01}$ & $\bf 44.95{\scriptstyle\pm 0.58}$\\
    Upper Bound & $96.33{\scriptstyle\pm 0.67}$ & $97.89{\scriptstyle\pm 0.11}$ & $95.52{\scriptstyle\pm 0.00}$ & $96.84{\scriptstyle\pm 0.03}$ & $81.13{\scriptstyle\pm 0.44}$ & $83.53{\scriptstyle\pm 0.24}$ & $76.26{\scriptstyle\pm 1.62}$ & $86.36{\scriptstyle\pm 0.29}$ \\
    \midrule
    \multicolumn{9}{c}{\emph{Llama-3.1-70B-Instruct}} \\
    \midrule
    Reasoning Confidence               & $\bf 94.44$${\scriptstyle\pm 0.11}$ & $94.80{\scriptstyle\pm 0.19}$ & $94.46{\scriptstyle\pm 0.08}$ & $93.62{\scriptstyle\pm 0.18}$ & $63.47{\scriptstyle\pm 1.35}$ & $63.00{\scriptstyle\pm 0.10}$ & $43.94{\scriptstyle\pm 2.62}$ & $45.96{\scriptstyle\pm 2.54}$ \\
    Answer Confidence                  & $93.89{\scriptstyle\pm 0.22}$ & $94.67{\scriptstyle\pm 0.38}$ & $94.10{\scriptstyle\pm 0.25}$ & $94.68{\scriptstyle\pm 0.23}$ & $59.40{\scriptstyle\pm 1.30}$ & $60.07{\scriptstyle\pm 1.09}$ & $45.12{\scriptstyle\pm 0.45}$ & $42.26{\scriptstyle\pm 1.78}$ \\
    Reasoning Confidence (normalised) & $93.33{\scriptstyle\pm 0.38}$ & $93.89{\scriptstyle\pm 0.22}$ & $93.37{\scriptstyle\pm 0.03}$ & $93.34{\scriptstyle\pm 0.26}$ & $65.60$${\scriptstyle\pm 0.60}$ & $65.13$${\scriptstyle\pm 0.13}$ & $40.07{\scriptstyle\pm 1.87}$ & $37.04{\scriptstyle\pm 0.89}$\\
    \rowcolor{blue!10} \methodabbrev & $94.10{\scriptstyle\pm 0.11}$ & $\bf 95.58$${\scriptstyle\pm 0.22}$ & $\bf 94.58$${\scriptstyle\pm 0.03}$ & $\bf 94.81$${\scriptstyle\pm 0.13}$ & $ 63.67$${\scriptstyle\pm 1.51}$ & $ 64.07$${\scriptstyle\pm 0.87}$ & $ 46.91$${\scriptstyle\pm 2.65}$ & $\bf 46.46$${\scriptstyle\pm 2.59}$ \\
    \rowcolor{blue!10} \methodabbrev-N & $\bf 94.44{\scriptstyle\pm 0.11}$ & $94.56{\scriptstyle\pm 0.59}$ & $94.07{\scriptstyle\pm 0.00}$ & $94.14{\scriptstyle\pm 0.13}$ & $\bf 72.00$${\scriptstyle\pm 0.20}$ & $\bf 70.33$${\scriptstyle\pm 0.98}$ & $\bf 47.98$${\scriptstyle\pm 1.01}$ & $44.95{\scriptstyle\pm 0.58}$\\
    Upper Bound & $97.22{\scriptstyle\pm 0.22}$ & $97.78{\scriptstyle\pm 0.22}$ & $96.91{\scriptstyle\pm 0.03}$ & $97.44{\scriptstyle\pm 0.03}$ & $77.07{\scriptstyle\pm 0.47}$ & $81.67{\scriptstyle\pm 0.18}$ & $75.59{\scriptstyle\pm 0.61}$ & $87.71{\scriptstyle\pm 0.45}$ \\
    \midrule
    \multicolumn{9}{c}{\emph{Qwen3-32B (Non-thinking)}} \\
    \midrule
    Reasoning confidence               & $92.78{\scriptstyle\pm 0.22}$ & $93.33{\scriptstyle\pm 0.29}$ & $93.19{\scriptstyle\pm 0.28}$ & $94.54{\scriptstyle\pm 0.22}$ & $76.47{\scriptstyle\pm 0.07}$ & $75.87{\scriptstyle\pm 0.18}$ & $44.78{\scriptstyle\pm 0.94}$ & $42.59{\scriptstyle\pm 1.02}$ \\
    Answer confidence                  & $92.56{\scriptstyle\pm 0.11}$ & $92.22{\scriptstyle\pm 0.29}$ & $93.84{\scriptstyle\pm 0.05}$ & $93.42{\scriptstyle\pm 0.13}$ & $75.40{\scriptstyle\pm 0.46}$ & $74.67{\scriptstyle\pm 0.18}$ & $\bf 51.85$${\scriptstyle\pm 0.61}$ & $44.11{\scriptstyle\pm 0.94}$ \\
    Reasoning Confidence (normalised) & $93.33{\scriptstyle\pm 0.19}$ & $94.11{\scriptstyle\pm 0.29}$ & $93.39{\scriptstyle\pm 0.00}$ & $93.44{\scriptstyle\pm 0.30}$ & $75.47{\scriptstyle\pm 0.27}$ & $75.53{\scriptstyle\pm 0.18}$ & $49.33{\scriptstyle\pm 1.18}$ & $37.88{\scriptstyle\pm 1.27}$\\
    \rowcolor{blue!10} \methodabbrev & $ 93.22$${\scriptstyle\pm 0.22}$ & $ 93.55$${\scriptstyle\pm 0.33}$ & $ 93.90$${\scriptstyle\pm 0.28}$ & $ 93.88$${\scriptstyle\pm 0.22}$ & $\bf 77.00$${\scriptstyle\pm 0.18}$ & $ 75.93{\scriptstyle\pm 0.13}$ & $46.91$${\scriptstyle\pm 1.02}$ & $\bf 44.44$${\scriptstyle\pm 2.28}$ \\
    \rowcolor{blue!10} \methodabbrev-N & $\bf 93.33{\scriptstyle\pm 0.38}$ & $\bf 93.89$${\scriptstyle\pm 0.22}$ &  $\bf 94.12{\scriptstyle\pm 0.03}$ & $\bf 94.09{\scriptstyle\pm 0.26}$ & $76.40{\scriptstyle\pm 0.60}$ & $75.13{\scriptstyle\pm 0.13}$ & $40.07{\scriptstyle\pm 1.87}$ & $37.04{\scriptstyle\pm 0.89}$\\
    Upper Bound & $96.78{\scriptstyle\pm 0.11}$ & $98.00{\scriptstyle\pm 0.00}$ & $96.28{\scriptstyle\pm 0.13}$ & $96.99{\scriptstyle\pm 0.07}$ & $82.27{\scriptstyle\pm 0.13}$ & $83.73{\scriptstyle\pm 0.07}$ & $72.56{\scriptstyle\pm 1.87}$ & $86.20{\scriptstyle\pm 1.02}$ \\
    \bottomrule
  \end{tabular}
  }
  \caption{
  \textbf{Performance comparison on benchmarks across methods on LLMs.}
  Values represent mean accuracy ± standard error over three independent evaluation runs.
  \textbf{Bold} indicates the best-performing method per column based on the mean accuracy. 
  %
  Sampling parameters: $k=\{6,32\}$ for Gemma-2-9B, Llama-3.1-8B, and Qwen3-8B; $k=\{6,16\}$ for Llama-3.1-70B and Qwen3-32B.
  }
  \label{tab:model-comparison_breakdown}
\end{table*}
\subsection{Analysis between Token Length, PiCSAR score, and Model Performance}
\begin{figure*}[ht]
    \centering
    \begin{subfigure}{0.45\textwidth}
        \centering
        \includegraphics[width=\textwidth]{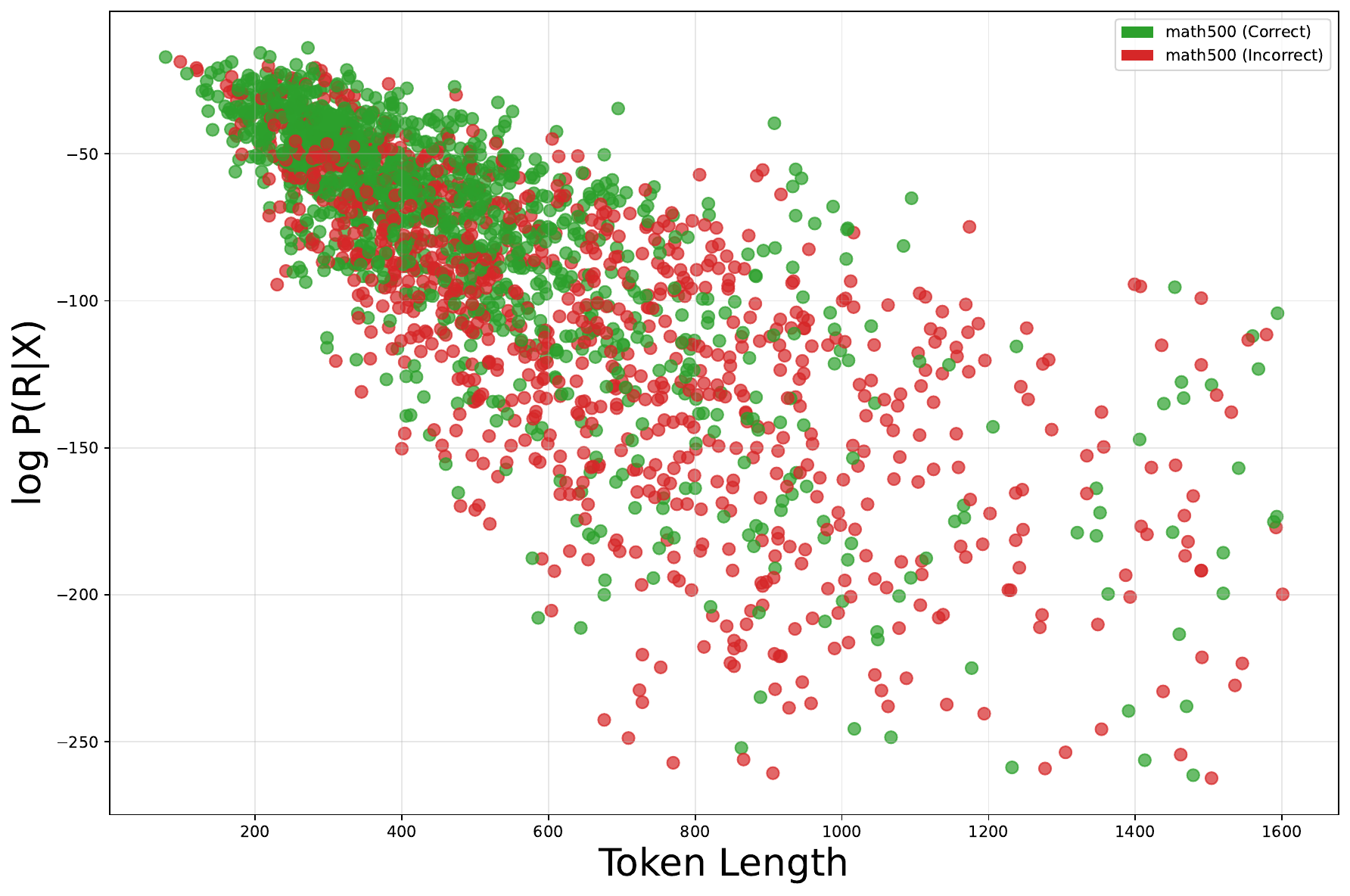}
        \caption{Length vs. Accuracy}
        \label{fig:length_vs_accuracy}
    \end{subfigure}
    \hfill
    \begin{subfigure}{0.45\textwidth}
        \centering
        \includegraphics[width=\textwidth]{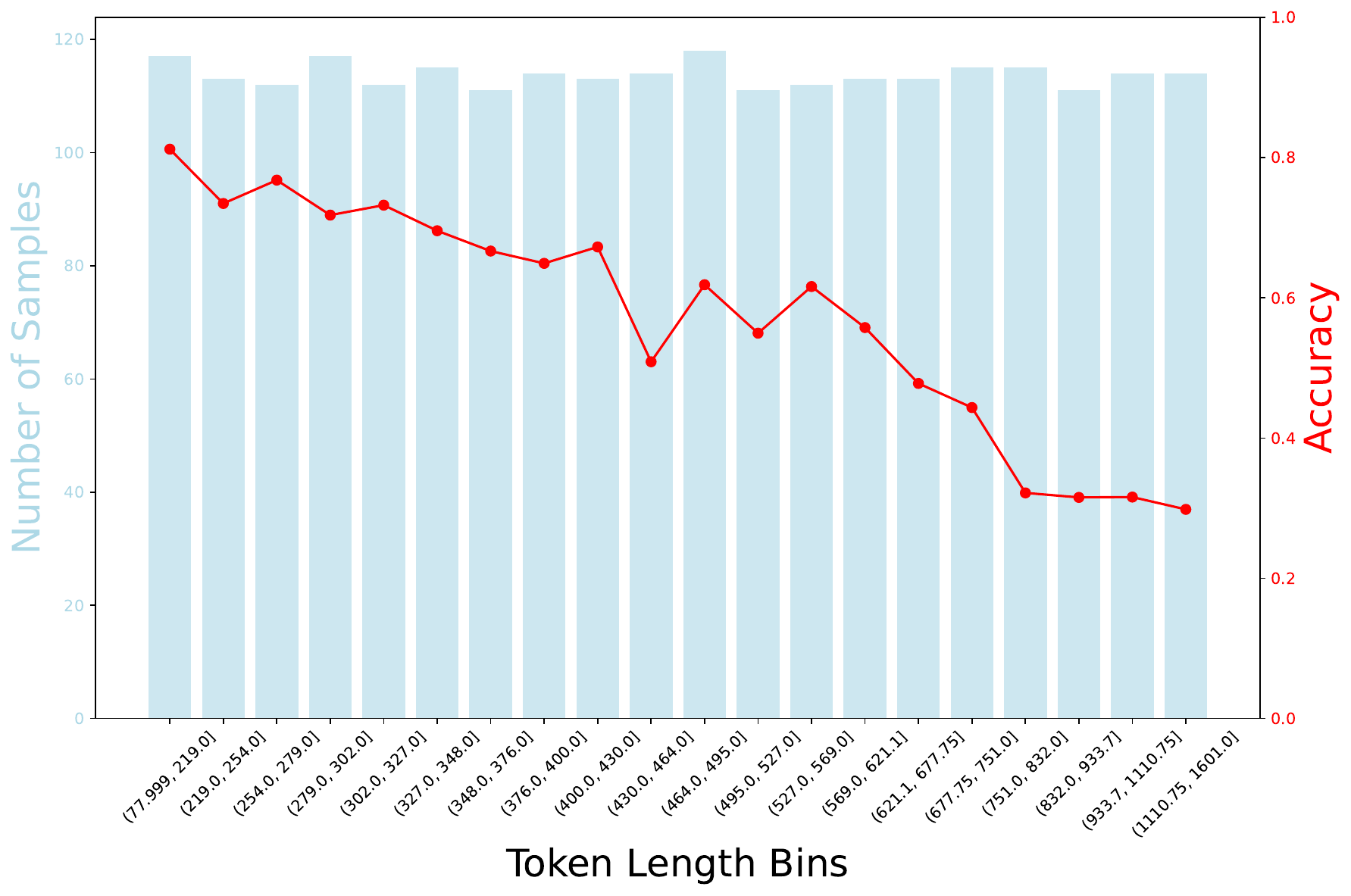}
        \caption{Information Plane Quadrants}
        \label{fig:info_plane_quadrants}
    \end{subfigure}

    \caption{Relationship between token length, probability, and accuracy.}
    \label{fig:merged_length_accuracy}
\end{figure*}
Figure \ref{fig:length_vs_accuracy} shows that correct instances predominantly cluster in regions of high probability and short sequence length, indicating that concise reasoning is strongly associated with higher quality. This pattern is reinforced by Figure \ref{fig:info_plane_quadrants}, which demonstrates a consistent decline in accuracy as sequence length grows. Together, the two figures highlight that shorter, more confident reasoning trajectories tend to yield more accurate performance.

\subsection{Ablation Studies on LLMs and LRMs}

The results for \methodabbrev-N are included in Table~\ref{tab:model-comparison_breakdown} and Table~\ref{tab:deepseek_qwen_ablation}. As shown, both \methodabbrev and \methodabbrev-N consistently surpass the other baselines, including their corresponding reasoning confidence metrics (with and without normalisation). The performance difference between \methodabbrev and \methodabbrev-N is not consistently in one direction; each variant excels on different model-dataset combinations. For instance, \methodabbrev-N shows stronger performance with Gemma-2-9B on MATH500 ($k=6$) and GPQA-Diamond, whereas the non-normalised \methodabbrev is clearly superior for Llama-3.1-8B across most settings. This suggests that the utility of length normalisation may depend on model-specific characteristics, such as tendencies towards verbosity.  

Based on Table~\ref{tab:model-comparison_breakdown}, we also observe that 20/40 results of the length-normalised (PiCSAR-N) versions outperform the non-length normalised versions (PiCSAR), demonstrating that length-normalisation does not perform worse than the non-length normalised version. This suggests that length normalisation is not detrimental and does not consistently weaken PiCSAR.


\begin{table*}[t]
  \centering
  \resizebox{\textwidth}{!}{
  \renewcommand{\arraystretch}{0.9}
  \begin{tabular}{llcccccc}
    \toprule
    \textbf{Method} & \textbf{AIME\,2024} & \textbf{AIME\,2025} & \textbf{MATH500} & \textbf{SVAMP} & \textbf{GSM8K} & \textbf{GPQA-Diamond} \\
    \midrule
    \multicolumn{7}{c}{\emph{DS-Distill-llama-3-8B}} \\ \midrule
    Reasoning Confidence & $44.43{\scriptstyle\pm 5.56}$ & $\bf 35.56$${\scriptstyle\pm 1.11}$ & $66.60{\scriptstyle\pm 0.60}$ & $83.67{\scriptstyle\pm 0.00}$ & $72.97{\scriptstyle\pm 0.30}$ & $46.97{\scriptstyle\pm 0.29}$\\
    Reasoning Confidence (Normalised) & $33.33{\scriptstyle\pm 3.85}$ & $28.89{\scriptstyle\pm 1.12}$ & $65.70{\scriptstyle\pm 1.30}$ & $83.00{\scriptstyle\pm 0.13}$ & $76.08{\scriptstyle\pm 0.23}$ & $41.41{\scriptstyle\pm 1.05}$ \\
    Answer Confidence  & $42.22{\scriptstyle\pm 4.01}$ & $32.22{\scriptstyle\pm 1.11}$ & $\bf 67.60$${\scriptstyle\pm 1.80}$ & \bf $88.33{\scriptstyle\pm 0.16}$ & $76.06{\scriptstyle\pm 0.43}$ & $\bf 48.99$${\scriptstyle\pm 1.62}$\\
    \rowcolor{blue!10}   \methodabbrev           & $\bf 47.78$${\scriptstyle\pm 4.01}$ & $ 33.33$${\scriptstyle\pm 1.13}$ & 
    $67.20$${\scriptstyle\pm 0.60}$ & 
    $85.67{\scriptstyle\pm 0.07}$ &  $\bf76.42$${\scriptstyle\pm 0.16}$ & 
    $ 47.31$${\scriptstyle\pm 0.17}$\\
    \rowcolor{blue!10}   \methodabbrev-N & $40.00{\scriptstyle\pm 5.09}$ & $32.22{\scriptstyle\pm 1.13}$ & $ 67.40$${\scriptstyle\pm 1.00}$ & $\bf 89.00$${\scriptstyle\pm 0.00}$ & $75.73{\scriptstyle\pm 0.41}$ & $ 47.47$${\scriptstyle\pm 2.78}$ \\
    Upper Bound & $66.67{\scriptstyle\pm 5.09}$ & $51.11{\scriptstyle\pm 1.11}$ & $82.00{\scriptstyle\pm 0.13}$ & $95.67{\scriptstyle\pm 0.00}$ & $92.91{\scriptstyle\pm 0.35}$ & $77.27{\scriptstyle\pm 0.77}$\\
    \midrule
    \multicolumn{7}{c}{\emph{DS-Distill-Qwen-2.5-7B}} \\ \midrule
    Reasoning Confidence & $57.78{\scriptstyle\pm 1.11}$ & $51.11{\scriptstyle\pm 1.11}$ & $72.93{\scriptstyle\pm 0.81}$ & $91.33{\scriptstyle\pm 0.58}$ & $87.83{\scriptstyle\pm 0.13}$ & $52.02{\scriptstyle\pm 2.81}$ & \\
      Reasoning Confidence (Normalised)     & $54.44{\scriptstyle\pm 2.22}$ & $45.56{\scriptstyle\pm 2.22}$ & $74.20{\scriptstyle\pm 1.10}$ & $90.33{\scriptstyle\pm 0.58}$ & $88.26{\scriptstyle\pm 0.20}$ & $45.96{\scriptstyle\pm 2.67}$ \\
      Answer Confidence  & $50.00{\scriptstyle\pm 5.09}$ & $44.44{\scriptstyle\pm 2.22}$ & $72.60{\scriptstyle\pm 0.23}$ & $91.00{\scriptstyle\pm 0.51}$ & $88.91{\scriptstyle\pm 0.08}$ & $\bf 53.20$${\scriptstyle\pm 2.19}$\\
    \rowcolor{blue!10}   \methodabbrev            & $\bf 61.11$${\scriptstyle\pm 1.11}$ & $\bf 51.11$${\scriptstyle\pm 1.11}$ & $\bf 74.00$${\scriptstyle\pm 0.70}$ & $\bf 91.78$${\scriptstyle\pm 0.48}$ & $88.18{\scriptstyle\pm 0.07}$ & $ 52.36$${\scriptstyle\pm 2.88}$\\
    \rowcolor{blue!10}   \methodabbrev-N & $57.78{\scriptstyle\pm 2.22}$ & $48.89{\scriptstyle\pm 2.22}$ & $73.40{\scriptstyle\pm 1.10}$ & $\bf 91.78{\scriptstyle\pm 0.29}$ & $\bf 89.60$${\scriptstyle\pm 0.18}$ & $50.34{\scriptstyle\pm 2.19}$ \\
    Upper Bound & $72.22{\scriptstyle\pm 1.11}$ & $70.00{\scriptstyle\pm 0.00}$ & $83.33{\scriptstyle\pm 0.18}$ & $96.33{\scriptstyle\pm 0.38}$ & $96.79{\scriptstyle\pm 0.13}$ & $79.12{\scriptstyle\pm 2.07}$\\
    \midrule
    \multicolumn{7}{c}{\emph{Qwen3-8B}} \\ \midrule
     Reasoning Confidence & $80.00{\scriptstyle\pm 0.00}$ & $68.89{\scriptstyle\pm 2.22}$ & $79.20{\scriptstyle\pm 0.00}$ & $93.00{\scriptstyle\pm 0.33}$ & $95.92{\scriptstyle\pm 0.03}$ & $58.59{\scriptstyle\pm 1.62}$\\
     Reasoning Confidence (Normalised) & $67.78{\scriptstyle\pm 2.22}$ & $65.56{\scriptstyle\pm 4.01}$ & $80.00{\scriptstyle\pm 0.00}$ & $93.56{\scriptstyle\pm 0.56}$ & $95.72{\scriptstyle\pm 0.05}$ & $56.23{\scriptstyle\pm 1.76}$ & \\
    Answer Confidence & $76.67{\scriptstyle\pm 0.00}$ & $\bf 73.33$${\scriptstyle\pm 1.92}$ & $80.13{\scriptstyle\pm 0.33}$ & $93.78{\scriptstyle\pm 0.11}$ & $95.37{\scriptstyle\pm 0.00}$ & $60.61{\scriptstyle\pm 0.29}$\\
    \rowcolor{blue!10}   \methodabbrev            & $\bf 81.33$${\scriptstyle\pm 1.34}$ & $ 68.89$${\scriptstyle\pm 2.22}$ & $  80.60$${\scriptstyle\pm 0.13}$ & $\bf 94.33$${\scriptstyle\pm 0.33}$ & $\bf 95.94$${\scriptstyle\pm 0.04}$ & $  59.43$${\scriptstyle\pm 1.61}$\\
    \rowcolor{blue!10}   \methodabbrev-N & $76.67{\scriptstyle\pm 3.33}$ & $ 70.00$${\scriptstyle\pm 5.09}$ & $\bf 89.67{\scriptstyle\pm 0.37}$ & $94.22{\scriptstyle\pm 0.56}$ & $ 95.08{\scriptstyle\pm 0.03}$ & $\bf 61.11$${\scriptstyle\pm 1.77}$ &\\
    Upper Bound & $87.78{\scriptstyle\pm 1.11}$ & $82.22{\scriptstyle\pm 1.11}$ & $84.00{\scriptstyle\pm 0.12}$ & $97.56{\scriptstyle\pm 0.11}$ & $97.54{\scriptstyle\pm 0.03}$ & $80.13{\scriptstyle\pm 0.45}$\\
    \bottomrule
  \end{tabular}
  }
  \caption{
    \textbf{Performance comparison of model across various baselines and benchmarks on LRMs, measured in terms of accuracy (\%).} For all the evaluations, we use $k=6$ sampling. \emph{PiCSAR outperforms all baselines with more pronounced gains in more challenging benchmarks.}}
  %
  \label{tab:deepseek_qwen_ablation}
\end{table*}


We further conducted ablation studies on LRMs, with results reported in Table~\ref{tab:deepseek_qwen_ablation}. Here, we compare \methodabbrev and \methodabbrev-N against both standard and normalised reasoning confidence, as well as answer confidence. The results confirm that our joint probability methods, \methodabbrev and \methodabbrev-N, consistently achieve top performance, similar to the findings with LLMs. Interestingly, we observe that maximising answer confidence alone yields strong results, sometimes comparable to \methodabbrev, particularly on the DS-Distill-llama-3-8B model. This reinforces the value of the answer confidence signal while highlighting the general effectiveness of \methodabbrev's approach in combining both reasoning and answer confidence.

\subsection{Further Analysis on Length-Normalised Variant: \methodabbrev-N}
\label{sec: picsarn_length}
In this section, we clarify the distinctions between \methodabbrev and its length-normalised counterpart, \methodabbrev-N, establishing empirical evidence for when length normalisation should be applied.

As shown in \cref{tab:model-comparison_breakdown}, the performance gap between \methodabbrev and \methodabbrev-N is generally marginal, with neither variant strictly dominating across all model-task configurations. We introduce \methodabbrev-N primarily as an ablation to confirm that \methodabbrev's strong performance is not merely an artifact of systematically penalising longer generations. Therefore, we recommend the unnormalised \methodabbrev as the default selection strategy.
Our empirical analysis (see\S~\ref{sec:section3}) suggests that correct reasoning chains exhibit high ``information density''. They accumulate log-probability mass efficiently as they converge toward the final answer. The unnormalised joint log-likelihood naturally favours reasoning paths that are both highly probable and structurally concise, effectively penalising indirect reasoning paths.

Conversely, \methodabbrev-N proves beneficial primarily for weaker models prone to ``verbose hallucinations'', instances where a model generates locally plausible but logically stagnant text that accumulates massive negative log-probabilities strictly due to sequence length. For highly capable reasoners (\eg Llama-3.1, Qwen3), the unnormalised score remains highly robust, as these models' sequence-level log-probabilities serve as well-calibrated proxies for both logical coherence and problem-solving efficiency.

\subsection{The Importance of Selection: Interpreting the Upper Bound} 
\label{sec:upperbound}
While \methodabbrev consistently outperforms other heuristics, it necessarily falls short of the oracle \emph{Upper Bound}, whose behaviour provides insight into the underlying challenges. On easier benchmarks such as SVAMP and GSM8K, the upper bound saturates quickly. For instance, increasing the sample size from $k=6$ to $k=32$ with Llama-3.1-70B on GSM8K raises accuracy only marginally from 96.91\% to 97.44\%, indicating that correct reasoning paths are usually present in small sample sets, and that selection rather than generation is the main bottleneck. In contrast, on more demanding tasks such as MATH500 and GPQA-Diamond, the upper bound continues to rise with larger $k$, as seen with Gemma-2-9B on GPQA-Diamond where accuracy jumps from 55.22\% to 82.49\%, reflecting the intrinsic difficulty of generating correct answers. In both regimes, \methodabbrev demonstrates its value: in selection-limited settings, it reliably identifies correct candidates from small pools, while in generation-limited scenarios, it narrows the gap to the oracle by detecting correct reasoning even when correct answers are sparse, highlighting that improving selection is often as important as 
enlarging the sampling budget.

\subsection{Analysis of Fallback Mechanism}
\label{sec:fallback}
To assess how sensitive our method is to the penalty assigned when a generation fails, \ie no answer token is produced and the answer-confidence term cannot be calculated, we tested several fallback values for the Answer Confidence score ($Y$). Specifically, we compared our default setting of $\log p(y \mid r, X) = -10$ with more conservative penalties of $Y = -20$ and $Y = -100$. As shown in Table~\ref{tab:fallback_sensitivity}, downstream accuracy is unchanged across all configurations. This indicates that, as long as the fallback value is sufficiently low to denote a failure state, its precise magnitude does not affect candidate rankings.

{\color{blue}
\begin{table}[h!]
\centering
\begin{tabular}{l c}
\toprule
\textbf{ $\log p(y\mid r,X)$} & \textbf{Accuracy} \\
\midrule
$ -10$  & 53.40\% \\
$ -20$  & 53.40\% \\
$-100$ & 53.40\% \\
\bottomrule
\end{tabular}
\caption{Sensitivity analysis of the Answer Confidence fallback value ($Y$) on model accuracy. The performance is robust to the magnitude of the penalty.}
\label{tab:fallback_sensitivity}
\vspace{-2ex}
\end{table}
}

\subsection{Analysis of Performance with Number of Samples and Temperature}

\begin{table}[h!]
\centering
\small
\begin{tabular}{c c c}
\toprule
\textbf{Samples} & \textbf{PiCSAR (\%)} & \textbf{Self-Consistency (\%)} \\
\midrule
6  & 89.11 & 88.15 \\
10 & 89.89 & 88.56 \\
16 & 89.89 & 88.11 \\
32 & 90.22 & 88.89 \\
\bottomrule
\end{tabular}
\caption{Scaling analysis of \textsc{Gemma-2-9B} on \textsc{SVAMP} comparing \methodabbrev{} against Self-Consistency across varying sample counts.}
\label{tab:svamp_samples}
\end{table}

We first examine the scaling behavior of \methodabbrev{} regarding the number of candidate generations ($k$). We evaluate \textsc{Gemma-2-9B} on the \textsc{SVAMP} dataset with sample budgets ranging from $k=6$ to $k=32$. As shown in Table~\ref{tab:svamp_samples}, \methodabbrev{} exhibits scaling properties, with accuracy consistently improving as the candidate pool expands (rising from 89.11\% at $k=6$ to 90.22\% at $k=32$). In contrast, Self-Consistency plateaus earlier and is consistently outperformed by \methodabbrev{}. This indicates that \methodabbrev{} is more effective at leveraging larger compute budgets to identify correct reasoning chains.

Additionally, we assess the stability of our method with respect to generation stochasticity by comparing performance at sampling temperatures of $T=0.7$ and $T=1.0$. The results, summarized in Table~\ref{tab:svamp_samples}, reveal negligible performance variance (89.89\% vs. 89.67\%). These results indicate that \methodabbrev{} is robust to moderate changes in generation hyperparameters and maintains high precision even under more stochastic sampling conditions ($T=1.0$).

\begin{table*}[!t]
\centering
\small
\setlength{\tabcolsep}{6pt} 
\renewcommand{\arraystretch}{0.85}

\begin{tabular}{l l r c r c c}
\toprule
\textbf{Model} & \textbf{Metric} & \textbf{$t$} & \textbf{$p$-value} & \textbf{$U$-stat} & \textbf{$d$} & \textbf{Mean (C / I)} \\
\midrule

\multirow{2}{*}{\textbf{LLaMA-3.1-8B}} 
 & $\log p(y \mid r, x)$ & 4.57 & $6.06 \times 10^{-6}$ & 38441 & 0.41 & $-4.21$ / $-5.75$ \\
 & $\log p(r \mid x)$    & 9.11 & $2.00 \times 10^{-18}$ & 45115 & 0.82 & $-45.78$ / $-67.43$ \\
\addlinespace

\multirow{2}{*}{\textbf{LLaMA-3.1-70B}} 
 & $\log p(y \mid r, x)$ & 5.76 & $1.48 \times 10^{-8}$ & 41596 & 0.54 & $-0.41$ / $-1.47$ \\
 & $\log p(r \mid x)$    & 6.99 & $8.76 \times 10^{-12}$ & 39096 & 0.66 & $-39.87$ / $-53.69$ \\
\addlinespace

\multirow{2}{*}{\textbf{Gemma-2-9B}} 
 & $\log p(y \mid r, x)$ & 9.03 & $3.70 \times 10^{-18}$ & 42086 & 0.81 & $-0.37$ / $-2.68$ \\
 & $\log p(r \mid x)$    & 9.03 & $3.85 \times 10^{-18}$ & 45831 & 0.81 & $-18.64$ / $-30.80$ \\
\addlinespace

\multirow{2}{*}{\textbf{Qwen3-8B}} 
 & $\log p(y \mid r, x)$ & 5.37 & $1.24 \times 10^{-7}$ & 36835 & 0.54 & $-0.94$ / $-2.36$ \\
 & $\log p(r \mid x)$    & 5.17 & $3.39 \times 10^{-7}$ & 31131 & 0.52 & $-41.88$ / $-68.41$ \\
\addlinespace

\multirow{2}{*}{\textbf{Qwen3-32B}} 
 & $\log p(y \mid r, x)$ & 6.09 & $2.26 \times 10^{-9}$ & 34500 & 0.64 & $-0.38$ / $-1.82$ \\
 & $\log p(r \mid x)$    & 4.98 & $8.81 \times 10^{-7}$ & 27660 & 0.52 & $-61.92$ / $-95.85$ \\
\addlinespace

\multirow{2}{*}{\textbf{Think-Qwen3-8B}} 
 & $\log p(y \mid r, x)$ & 4.97 & $9.11 \times 10^{-7}$ & 27177 & 0.56 & $-2.17$ / $-4.55$ \\
 & $\log p(r \mid x)$    & 2.67 & $0.008$ & 21190 & 0.30 & $-418.77$ / $-587.10$ \\
\addlinespace

\multirow{2}{*}{\shortstack[l]{\textbf{Think-DS-R1}\\\textbf{Distill-Qwen-7B}}} 
 & $\log p(y \mid r, x)$ & 3.87 & $1.21 \times 10^{-4}$ & 29105 & 0.39 & $-1.69$ / $-2.76$ \\
 & $\log p(r \mid x)$    & 2.04 & $0.042$ & 29023 & 0.21 & $-174.75$ / $-254.23$ \\
\addlinespace

\multirow{2}{*}{\shortstack[l]{\textbf{Think-DS-R1}\\\textbf{Distill-LLaMA-8B}}} 
 & $\log p(y \mid r, x)$ & 5.99 & $4.00 \times 10^{-9}$ & 39822 & 0.57 & $-0.97$ / $-3.20$ \\
 & $\log p(r \mid x)$    & 4.63 & $4.60 \times 10^{-6}$ & 31908 & 0.44 & $-246.18$ / $-500.00$ \\

\bottomrule
\end{tabular}
\caption{Statistical tests on Math500 comparing Correct (C) and Incorrect (I) samples. We report $t$-statistic ($t$), Mann-Whitney $U$ ($U$), and Cohen's $d$ ($d$) for both prediction confidence ($\log p(y \mid r, x)$) and reasoning compression ($\log p(r \mid x)$). All differences are significant at $p < 0.05$.}
\label{tab:math500-stats}
\vspace{-2ex}
\end{table*}

\subsection{Nemenyi Post-hoc Test for PiCSAR}
\label{sec:nemenyi-test}

\begin{figure}[ht]
  \centering
  \includegraphics[width=0.48\textwidth]
  {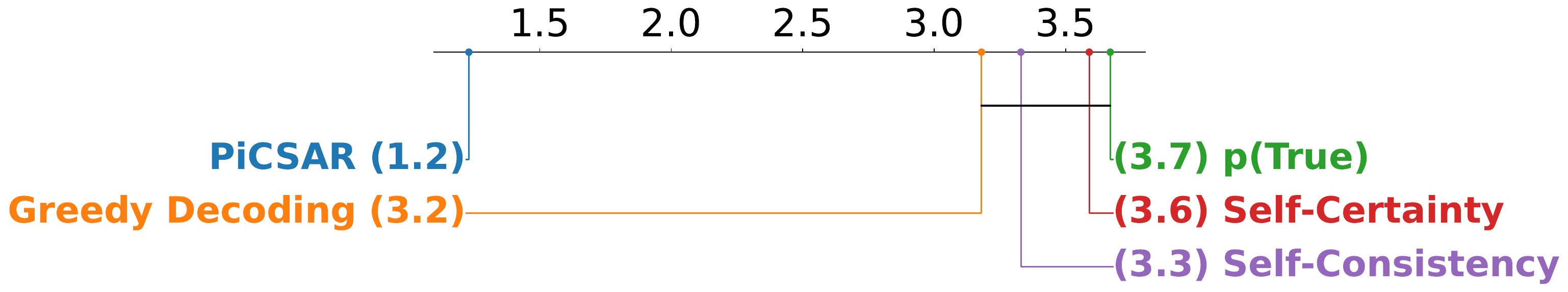}
  \caption{Critical difference diagram based on Nemenyi test ($p<0.05$) for LLM. PiCSAR significantly outperforms all competing methods, while Self-Consistency, Self-Certainty, and p(True) show no significant difference from each other.}
  \label{fig:cd_nemenyi}
\end{figure}

\begin{figure}[ht]
  \centering
  \includegraphics[width=0.48\textwidth]
  {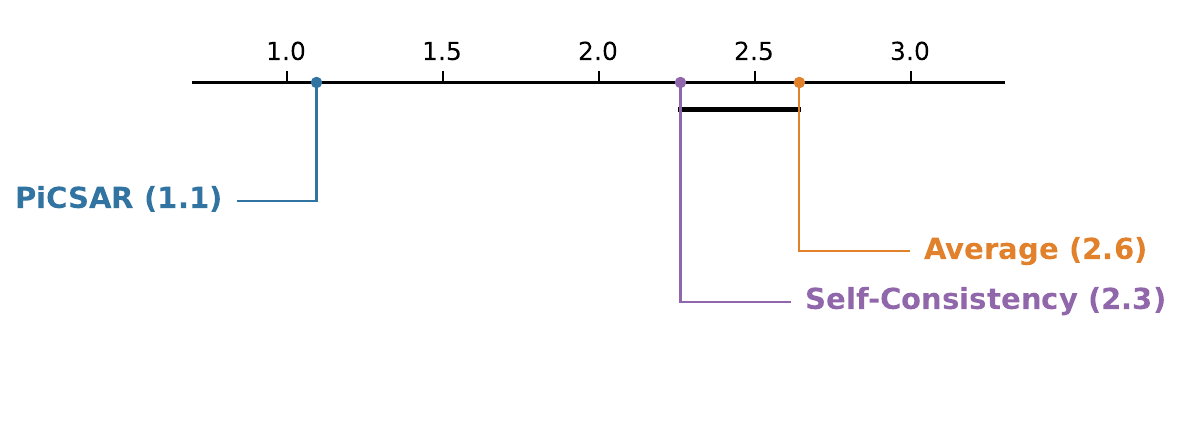}
  \caption{Critical difference diagram based on Nemenyi test ($p<0.05$) for LRM. PiCSAR significantly outperforms the next-best methods, while Self-Consistency and Average show no significant difference from each other.}
  \label{fig:cd_nemenyi_lrm}
\vspace{-3ex}
\end{figure}

In Figure~\ref{fig:cd_nemenyi} and Figure~\ref{fig:cd_nemenyi_lrm} we show the critical diagrams obtained by performing the Nemenyi post-hoc test. In the critical diagram, the group of methods that do not differ significantly (significance level 0.05) are connected through a horizontal line.

At $p<0.05$, the Nemenyi post-hoc test shows that PiCSAR (average rank $\approx 1.2$) is significantly better than all other methods, as its rank is well separated beyond the critical difference from Greedy Decoding, Self-Consistency, Self-Certainty, and \emph{p(True)}. This indicates that PiCSAR consistently outperforms the alternatives across datasets, and its superior performance is statistically robust rather than due to random variation.

As for LRM, statistical analysis utilising the Friedman test revealed highly significant performance differences across the methods ($p < 9.96 \times 10^{-7}$). Subsequent Nemenyi post-hoc comparisons confirmed that PiCSAR significantly outperforms both the Average and Self-Consistency baselines, showing mean rank differences of $1.55$ and $1.17$ respectively, both of which substantially exceed the critical difference of $0.72$ at $\alpha=0.05$. 

\section{Sentence Level Analysis}
\label{appendix:sentence_level_analysis}
In this section, we provide analysis between senetence level of PiCSAR and PiCSAR-N. 

Our empirical analysis across \textsc{AIME2025}, \textsc{AIME2024}, and \textsc{MATH500} with \textsc{DeepSeek-R1-Distill-Qwen-2.5-7B} demonstrates that this sensitivity is well-calibrated and reflects genuine quality differences rather than arbitrary length penalisation. Excessively long reasoning chains ($>$10K tokens) achieve only 14.1\%, 13.5\%, and 36.5\% accuracy on these three benchmarks, compared to 92.7\%, 82.1\%, and 69.3\% for the 1K--5K token range. Wrong answers are consistently 2--3$\times$ longer than correct ones (e.g., 15,761 vs.\ 5,651 tokens on \textsc{AIME2025}), confirming that excessive length may signal model uncertainty.

We further conducted a sentence-level trajectory analysis of $\log p(y|r,x)$ across reasoning chains for \textsc{Qwen3-8B}.

On \textsc{AIME2024} (thinking mode):
\begin{itemize}
    \item The highest-\textsc{PiCSAR} generations (avg $\sim$650 sentences) begin with $\log p(y \mid r,x)$ around $-9.5$ at early sentence positions and rise steadily, peaking near $-0.5$ --- demonstrating that each reasoning step productively advances toward the correct answer.
    \item The lowest-\textsc{PiCSAR} generations (avg $\sim$1,700 sentences, $\sim$2$\times$ longer) have $\log p(y \mid r,x)$ oscillating erratically between $-5$ and $-7$ throughout, never reaching the sharp convergent peak observed in high-\textsc{PiCSAR} samples.
\end{itemize}

On \textsc{MATH500} (no-think mode), the same pattern holds:
\begin{itemize}
    \item High-\textsc{PiCSAR} generations (avg $\sim$660 sentences) show $\log p(y \mid r,x)$ climbing monotonically from $-9.6$ to a peak near $-0.05$.
    \item Low-\textsc{PiCSAR} generations (avg $\sim$1,057 sentences, 1.6$\times$ longer) rise more slowly, plateau at substantially worse values ($-12.0$), and exhibit noisy fluctuations rather than clean convergence.
\end{itemize}

These trajectory analyses demonstrate that the additional length in low-\textsc{PiCSAR} generations does not yield meaningful progress in reasoning tasks --- the model cycles through unproductive loops without converging. \textsc{PiCSAR}'s unnormalised $\log p(r \mid x)$ faithfully captures this distinction by encoding the model's own confidence in a reasoning path, naturally assigning lower scores to long, uncertain chains. However, as shown in the example above regarding Qwen3-8B on AIME2024 and MATH500, when each generation converges to the answer, accuracy increases with the length. This length sensitivity is therefore a desirable property that rewards efficient, convergent reasoning over verbose, aimless generation. 

We further provide three different cases in Appendix~\ref{sec:further_example_picsar_scoring}: (1) a lengthy generation that arrives at the correct answer through extended thinking, and (2) a concise generation with high information density that likewise yields the correct answer. (3) The impact of answer confidence in generation selection.

\section{Additional Experiments for Confidence Information Plane}
\label{sec:information_plane_selection}
\begin{figure*}[htbp]
  \centering
  \scalebox{1.0}{
  \begin{subfigure}[b]{0.45\textwidth}
    \centering
    \includegraphics[width=\textwidth]{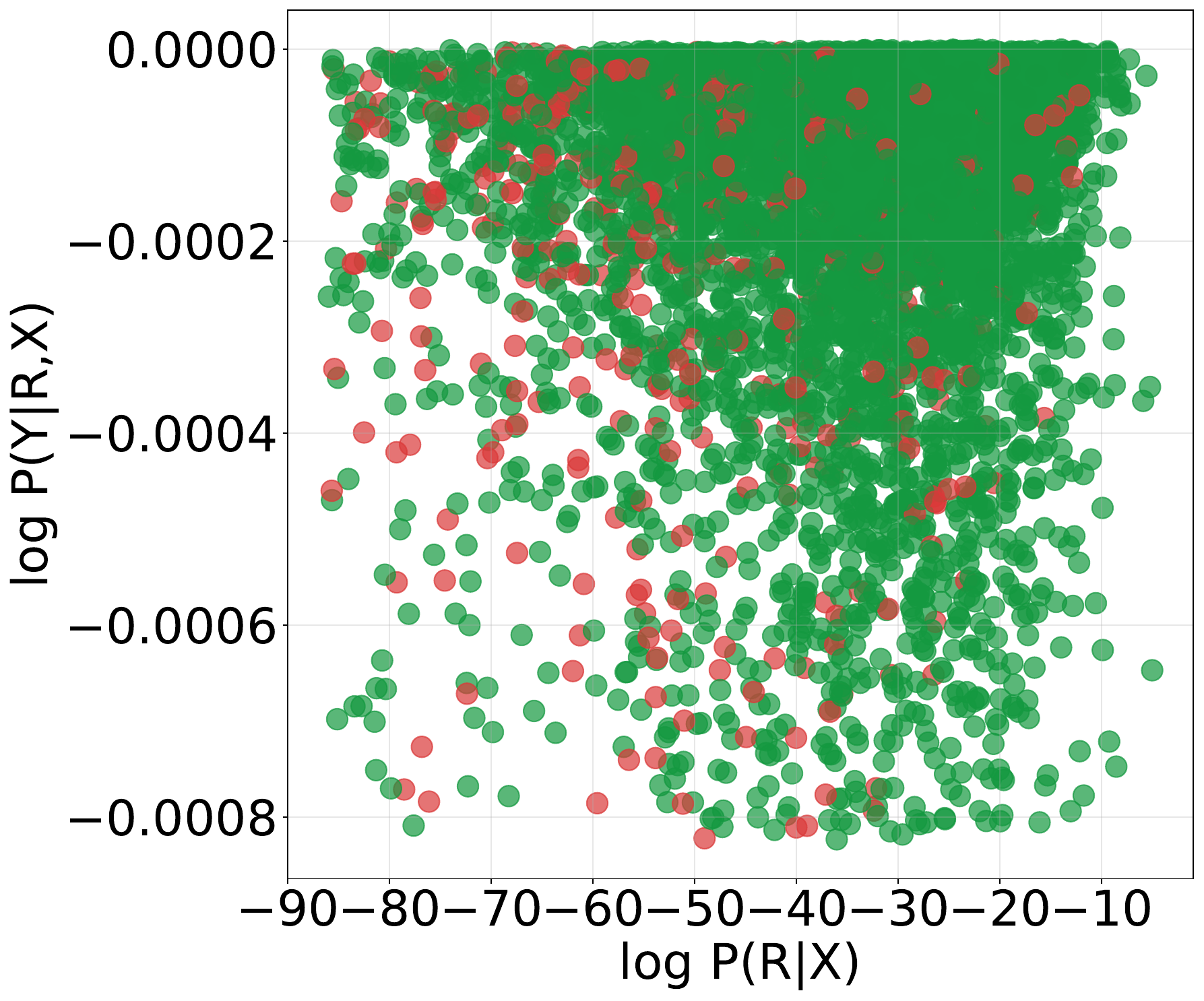}
    \caption{Continuous distribution without quadrant partitioning. }
    \label{fig:quadrats_split-gsm8k-llama-3.1}
  \end{subfigure}
  \hfill
  \begin{subfigure}[b]{0.44\textwidth}
    \centering
    \includegraphics[width=\textwidth]{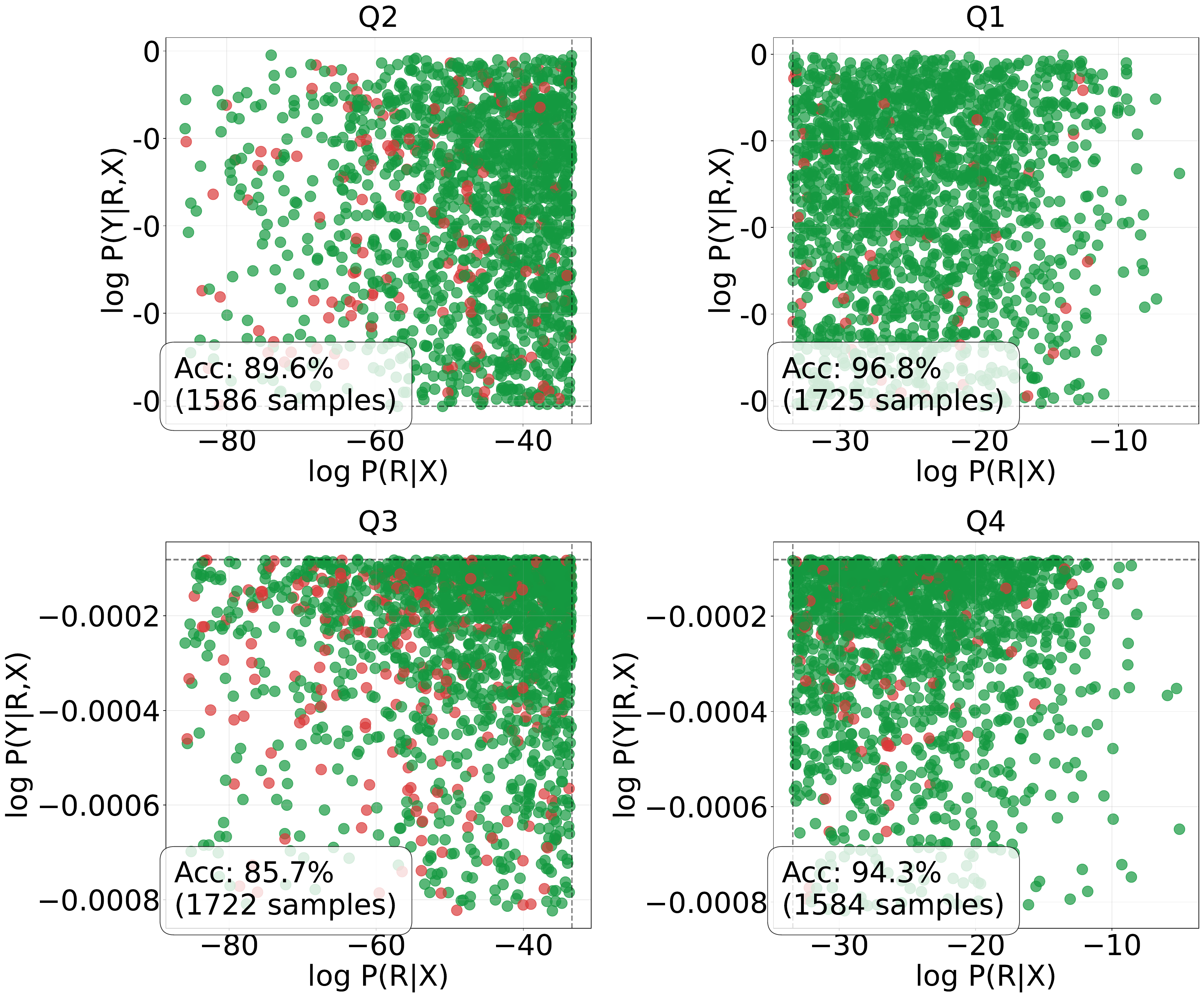}
    \caption{Quadrant-split representation with Q1–Q4 defined by median thresholds of reasoning and answer confidence. }
    \label{fig:quadrats_plane-gsm8k-llama-3.1}
  \end{subfigure}}
  \caption{Information Plane visualisations of Llama-3.1-8B on the GSM8K dataset ($k=6$). 
  Green indicates correct answers, Red indicates incorrect ones.}
  \label{fig:quadrats_subplot-llama-3.1}
\end{figure*}

\begin{figure*}[htbp]
  \centering
    \scalebox{1.0}{
  \begin{subfigure}[b]{0.4\textwidth}
    \centering
    \includegraphics[width=\textwidth]{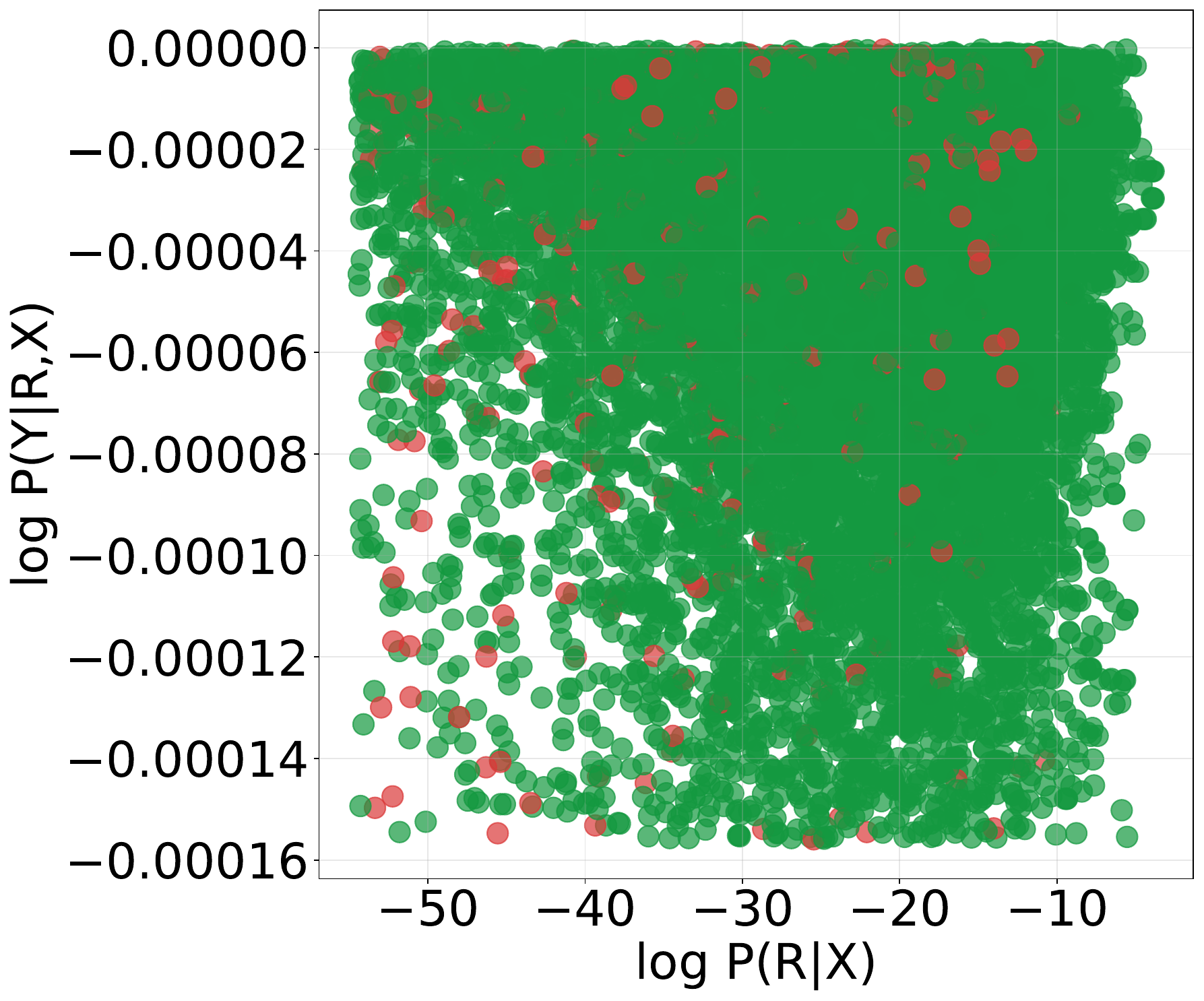}
    \caption{Continuous distribution without quadrant partitioning. }
    \label{fig:quadrats_split-qwen3-32b-gsm8k}
  \end{subfigure}
  \hfill
  \begin{subfigure}[b]{0.4\textwidth}
    \centering
    \includegraphics[width=\textwidth]{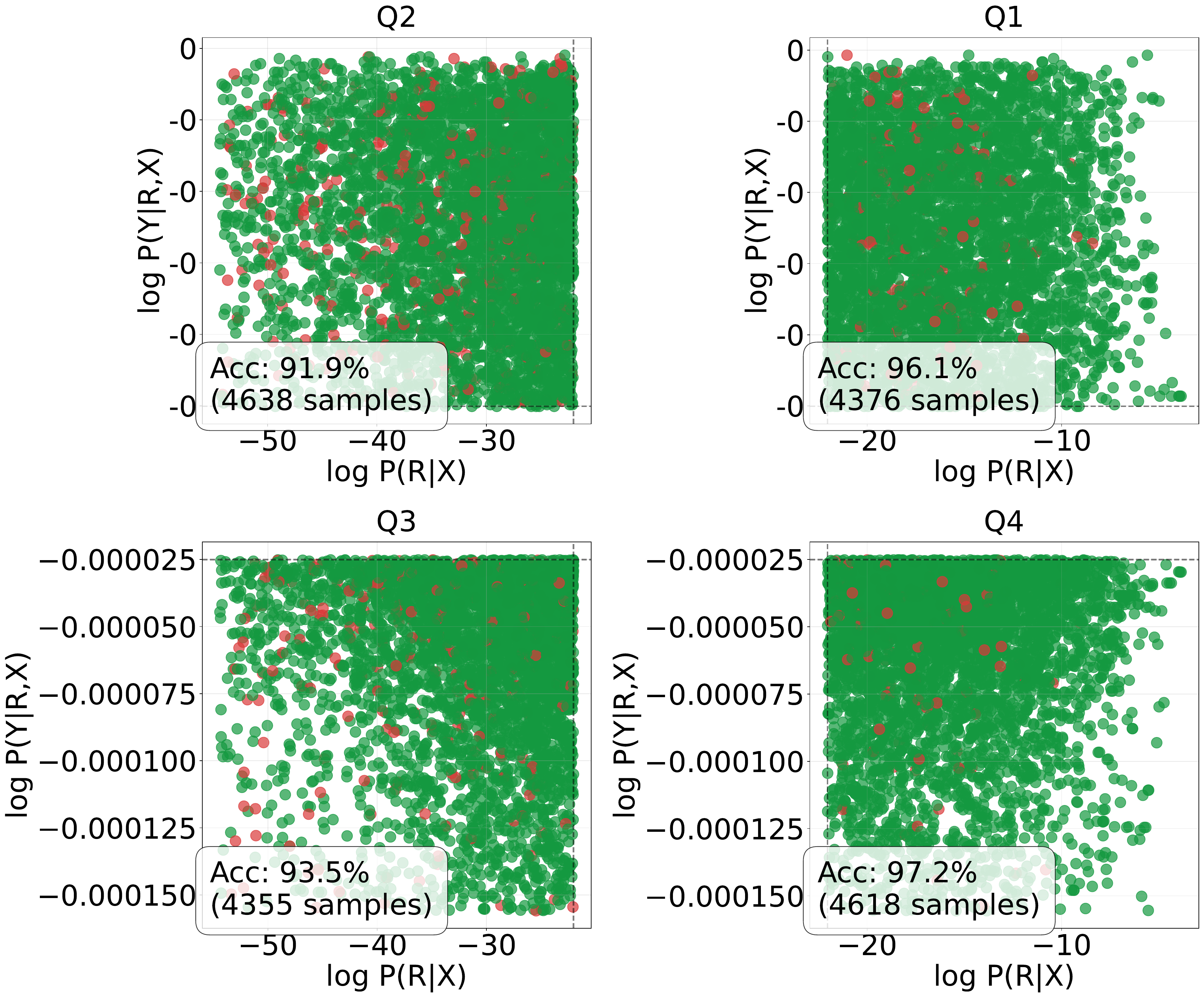}
    \caption{Quadrant-split representation with Q1–Q4 defined by median thresholds of reasoning and answer confidence. }
    \label{fig:quadrats_plane-qwen3-32b-gsm8k}
  \end{subfigure}}
  \caption{Information Plane visualisations of Gemma-2-9B on the GSM8K dataset ($k=6$). 
  Green indicates correct answers, Red incorrect ones.}
  \label{fig:quadrats_subplot-qwen3-32b-gsm8k}
\end{figure*}

\begin{figure*}[htbp]
  \centering
    \scalebox{1.0}{
  \begin{subfigure}[b]{0.45\textwidth}
    \centering
    \includegraphics[width=\textwidth]{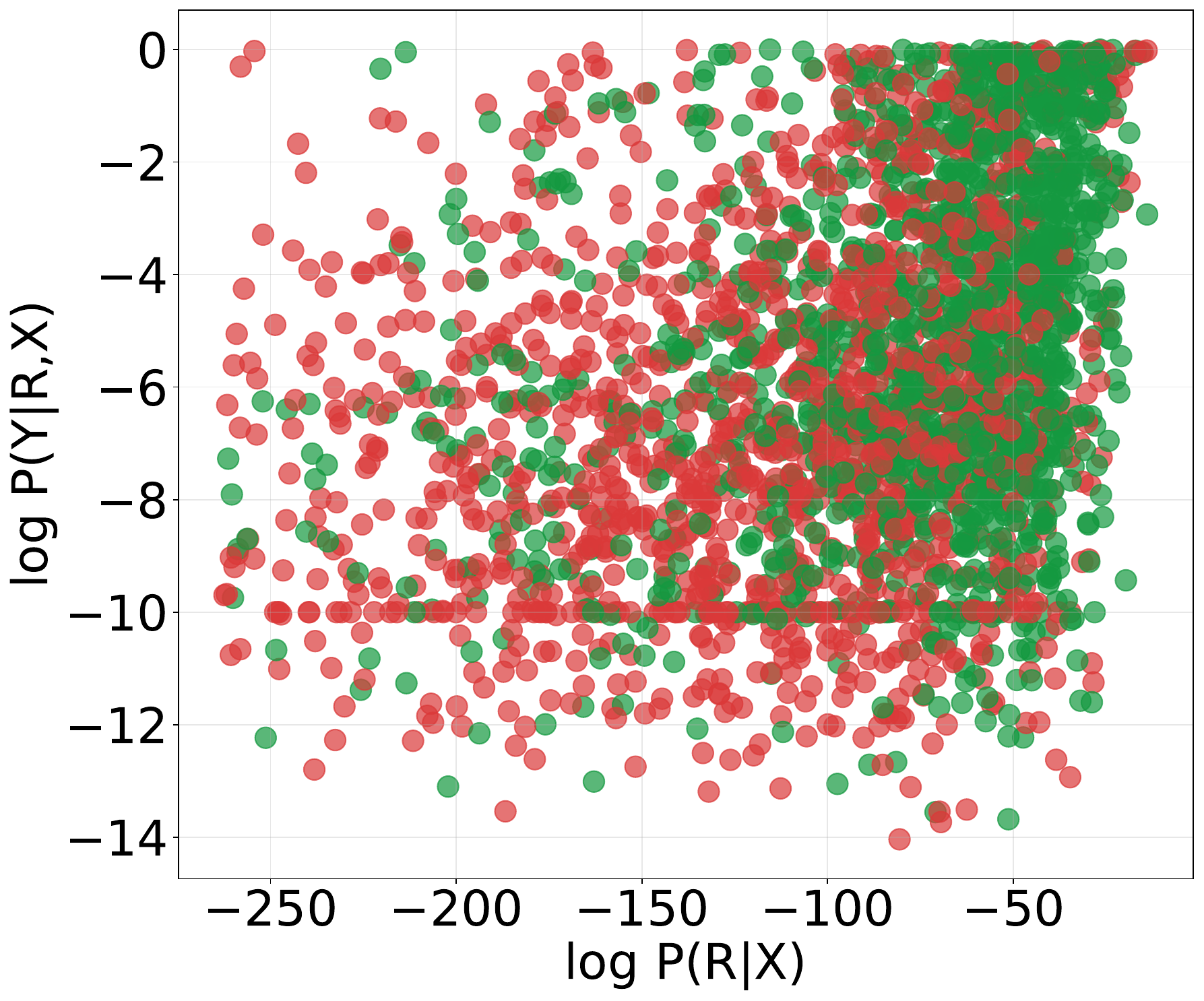}
    \caption{Continuous distribution without quadrant partitioning. }
    \label{fig:quadrats_split-llama-3.1-math500}
  \end{subfigure}
  \hfill
  \begin{subfigure}[b]{0.45\textwidth}
    \centering
    \includegraphics[width=\textwidth]{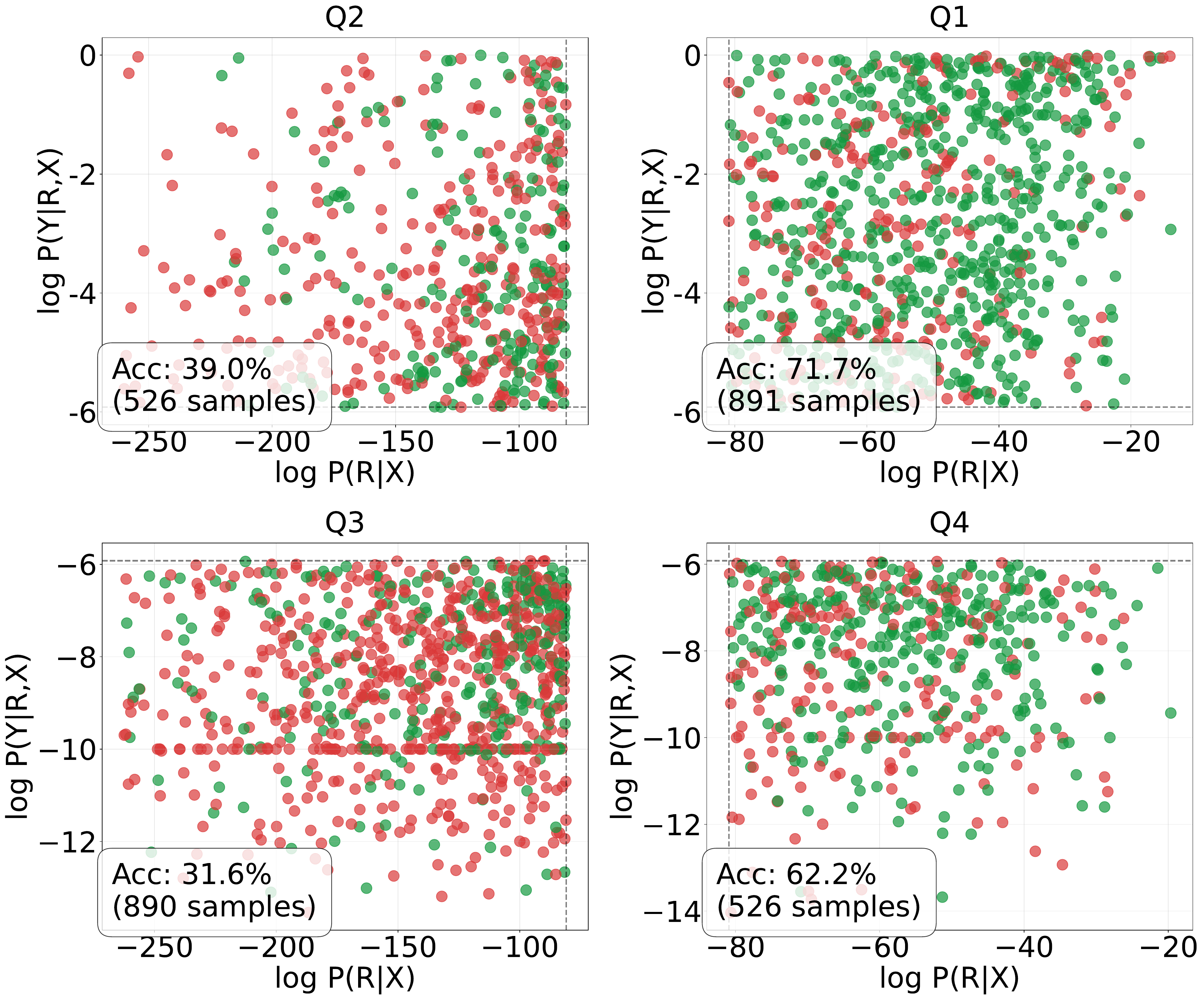}
    \caption{Quadrant-split representation with Q1–Q4 defined by median thresholds of reasoning and answer confidence. }
    \label{fig:quadrats_plane-llama-3.1-math500}
  \end{subfigure}}
  \caption{Information Plane visualisations of Llama-3.1-8B on the MATH500 dataset ($k=6$). 
  Green indicates correct answers, Red incorrect ones.}
  \label{fig:quadrats_subplot-llama-3.1-math500}
\end{figure*}

\begin{figure*}[htbp]
  \centering
    \scalebox{1.0}{
  \begin{subfigure}[b]{0.4\textwidth}
    \centering
    \includegraphics[width=\textwidth]{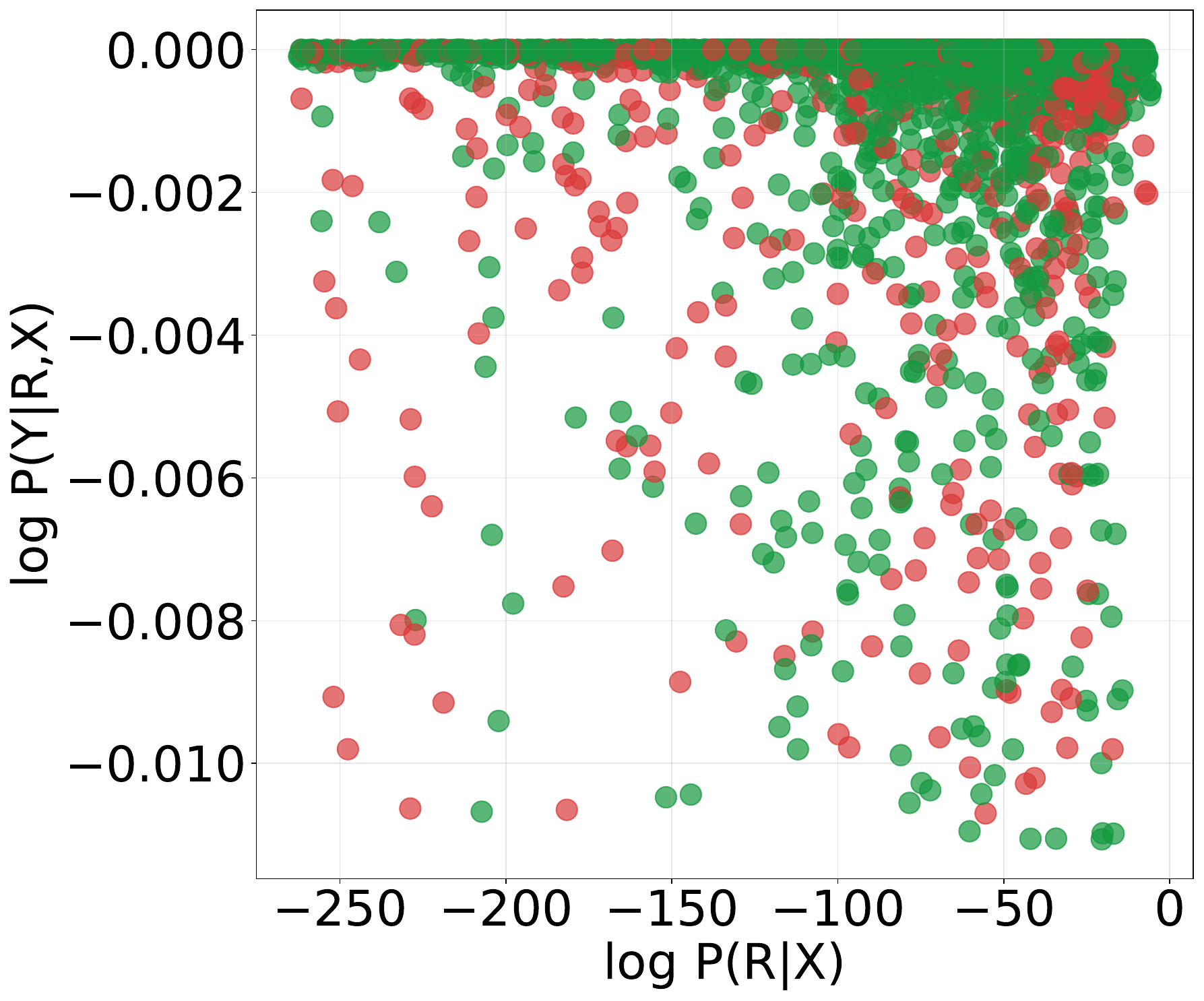}
    \caption{Continuous distribution without quadrant partitioning. }
    \label{fig:quadrats_split-qwen3-32b-math500}
  \end{subfigure}
  \hfill
  \begin{subfigure}[b]{0.4\textwidth}
    \centering
    \includegraphics[width=\textwidth]{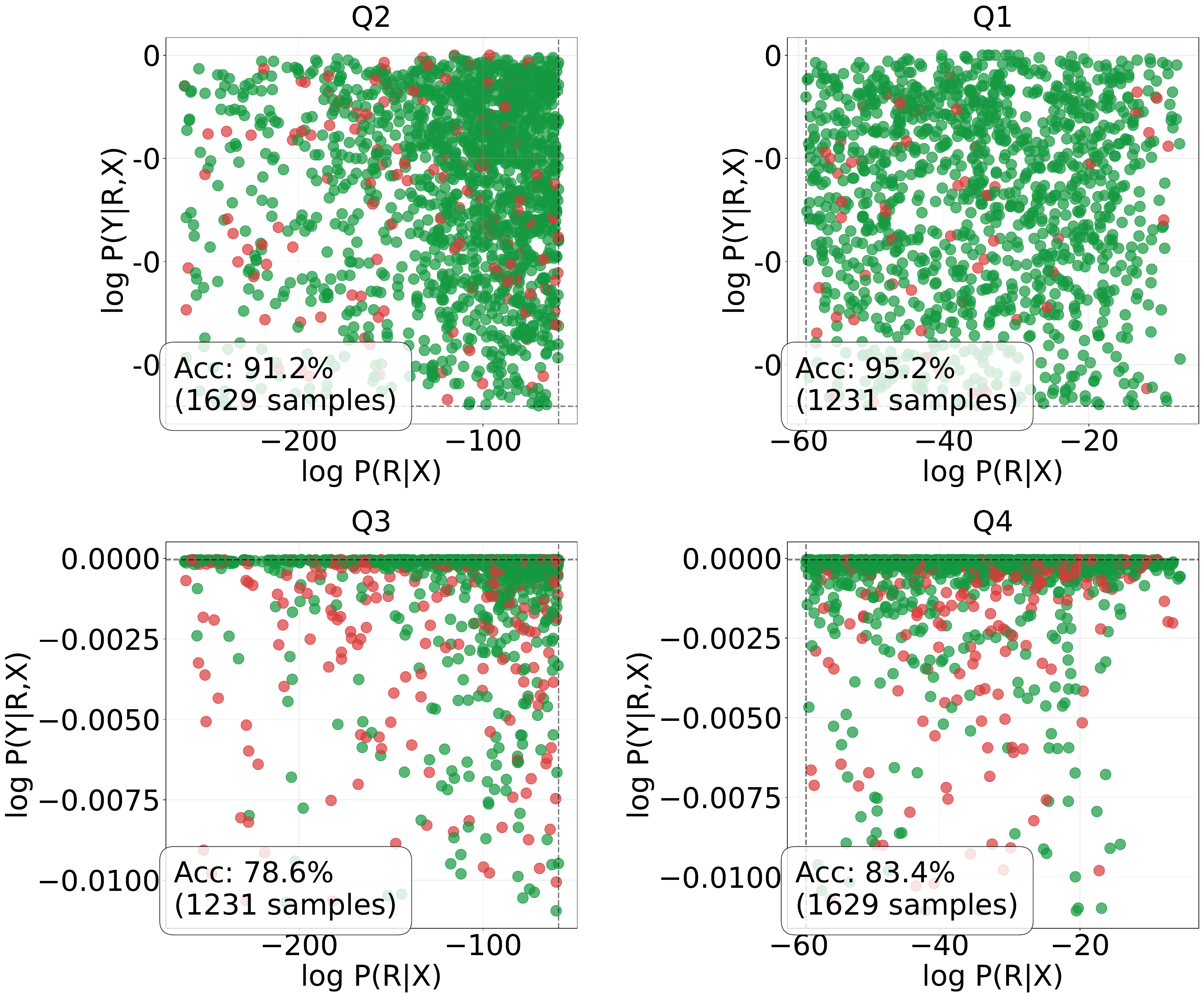}
    \caption{Quadrant-split representation with Q1–Q4 defined by median thresholds of reasoning and answer confidence. }
    \label{fig:quadrats_plane-qwen3-32b-math500}
  \end{subfigure}}
  \caption{Information Plane visualisations of Qwen3-8B on the MATH500 dataset ($k=6$). 
  Green indicates correct answers, Red incorrect ones.}
  \label{fig:quadrats_subplot-qwen3-32b-math500}
\end{figure*}

\begin{figure*}[htbp]
  \centering
    \scalebox{1.0}{
  \begin{subfigure}[b]{0.4\textwidth}
    \centering
    \includegraphics[width=\textwidth]{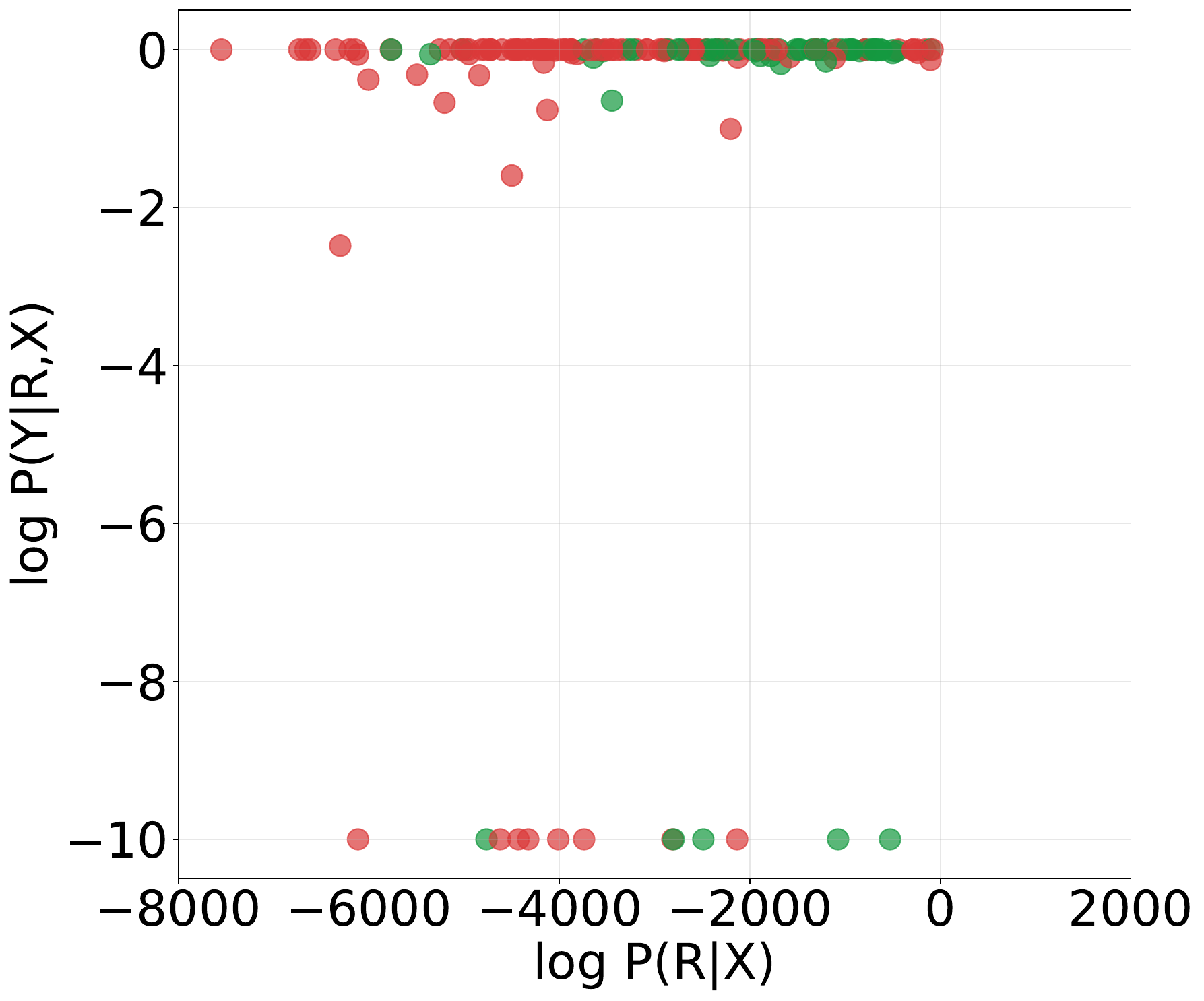}
    \caption{Continuous distribution without quadrant partitioning. }
    \label{fig:quadrats_split-DS-llamab-aime2024}
  \end{subfigure}
  \hfill
  \begin{subfigure}[b]{0.4\textwidth}
    \centering
    \includegraphics[width=\textwidth]{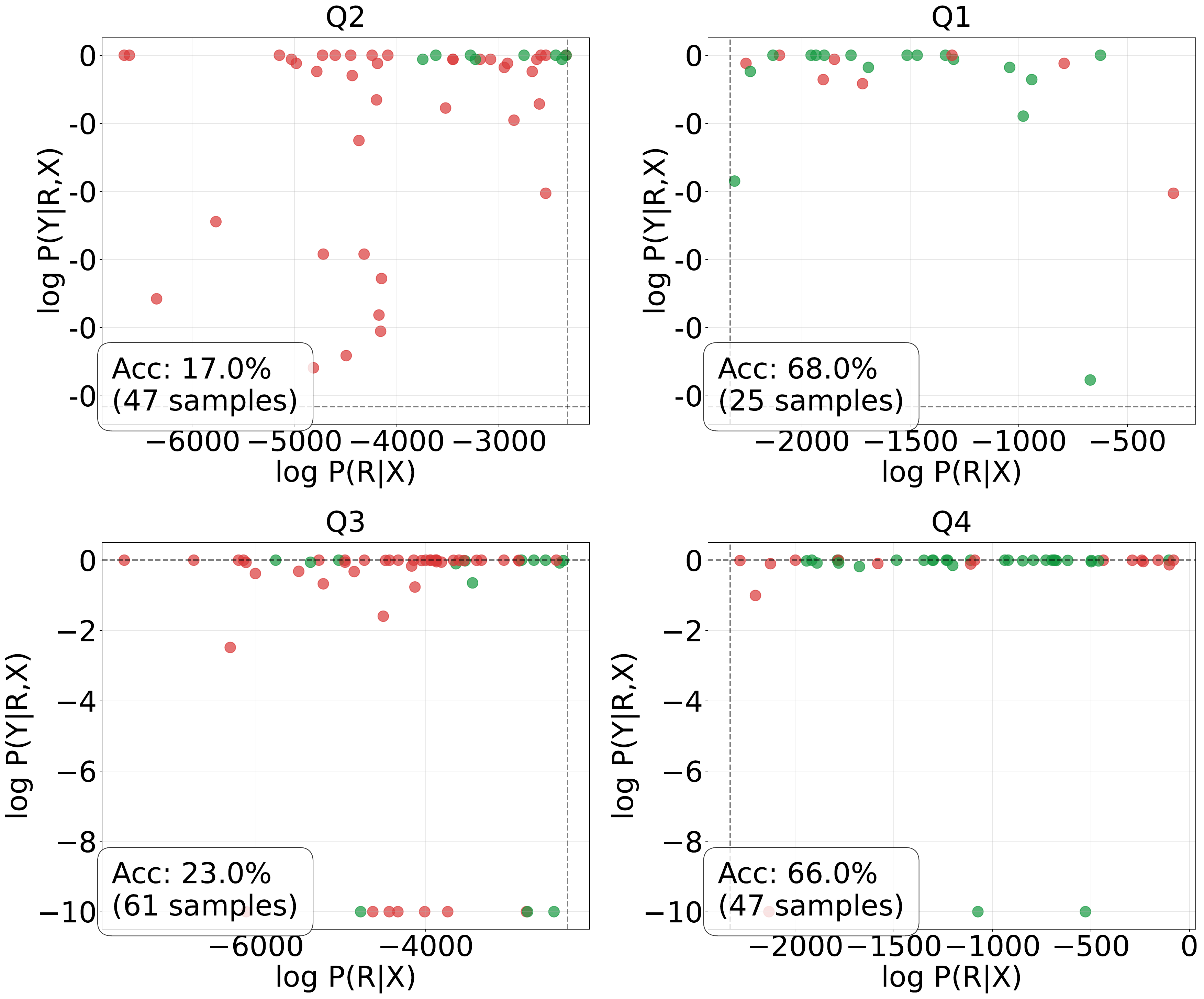}
    \caption{Quadrant-split representation with Q1–Q4 defined by median thresholds of reasoning and answer confidence. }
    \label{fig:quadrats_plane-DS-llama-aime2024}
  \end{subfigure}}
  \caption{Information Plane visualisations of DS-Distilled-Llama-8B on the AIME2024 dataset ($k=6$). 
  Green indicates correct answers, Red incorrect ones.}
  \label{fig:quadrats_subplot-DS-llama-aime2024}
\end{figure*}

\begin{figure*}[ht]
  \centering
    \scalebox{1.0}{
  \begin{subfigure}[b]{0.4\textwidth}
    \centering
    \includegraphics[width=\textwidth]{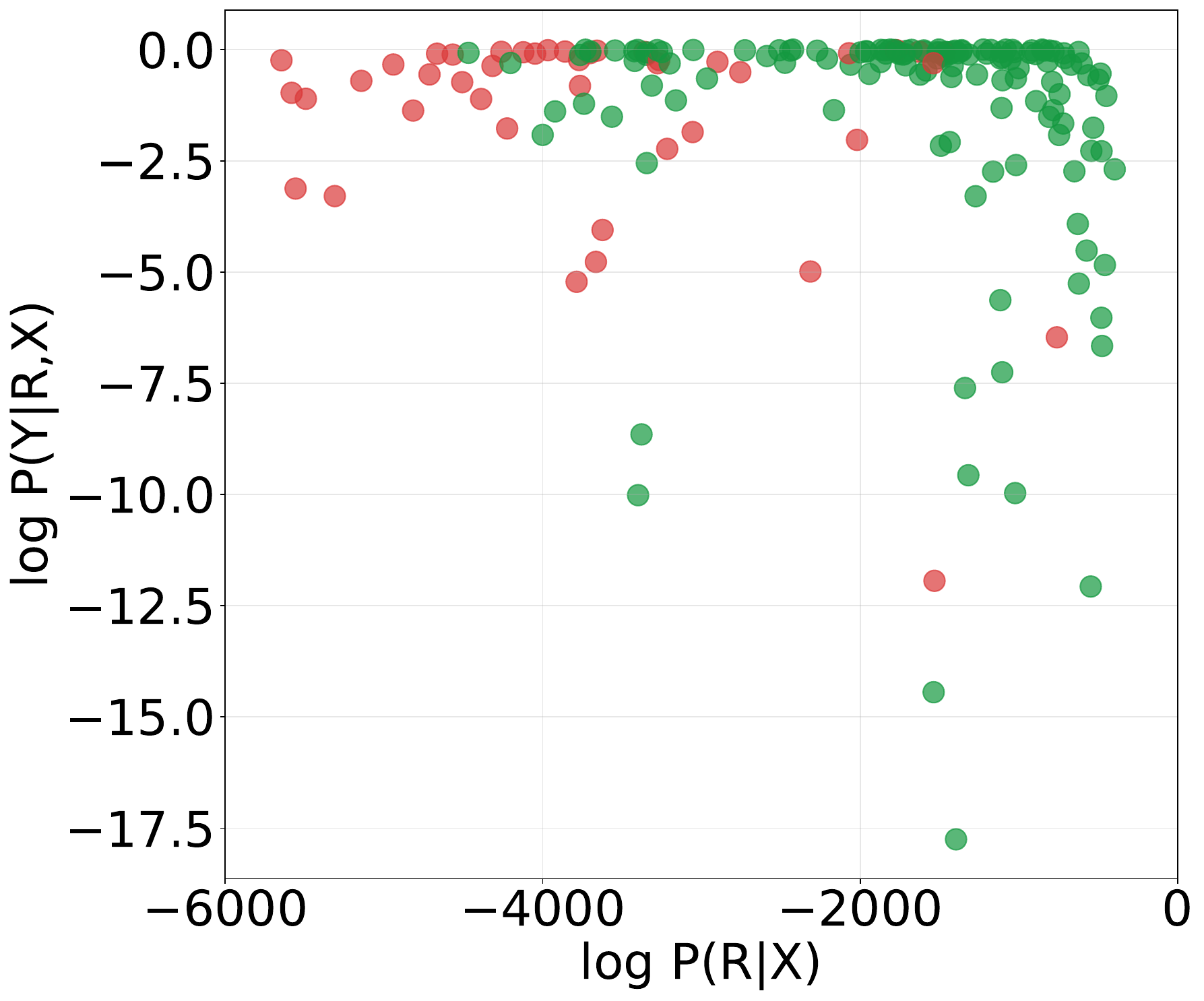}
    \caption{Continuous distribution without quadrant partitioning. }
    \label{fig:quadrats_split-qwen3-aime2024}
  \end{subfigure}
  \hfill
  
  \begin{subfigure}[b]{0.3\textwidth}
    \centering
    \includegraphics[width=\textwidth]{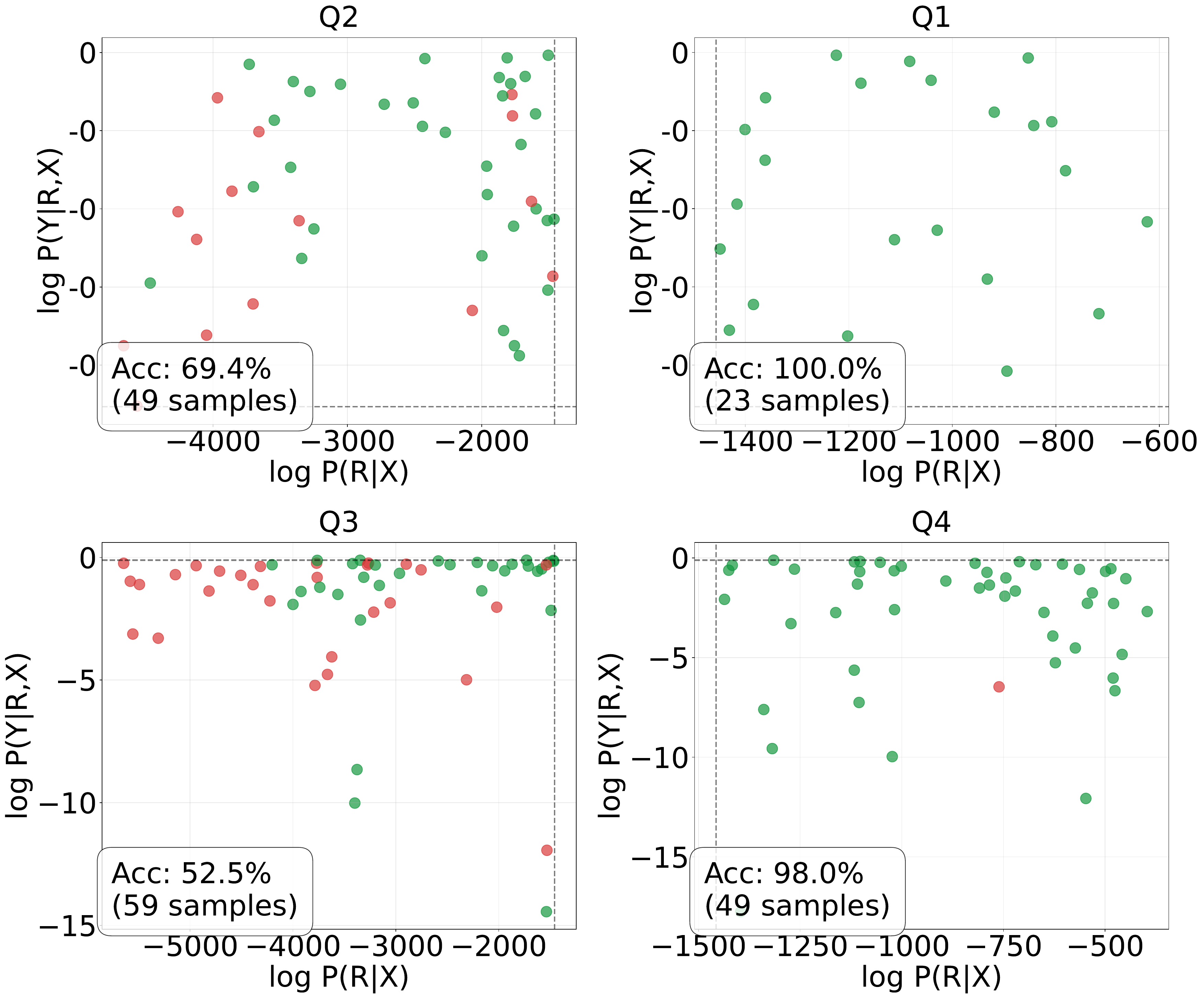}
    \caption{Quadrant-split representation with Q1–Q4 defined by median thresholds of reasoning and answer confidence. }
    \label{fig:quadrats_plane-qwen3-aime2024}
  \end{subfigure}}
  \caption{Information Plane visualisations of DS-Distilled-Llama-8B on the AIME2024 dataset ($k=6$). 
  Green indicates correct answers, Red incorrect ones.}
  \label{fig:quadrats_subplot-qwen3-aime2024}
\end{figure*}

\begin{figure*}[!ht]
  \centering
    \scalebox{1.0}{
  \begin{subfigure}[b]{0.47\textwidth}
    \centering
    \includegraphics[width=\textwidth]{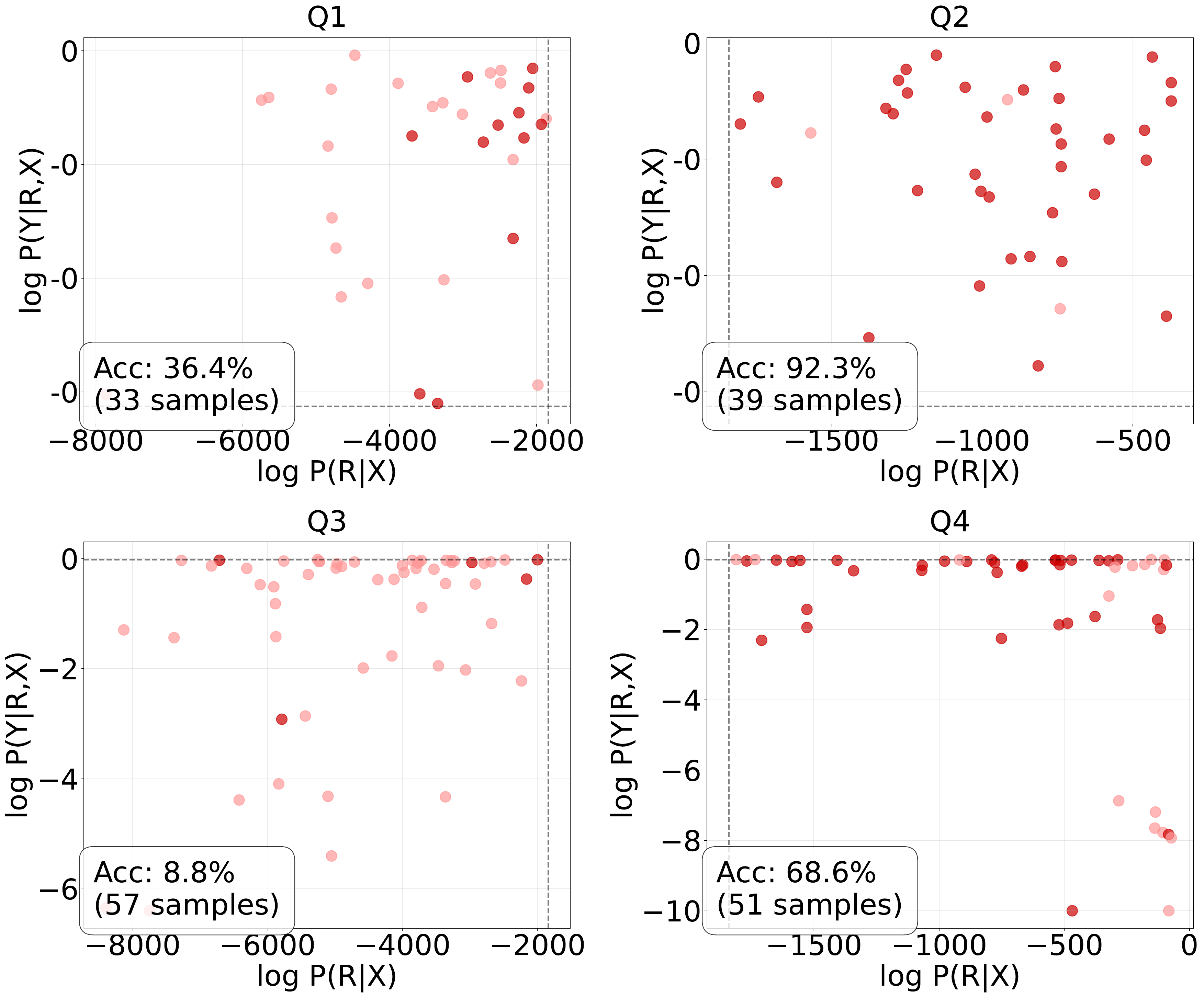}
    \caption{Quadrants defined by the 50th percentile (median)}
    \label{fig:info_plane_ds-distill-qwen}
  \end{subfigure}
  \hfill
  \begin{subfigure}[b]{0.47\textwidth}
    \centering
    \includegraphics[width=\textwidth]
    {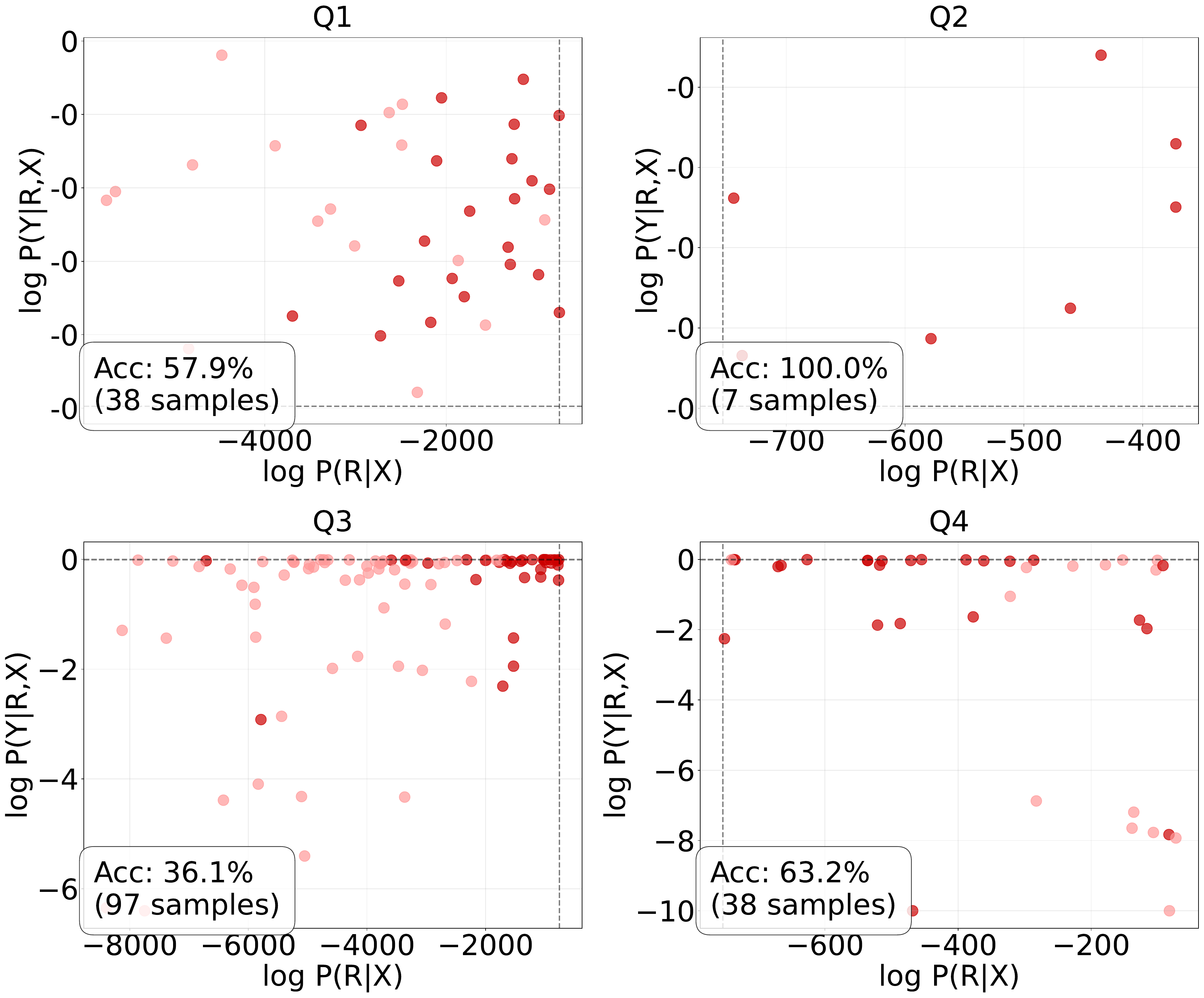}
    \caption{Quadrants defined by the 75th percentile. }
    \label{fig:info_plane_ds-distill-qwen-75c}
  \end{subfigure}}
  \caption{Effect of confidence thresholding on the Information Plane for DS-Distilled-Qwen-2.5-7B ($k=6$) on AIME2024. }
  \label{fig:info_plane_comparison}
\end{figure*}

In this section, we show all the models across datasets (GSM8K, MATH500 and AIME2024), which consist of a variety of difficulties. We observe a consistent pattern across \methodabbrev. In addition, the utility of our confidence metric extends to filtering for high-reliability answers. For GSM8K and MATH500, we use the median as our threshold with outliers removed, similar to~\S\ref{sec:empirical_justification}. However, as for AIME2024, as the instance is similar, we include all the instances including the outliers, and set the threshold to 60\% for both x and y-axis. We show results on GSM8K in~\cref{fig:quadrats_subplot-llama-3.1}--%
\ref{fig:quadrats_subplot-qwen3-32b-gsm8k}. Similarly, results on MATH500 are provided in Figures~\ref{fig:quadrats_subplot-llama-3.1-math500}--%
\ref{fig:quadrats_subplot-qwen3-32b-math500}. We provide results on AIME 2024 in Figures~\ref{fig:quadrats_subplot-DS-llama-aime2024}--%
\ref{fig:quadrats_subplot-qwen3-aime2024}. In addition, we also show results of using 75th percentile as the threshold in \cref{fig:info_plane_comparison}. As shown in \cref{fig:info_plane_comparison}, increasing the confidence thresholds from the median to the 75th percentile isolates a region in the Information Plane with significantly higher accuracy, effectively identifying the most trustworthy solutions.

\begin{table*}[t]
\centering
\small
\renewcommand{\arraystretch}{0.85}
\begin{tabular}{l c c c c c}
\toprule
\textbf{Method} & \textbf{Gemma-2-9B} & \textbf{Qwen3-8B} & \textbf{Llama-3.1-70B} & \textbf{DS-Qwen-7B} & \textbf{Average} \\
\midrule
Skywork-Reward-V2-Llama-3.1-8B & $ 48.40$ & $74.20$ & $62.60$ & $74.80$ & $65.00$ \\
LMUnit-qwen2.5-72B             & $ 48.40$ & $ 75.40$ & $ 64.00$ & $ 75.00$ & $ 65.70$ \\
\rowcolor{blue!10} \methodabbrev & $46.53$ & $73.67$ & $63.67$ & $74.00$ & $64.47$ \\
\bottomrule
\end{tabular}
\caption{\textbf{Performance comparison of \methodabbrev against trained verifiable reward models on the MATH500 dataset (\%).} Despite being a zero-shot, training-free approach, \methodabbrev achieves highly competitive performance compared to state-of-the-art reward models trained on large preference datasets.}
\label{tab:math500_performance}
\end{table*}

\begin{table*}[t]
\centering
\small
\renewcommand{\arraystretch}{0.85}
\begin{tabular}{l c c c c c}
\toprule
\textbf{Method} & \textbf{Gemma-2-9B} & \textbf{Qwen3-8B} & \textbf{Llama-3.1-70B} & \textbf{DS-Qwen-7B} & \textbf{Average} \\
\midrule
Skywork-Reward-V2-Llama-3.1-8B & $ 89.60$ & $93.17$ & $89.83$ & $ 89.37$ & $90.49$ \\
LMUnit-qwen2.5-72B             & $88.84$ & $ 93.55$ & $94.53$ & $ 89.37$ & $ 91.57$ \\
\rowcolor{blue!10} \methodabbrev & $88.66$ & $92.33$ & $ 94.58$ & $88.18$ & $90.94$ \\
\bottomrule
\end{tabular}
\caption{\textbf{Performance comparison of \methodabbrev against trained verifiable reward models on the GSM8K dataset (\%).} Similar to the MATH500 results, the zero-shot, training-free \methodabbrev demonstrates highly competitive performance against trained reward models.}
\label{tab:gsm8k_performance}
\end{table*}

\subsection{Statistical Tests}
\label{sec:statsmodel}

In this part, we present detailed results of the statistical tests described in~\S\ref{sec:empirical_justification}. We conduct these tests on the MATH500 dataset, with all results reported in~\cref{tab:math500-stats}. 

The \emph{terms} reported in the table correspond to reasoning confidence and answer confidence. The \emph{t-statistic} measures the degree of separation between correct and incorrect responses. The associated \emph{p-value} confirms that both confidence metrics significantly contribute to this distinction rather than arising from random variation. The \emph{U-statistic} provides a non-parametric validation that the scores for correct responses are stochastically distinct from those for incorrect ones. The \emph{Cohen’s d} quantifies the magnitude of this effect size. The \emph{mean (C/I)} indicates the directionality of the relationship, showing that the average log-probabilities are consistently higher for correct responses than for incorrect ones.

To observe whether the ability of the base model itself influence the performance of PiCSAR, we perform a one-way ANOVA testing whether the choice of base model affects PICSAR's improvement over the best competing baseline across all benchmarks. The result $F=1.95$, $p=0.15$ indicates no statistically significant difference, confirming that PICSAR provides consistent gains regardless of the underlying model. This is supported by our findings in~\ref{sec:decoupled_confidence}, where we show that the answer-confidence component can be evaluated by a relatively smaller model without degrading selection quality, removing the requirement for the scoring model to match the generator in scale or architecture.

\subsection{In Comparison between PiCSAR and Trained Verifiable Rewards}
\label{sec:rm_comparison}
In this section, we compare PiCSAR with trained verifiable rewards. We evaluated PiCSAR with two best performing Reward Models from RewardBench~\citep{lambert-etal-2025-rewardbench}, \textit{Skywork-Reward-V2-Llama-3.1-8B} and \textit{LMUnit-qwen2.5-72B}.

Based on Table~\ref{tab:math500_performance} and~\ref{tab:gsm8k_performance}, PiCSAR achieves parity with (and occasionally outperforms) \textit{Skywork-Reward-V2-Llama-3.1-8B}, a model explicitly trained on massive preference datasets, despite PiCSAR being zero-shot and training-free. While larger RMs (\ie \textit{LMUnit-qwen2.5-72B}) generally perform better (avg ~+1-2\%), the fact that PiCSAR is competitive with trained verifiers confirms its high effectiveness, especially due to its zero additional training cost.

\section{Dataset Details}

\begin{itemize}

    \item SVAMP~\cite{patel2021nlp}:
    \url{https://github.com/arkilpatel/SVAMP}, License: \href{https://github.com/arkilpatel/SVAMP/blob/main/LICENSE}{SVAMP License}

    \item GSM8K~\cite{cobbe2021training}: \url{https://huggingface.co/datasets/openai/gsm8k}, License: \href{https://github.com/openai/grade-school-math/blob/master/LICENSE}{GSM8K License}

    \item MATH~\cite{hendrycks2021measuring}: \url{https://huggingface.co/datasets/HuggingFaceH4/MATH-500}, License: \href{https://github.com/hendrycks/math/blob/main/LICENSE}{MATH License}

    \item GPQA-Diamond~\cite{rein2024gpqa}: \url{https://huggingface.co/datasets/Idavidrein/gpqa/viewer/gpqa_diamond/train}, License: \href{https://github.com/idavidrein/gpqa/blob/main/LICENSE}{GPQA License}
    
    \item TheoremQA~\cite{chen2023theoremqa}: \url{https://huggingface.co/datasets/TIGER-Lab/TheoremQA}, License: \href{https://github.com/TIGER-AI-Lab/TheoremQA/blob/main/LICENSE}{GPQA License}

    \item AIME-2024: \url{https://huggingface.co/datasets/Maxwell-Jia/AIME_2024}, License: \href{https://huggingface.co/datasets/Maxwell-Jia/AIME_2024}{AIME-2024 License}

    \item AIME-2025: \url{https://huggingface.co/datasets/opencompass/AIME2025}, License: \href{https://huggingface.co/datasets/opencompass/AIME2025}{AIME-2024 License}

\end{itemize}

\onecolumn
\subsection*{GSM8K}

\begin{figure*}[htbp]
  \centering
  \scalebox{1.0}{
  \begin{subfigure}[b]{0.44\textwidth}
    \centering
    \includegraphics[width=\textwidth]{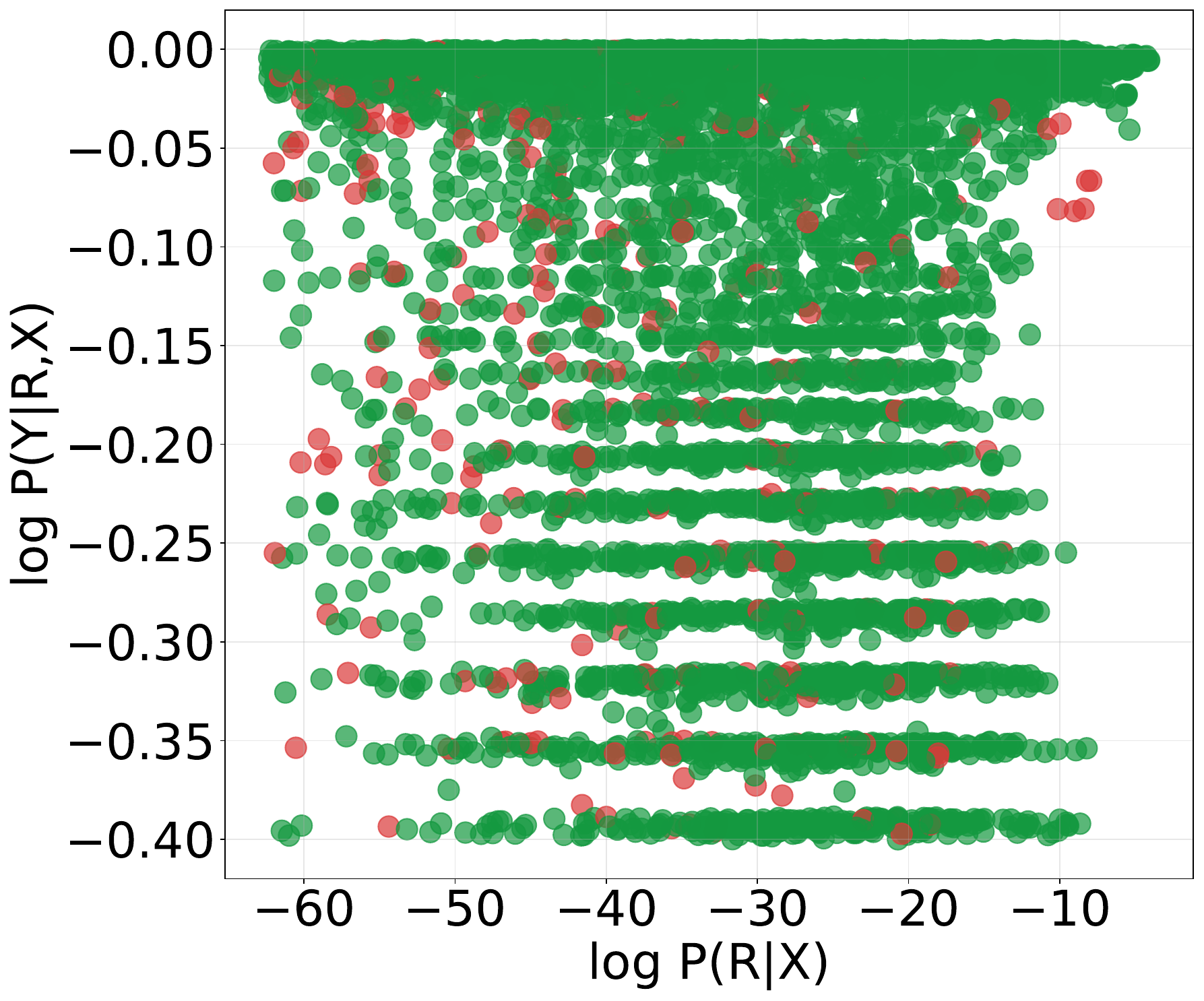}
    \caption{Continuous distribution without quadrant partitioning. }
    \label{fig:quadrats_split-llama-3.1-70b-gsm8k}
  \end{subfigure}
  \hfill
  \begin{subfigure}[b]{0.45\textwidth}
    \centering
    \includegraphics[width=\textwidth]{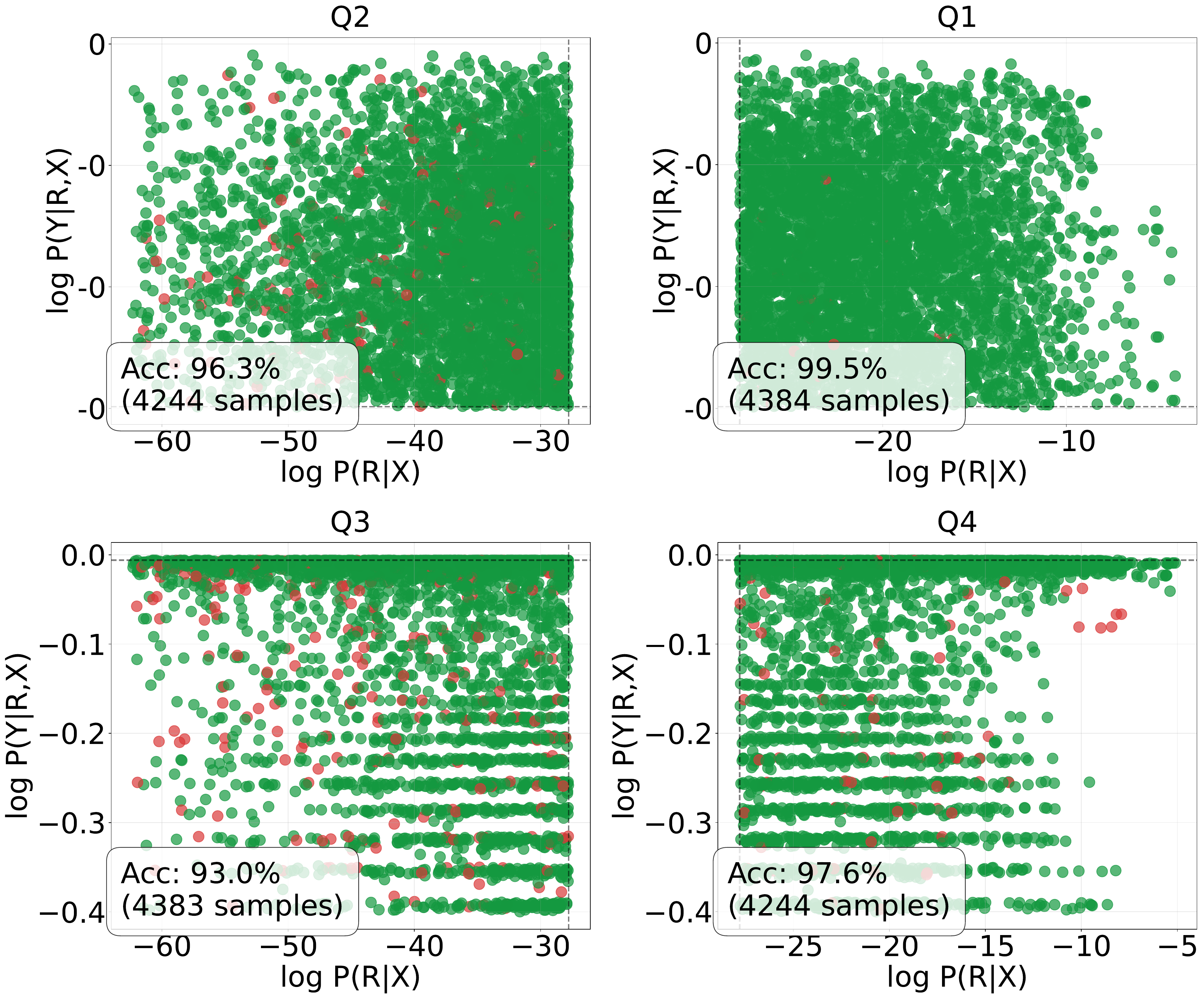}
    \caption{Quadrant-split representation with Q1–Q4 defined by median thresholds of reasoning and answer confidence. }
    \label{fig:quadrats_plane-llama-3.1-70b-gsm8k}
  \end{subfigure}}
  \caption{Information Plane visualisations of Gemma-2-9B on the GSM8K dataset ($k=6$). 
  Green indicates correct answers, Red incorrect ones.}
  \label{fig:quadrats_subplot-llama-3.1-70b-gsm8k}
\end{figure*}

\begin{figure*}[htbp]
  \centering
\scalebox{1.0}{
  \begin{subfigure}[b]{0.4\textwidth}
    \centering
    \includegraphics[width=\textwidth]{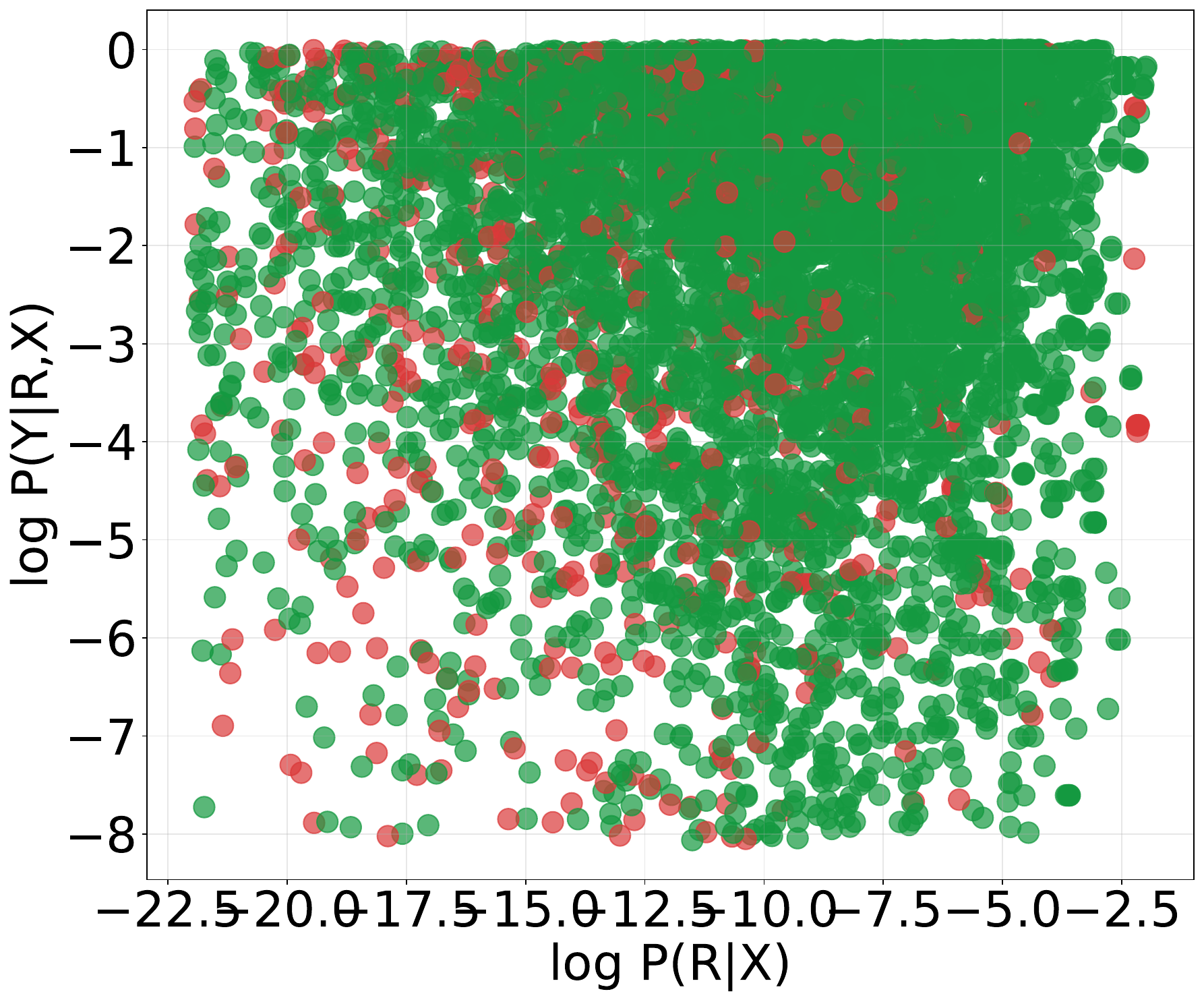}
    \caption{Continuous distribution without quadrant partitioning. }
    \label{fig:quadrats_split-gemma-2-gsm8k}
  \end{subfigure}
  \hfill
  \begin{subfigure}[b]{0.4\textwidth}
    \centering
    \includegraphics[width=\textwidth]{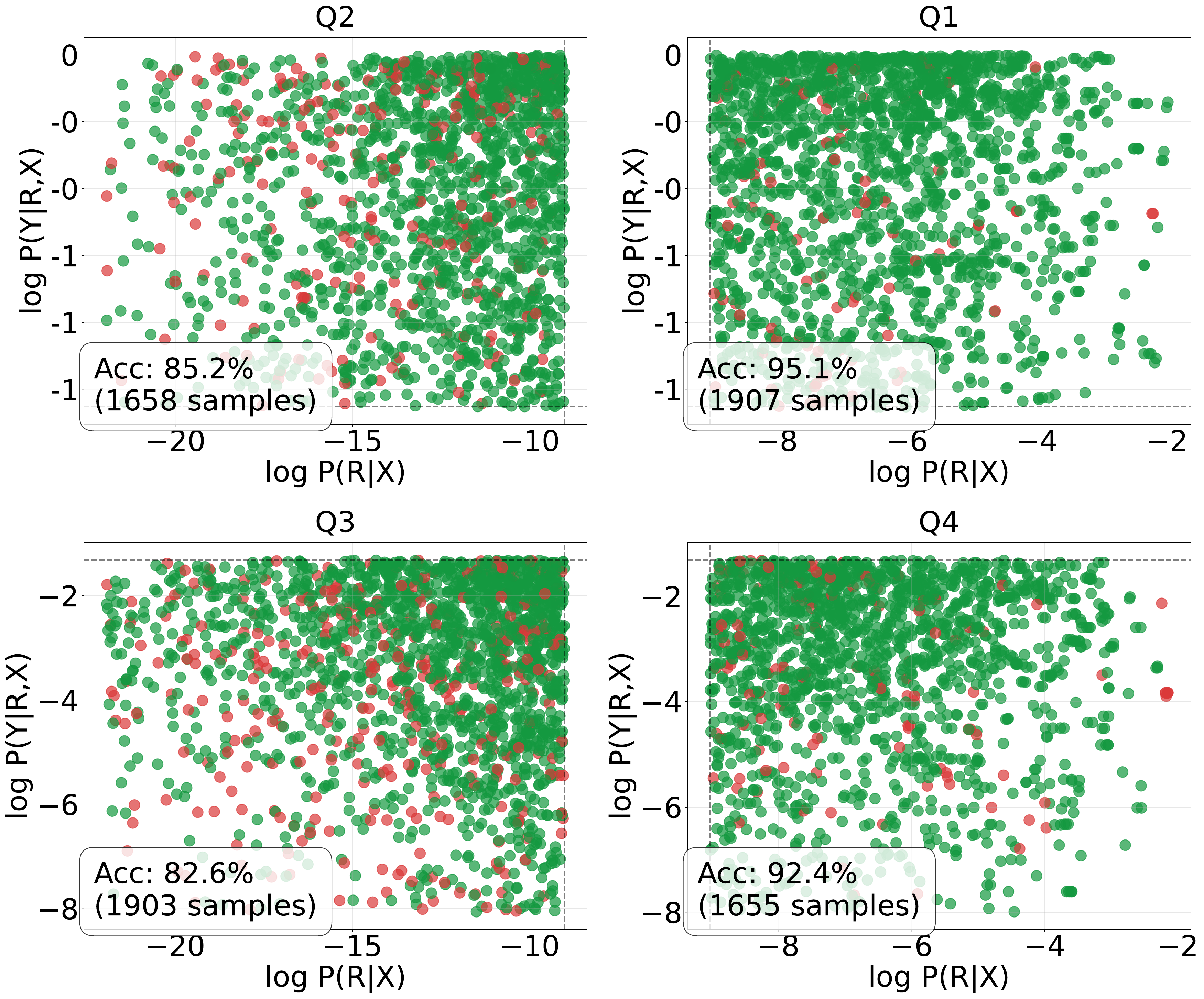}
    \caption{Quadrant-split representation with Q1–Q4 defined by median thresholds of reasoning and answer confidence. }
    \label{fig:quadrats_plane-gemma-2-gsm8k}
  \end{subfigure}}
  \caption{Information Plane visualisations of Gemma-2-9B on the GSM8K dataset ($k=6$). 
  Green indicates correct answers, Red incorrect ones.}
  \label{fig:quadrats_subplot-gemma-2-gsm8k}
\end{figure*}

\begin{figure*}[htbp]
  \centering
    \scalebox{1.0}{
  \begin{subfigure}[b]{0.4\textwidth}
    \centering
    \includegraphics[width=\textwidth]{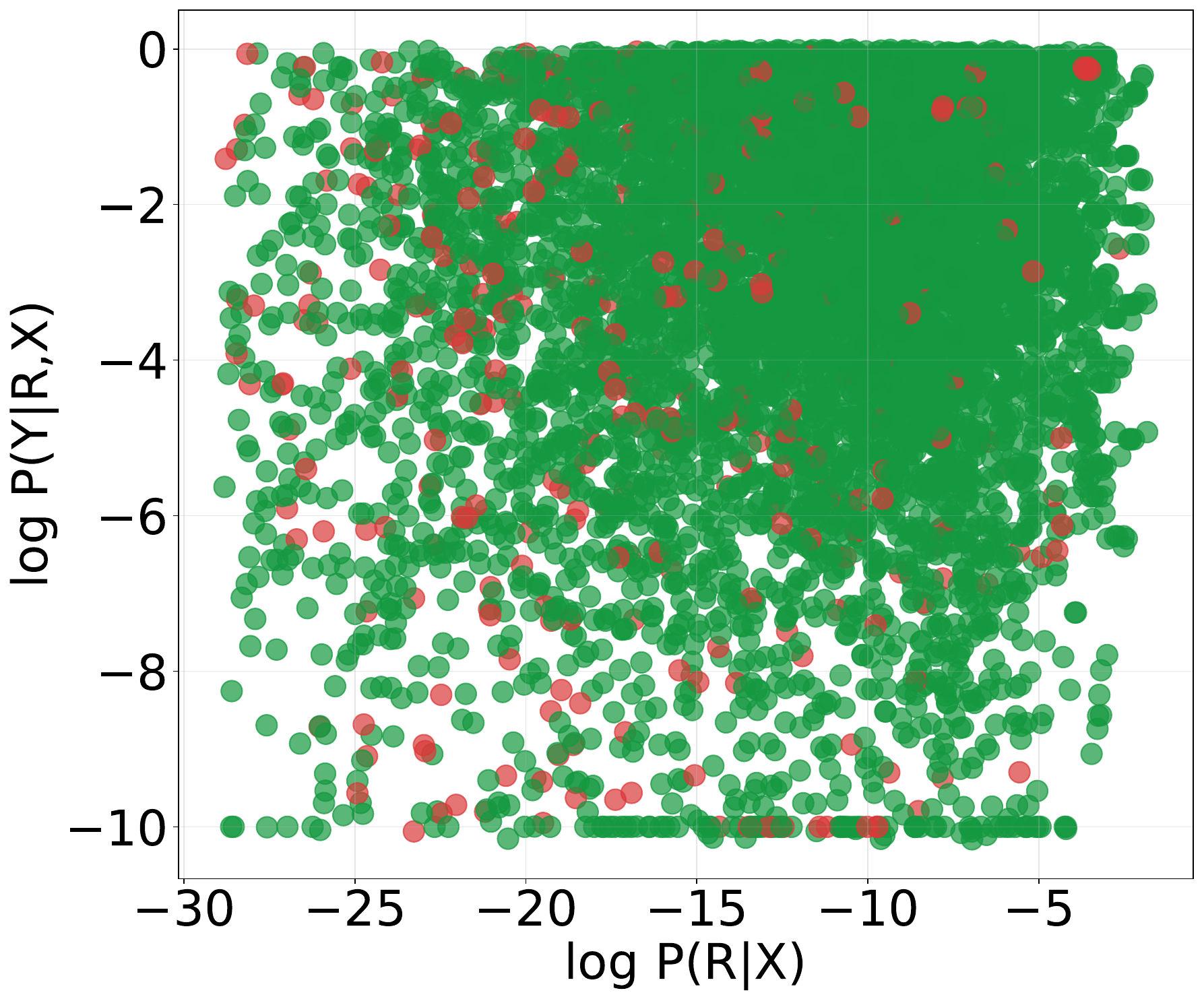}
    \caption{Continuous distribution without quadrant partitioning. }
    \label{fig:quadrats_split-qwen3-8b-gsm8k}
  \end{subfigure}
  \hfill
  \begin{subfigure}[b]{0.4\textwidth}
    \centering
    \includegraphics[width=\textwidth]{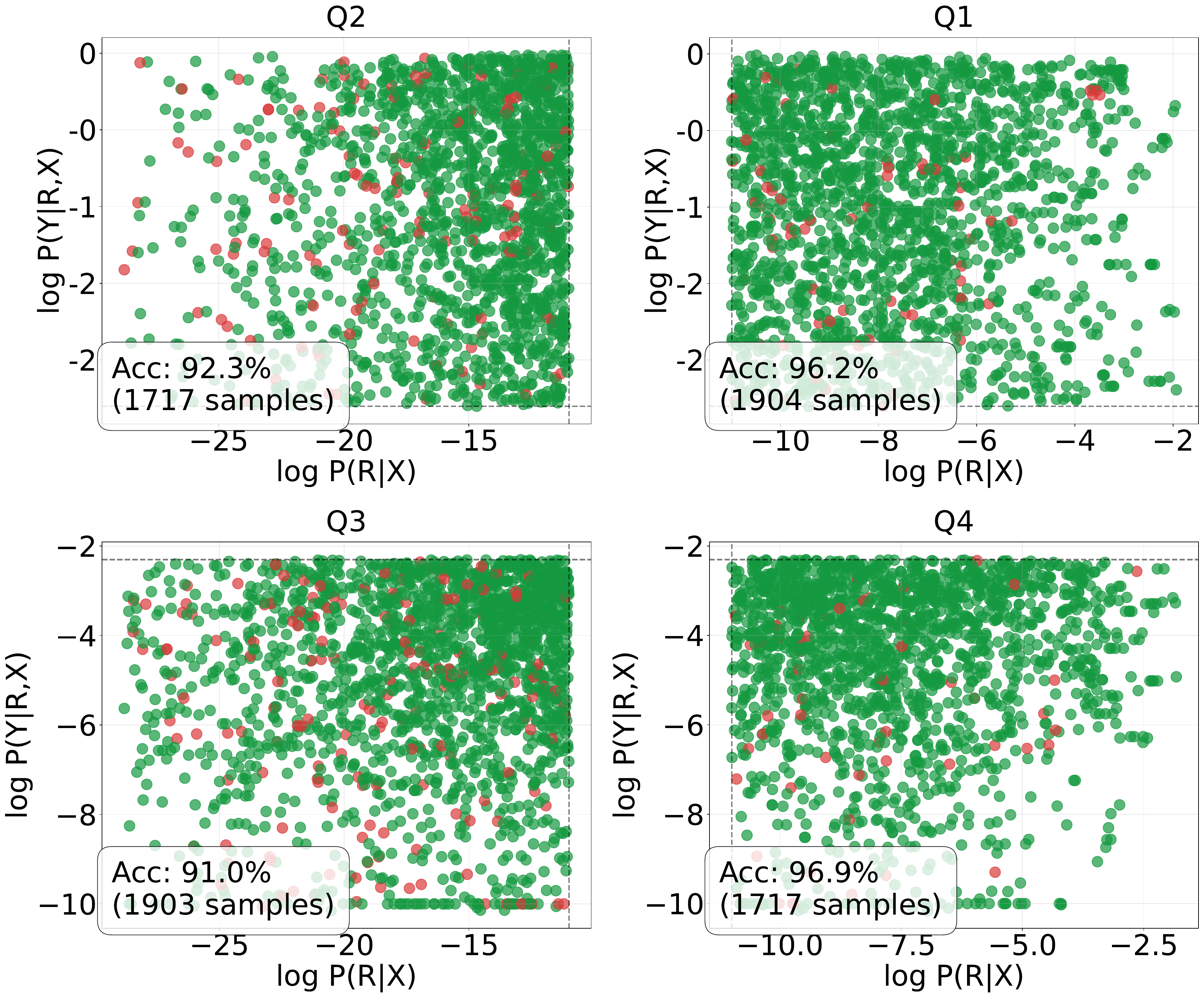}
    \caption{Quadrant-split representation with Q1–Q4 defined by median thresholds of reasoning and answer confidence. }
    \label{fig:quadrats_plane-qwen3-8b-gsm8k}
  \end{subfigure}}
  \caption{Information Plane visualisations of Gemma-2-9B on the GSM8K dataset ($k=6$). 
  Green indicates correct answers, Red incorrect ones.}
  \label{fig:quadrats_subplot-qwen3-8b-gsm8k}
\end{figure*}

\newpage
\subsection*{MATH500}

\begin{figure*}[htbp]
  \centering
    \scalebox{1.0}{
  \begin{subfigure}[b]{0.45\textwidth}
    \centering
    \includegraphics[width=\textwidth]{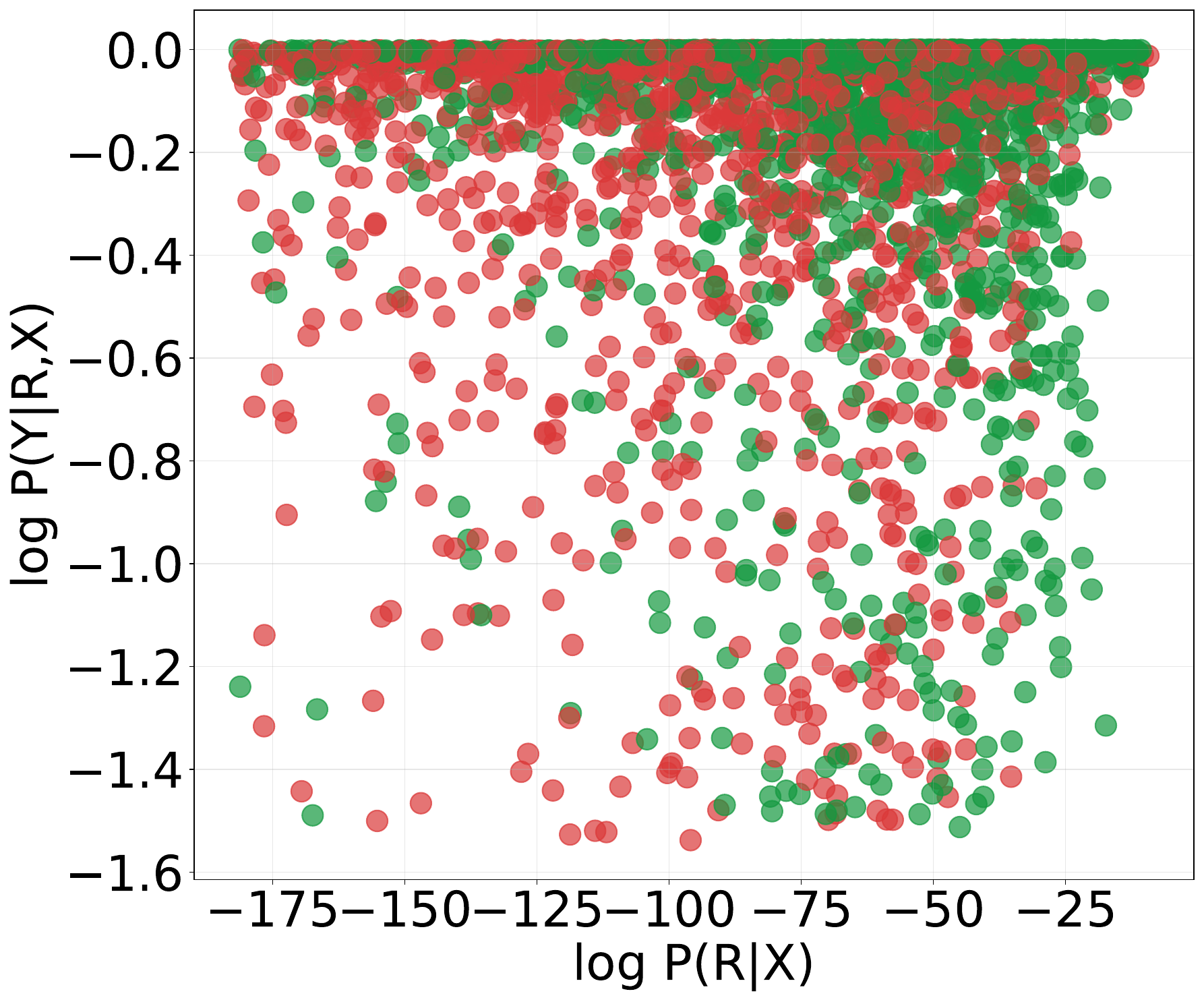}
    \caption{Continuous distribution without quadrant partitioning. }
    \label{fig:quadrats_split-llama-3.1-70b-math500}
  \end{subfigure}
  \hfill
  \begin{subfigure}[b]{0.45\textwidth}
    \centering
    \includegraphics[width=\textwidth]{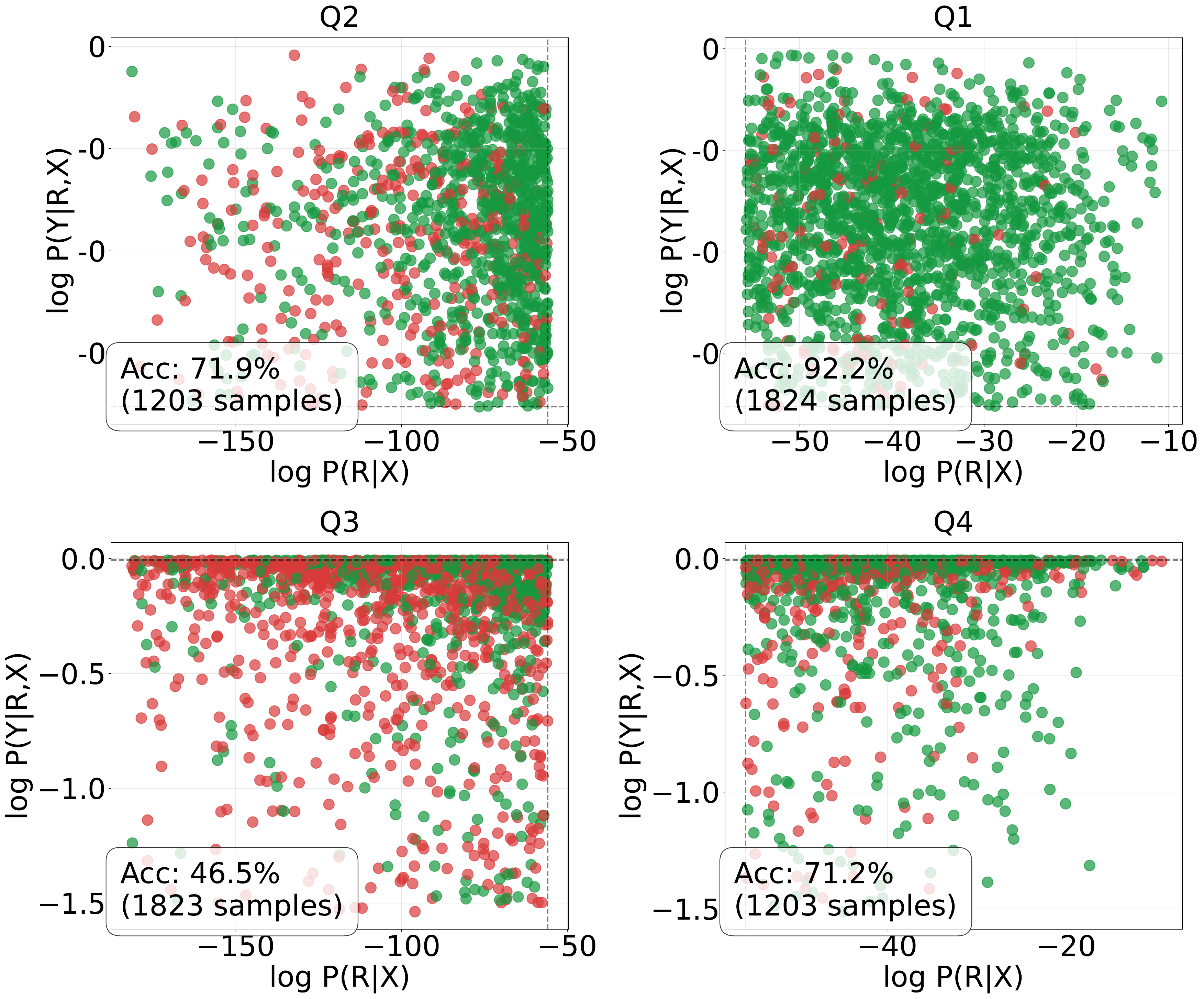}
    \caption{Quadrant-split representation with Q1–Q4 defined by median thresholds of reasoning and answer confidence. }
    \label{fig:quadrats_plane-llama-3.1-70b-math500}
  \end{subfigure}}
  \caption{Information Plane visualisations of Llama-3.1-70B on the MATH500 dataset ($k=6$). 
  Green indicates correct answers, Red incorrect ones.}
  \label{fig:quadrats_subplot-llama-3.1-70b-math500}
\end{figure*}



\begin{figure*}[htbp]
  \centering
    \scalebox{1.0}{
  \begin{subfigure}[b]{0.4\textwidth}
    \centering
    \includegraphics[width=\textwidth]{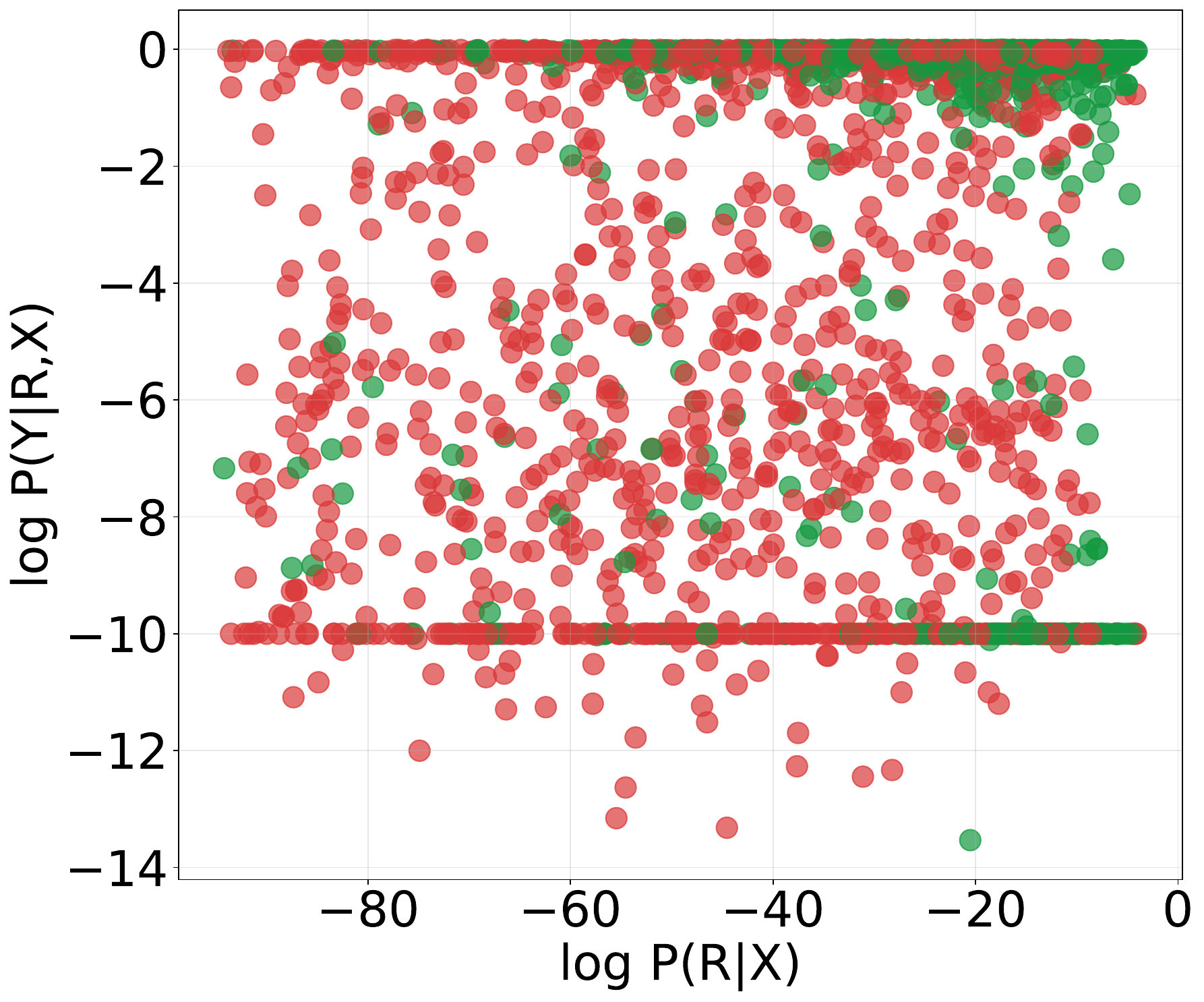}
    \caption{Continuous distribution without quadrant partitioning. }
    \label{fig:quadrats_split-gemma-2-9b-math500}
  \end{subfigure}
  \hfill
  \begin{subfigure}[b]{0.4\textwidth}
    \centering
    \includegraphics[width=\textwidth]{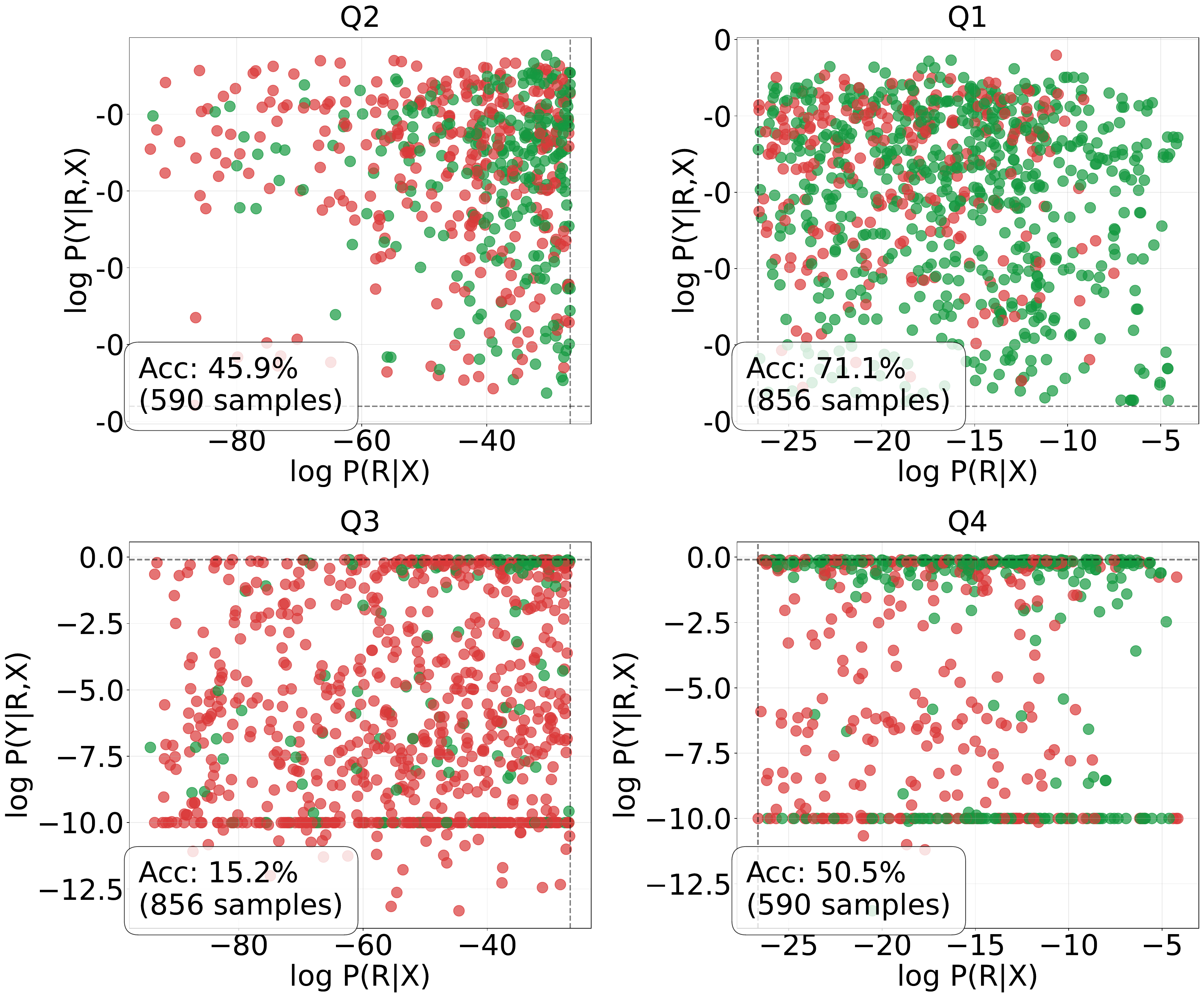}
    \caption{Quadrant-split representation with Q1–Q4 defined by median thresholds of reasoning and answer confidence. }
    \label{fig:quadrats_plane-gemma-2-9b-math500}
  \end{subfigure}}
  \caption{Information Plane visualisations of Gemma-2-9B on the MATH500 dataset ($k=6$). 
  Green indicates correct answers, Red indicates incorrect ones.}
  \label{fig:quadrats_subplot-gemma-2-9b-math500}
\end{figure*}

\begin{figure*}[ht]
  \centering
    \scalebox{1.0}{
  \begin{subfigure}[b]{0.4\textwidth}
    \centering
    \includegraphics[width=\textwidth]{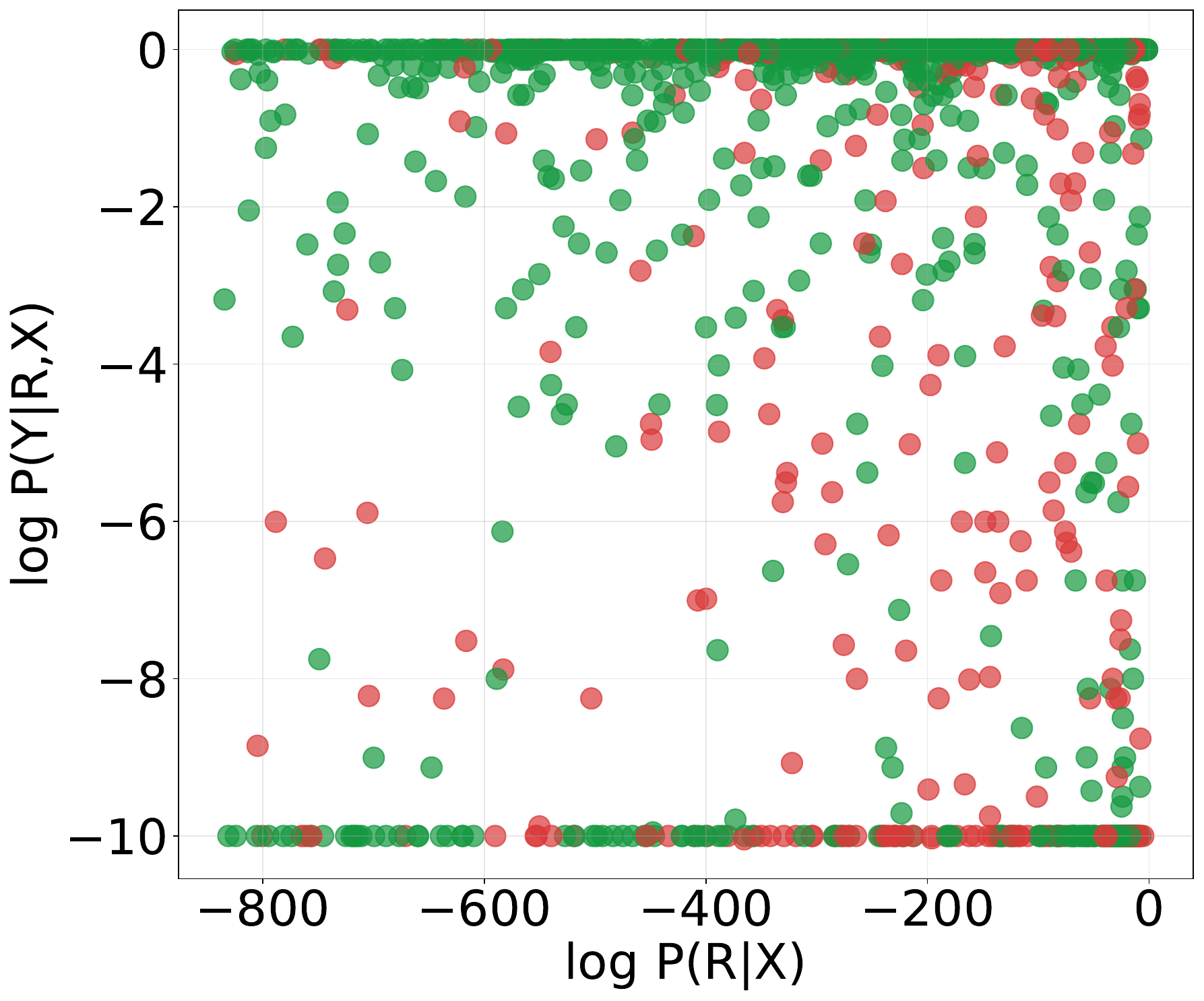}
    \caption{Continuous distribution without quadrant partitioning. }
    \label{fig:quadrats_split-qwen3-8b-math500}
  \end{subfigure}
  \hfill
  \begin{subfigure}[b]{0.4\textwidth}
    \centering
    \includegraphics[width=\textwidth]{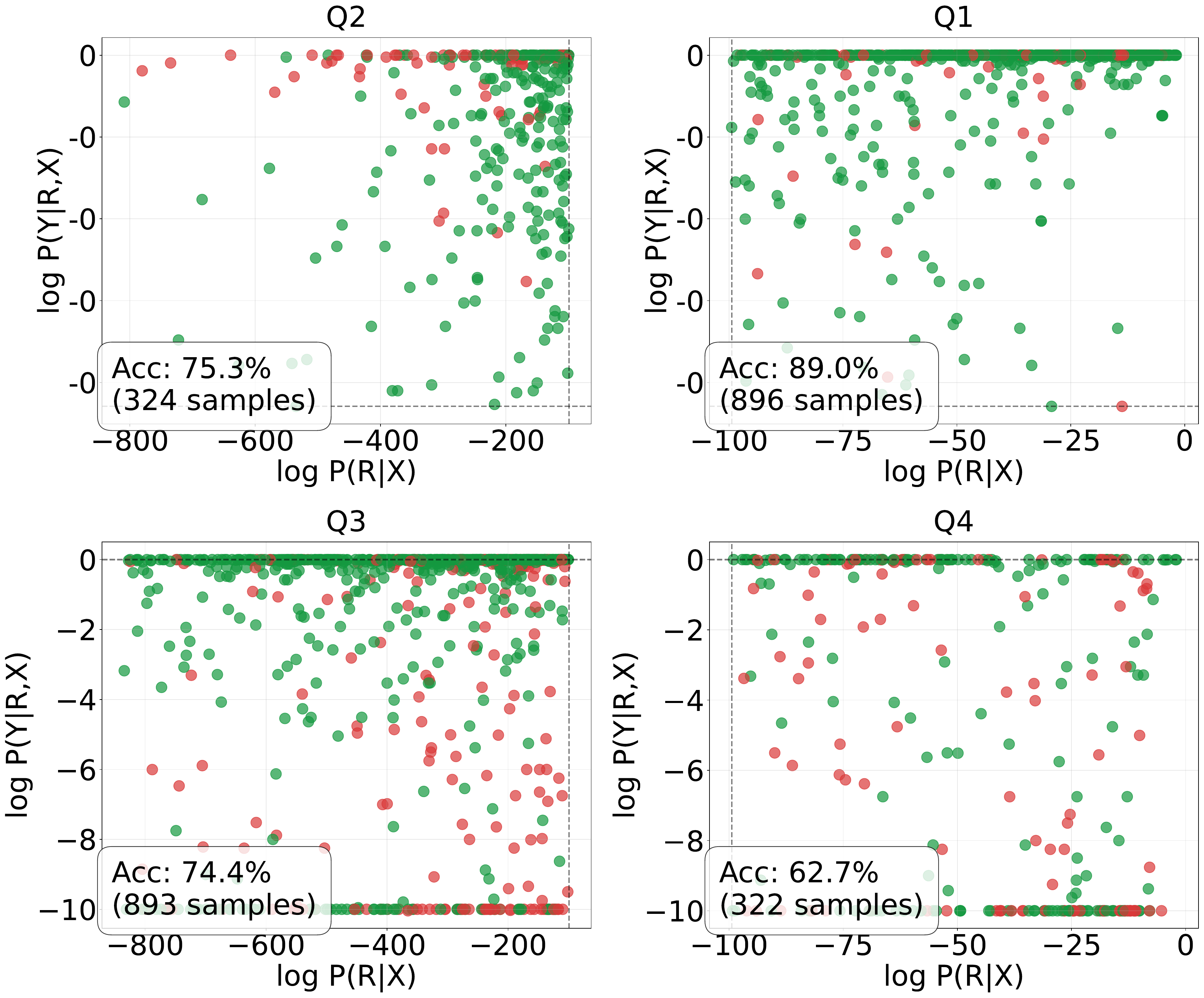}
    \caption{Quadrant-split representation with Q1–Q4 defined by median thresholds of reasoning and answer confidence. }
    \label{fig:quadrats_plane-qwen3-8b-math500}
  \end{subfigure}}
  \caption{Information Plane visualisations of Qwen3-8B on the MATH500 dataset ($k=6$). 
  Green indicates correct answers, Red incorrect ones.}
  \label{fig:quadrats_subplot-qwen3-8b-math500}
\end{figure*}

\clearpage

\subsection*{AIME2024}

\begin{figure*}[ht]
  \centering
    \scalebox{1.0}{
  \begin{subfigure}[b]{0.4\textwidth}
    \centering
    \includegraphics[width=\textwidth]{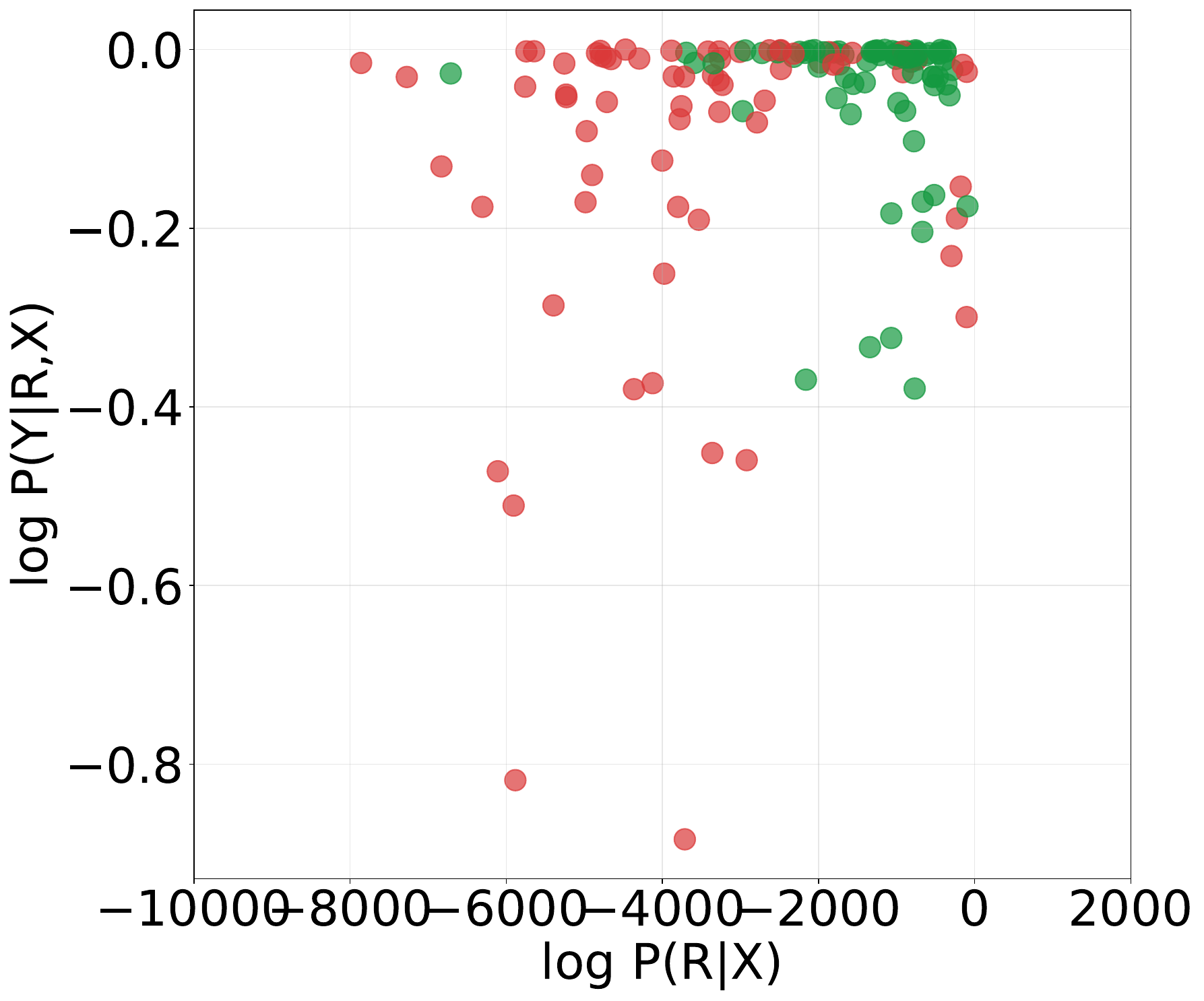}
    \caption{Continuous distribution without quadrant partitioning. }
    \label{fig:quadrats_split-DS-qwen-aime2024}
  \end{subfigure}
  \hfill
  \begin{subfigure}[b]{0.3\textwidth}
    \centering
    \includegraphics[width=\textwidth]{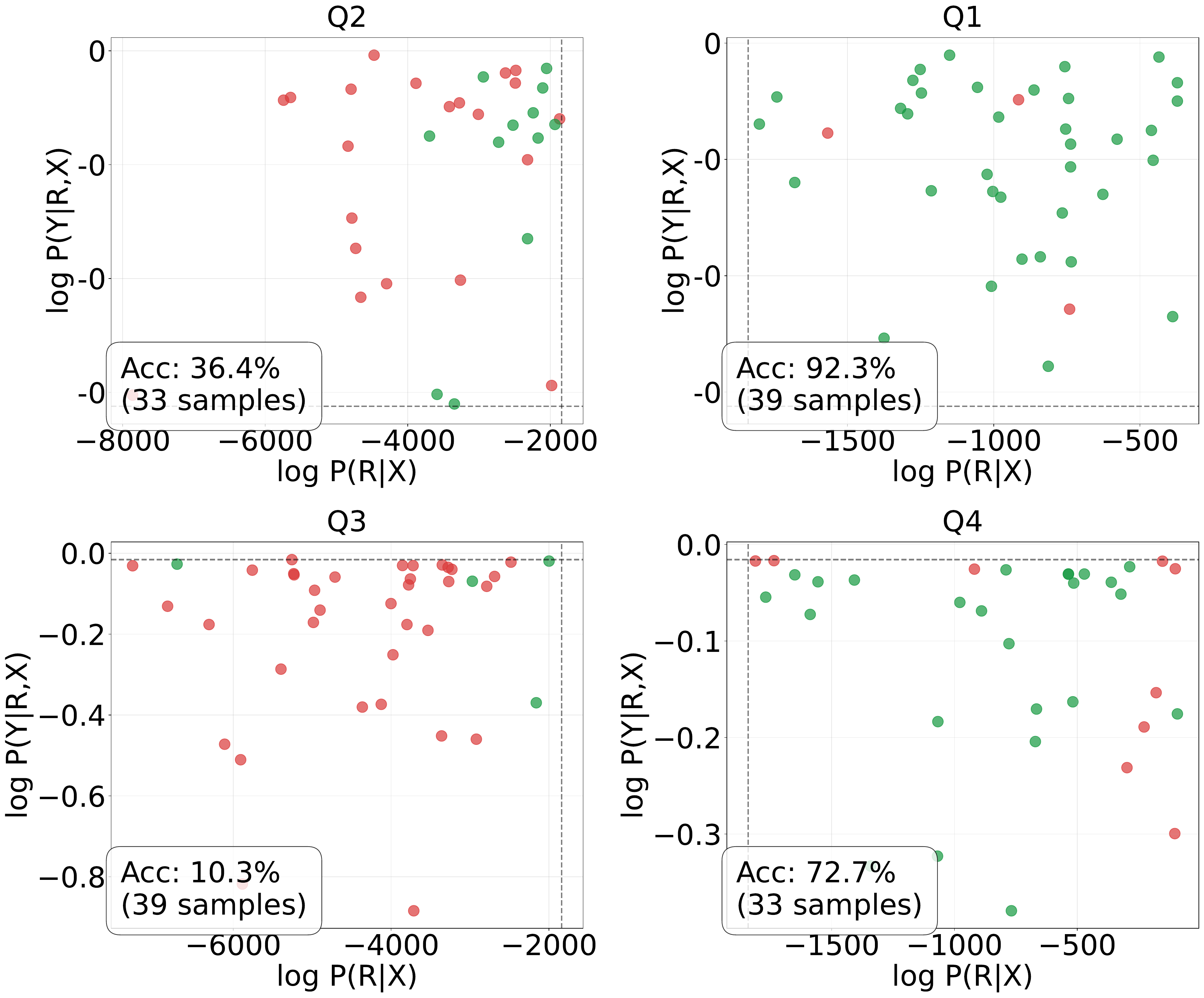}
    \caption{Quadrant-split representation with Q1–Q4 defined by median thresholds of reasoning and answer confidence. }
    \label{fig:quadrats_plane-DS-qwen-aime2024}
  \end{subfigure}}
  \caption{Information Plane visualisations of DS-Distilled-Llama-8B on the AIME2024 dataset ($k=6$). 
  Green indicates correct answers, Red incorrect ones.}
  \label{fig:quadrats_subplot-DS-qwen-aime2024}
\end{figure*}

\begin{figure*}[htbp]
  \centering
  \includegraphics[width=0.6\textwidth]{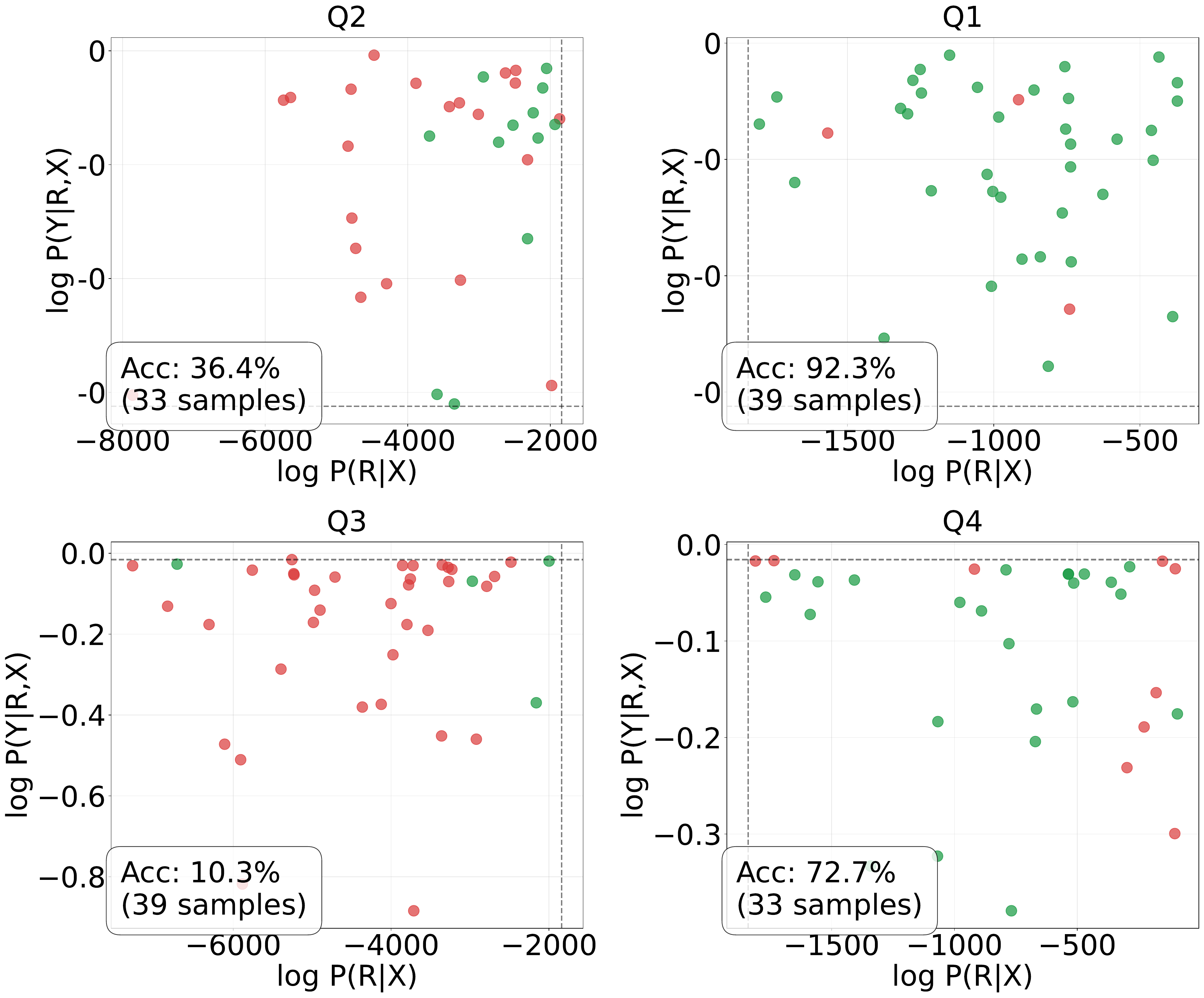}
  \caption{Information Plane of Deepseek-Distill-Qwen-2.5-7b on the AIME2024 dataset ($k=6$). Dark red indicates correct answers; light red indicates incorrect ones. Quadrants represent: Q1 (high answer confidence, low reasoning confidence), Q2 (high both), Q3 (low both), and Q4 (high reasoning confidence, low answer confidence).}
  \label{fig:quadrats_qwen2.5_aime2024}
\end{figure*}

\clearpage

\subsection*{AIME2024 with various Quadrants}

\clearpage

\section{Example of PiCSAR Scoring}
\label{sec:example_picsar_scoring}

\begin{figure*}[h]
    \centering
    \begin{tcolorbox}[colback=gray!10,colframe=gray!50,boxrule=0.5pt]
    {
    \small
    \ttfamily
    \textbf{Question:} Every day, Wendi feeds each of her chickens three cups of mixed feed. She gives the flock three meals per day. In the morning, she gives 15 cups of feed, and in the afternoon she gives another 25 cups. How many cups of feed must she give in the final meal of the day if her flock contains 20 chickens?

    \medskip
    \textbf{High PiCSAR Score}\\
    Step 1: Determine the total number of cups of feed given in the morning. Wendi gives 15 cups.\\
    Step 2: Determine the total number of cups of feed given in the afternoon. Wendi gives 25 cups.\\
    Step 3: Total feed given so far = 15 cups + 25 cups = 40 cups.\\
    Step 4: Each chicken needs 3 cups of feed per day.\\
    Step 5: Total daily feed required = 3 cups/chicken * 20 chickens = 60 cups.\\
    Step 6: Total given in morning and afternoon = 40 cups.\\
    Step 7: Final meal feed = 60 cups - 40 cups = 20 cups.\\
    The final answer is: 20.\\
    PiCSAR Score: -36.76

    \medskip
    \textbf{Low PiCSAR Score}\\
    Step 1: Determine the total morning feed. Wendi gives 15 cups.\\
    Step 2: Determine the total afternoon feed. Wendi gives 25 cups.\\
    Step 3: Total feed in morning and afternoon = 15 + 25 = 40 cups.\\
    Step 4: Since there are three meals, multiply the morning+afternoon total by 1.5 to estimate total daily feed. Total = 40 * 1.5 = 60 cups.\\
    Step 5: Each chicken is assumed to get 3 cups per meal, so 3 meals * 3 cups = 9 cups per chicken per day.\\
    Step 6: Total feed needed per day = 20 chickens * 9 cups = 180 cups.\\
    Step 7: Final meal feed = 180 cups - 40 cups = 140 cups.\\
    The final answer is: 140.\\
    PiCSAR Score: -112.79

    \medskip
    \rmfamily
    A concise generation with high token probability yields a higher PiCSAR score than a longer, lower-probability generation.
    }
    \end{tcolorbox}
    \caption{Example of PiCSAR scoring on a grade-school maths problem. The high-scoring generation is both correct and concise, while the low-scoring generation contains faulty reasoning and receives a substantially lower score.}
    \label{fig:picsar_example}
\end{figure*}

\newpage

\section{Further Case Study on of PiCSAR Scoring}
\label{sec:further_example_picsar_scoring}
\subsection{Case 1 \& 2}
\begin{figure*}[h]
\centering
\begin{tcolorbox}[colback=gray!10,colframe=gray!50,boxrule=0.5pt]
{\small\ttfamily

\textbf{Case 1: Correct generation is shorter yet yields a higher log-probability}

\medskip
\textbf{Question:} There is one set of twins and one set of triplets. One twin is 7 years older than one triplet. If their combined ages are 44, how old is one of the twins?

\medskip
\textbf{Correct Generation (PiCSAR Score: -46.12)}\\
Step 1: Let's denote the age of one of the twins as T and the age of one of the triplets as Tr. Since one twin is 7 years older than one triplet, we can express this relationship as T = Tr + 7.\\
Step 2: The combined ages of the twins and the triplets are 44. We can express this as 2T + 3Tr = 44.\\
Step 3: Substitute T = Tr + 7 into the equation: 2(Tr + 7) + 3Tr = 44.\\
Step 4: Expand: 2Tr + 14 + 3Tr = 44.\\
Step 5: Combine: 5Tr + 14 = 44.\\
Step 6: 5Tr = 30. \quad Step 7: Tr = 6. \quad Step 8: T = 6 + 7 = 13.\\
The final answer is: 13

\medskip
\textbf{Incorrect Generation (PiCSAR Score: -694.92)}\\
Step 1: Let's denote the age of one triplet as 'x'. Combined age of triplets: 3x.\\
Step 2: Let's denote the age of one twin as 'y'. Combined age of twins: y + (y + 7) = 2y + 7.\\
Step 3: Equation: 3x + 2y + 7 = 44. \quad Step 4: Simplify: 3x + 2y = 37.\\
Step 5: We can express 'y' in terms of 'x' from the equation y + (y + 7) = 2y + 7, which is y = (2y + 7) - 7. So, y = 2y. ...[TRUNCATED]... However, this is incorrect. Let's re-evaluate the equation y + (y + 7) = 2y + 7. We can represent it as y + 7 = y + 7. However, this is incorrect. Let's re-evaluate...

\vspace{0.8em}\hrule\vspace{0.8em}

\textbf{Case 2: Correct generation is longer yet yields a higher log-probability}

\medskip
\textbf{Question:} Milly needs to return a book she decided was really boring. The book weighs 4 pounds, cost \$32, and needs to be returned to a distribution center 20 miles away. If the shipping company charges \$0.35 per pound plus \$0.08 per mile, and Amazon will only refund 75\% of the book's purchase price, how much money will Milly lose?

\medskip
\textbf{Correct Generation (PiCSAR Score: higher by 105.90)}\\
Step 1: Shipping cost by weight = 4 $\times$ \$0.35 = \$1.40.\\
Step 2: Shipping cost by distance = 20 $\times$ \$0.08 = \$1.60.\\
Step 3: Total shipping = \$1.40 + \$1.60 = \$3.00.\\
Step 4: Refund = 75\% $\times$ \$32 = \$24.\\
Step 5: Loss from refund = \$32 - \$24 = \$8.\\
Step 6: Total loss = \$3.00 + \$8 = \$11.\\
The final answer is: \$11

\medskip
\textbf{Incorrect Generation}\\
Step 1--2: Shipping = 4 $\times$ \$0.35 + 20 $\times$ \$0.08 ...[TRUNCATED]... The total loss is actually \$2.99 - \$24 = -\$21.01. Since we cannot lose a negative amount, we must recalculate the total loss.\\
Step 6: Since the refund is more than the shipping cost, the total loss will be 0.\\
The final answer is: \$2.99

\medskip
\rmfamily\itshape
In Case 1, the correct generation is shorter and scores 648.80 points higher. In Case 2, the correct generation is 67 tokens longer yet scores 105.90 points higher, demonstrating that PiCSAR rewards reasoning quality rather than brevity.
}
\end{tcolorbox}
\caption{PiCSAR scoring examples illustrating that the score reflects reasoning quality independent of generation length.}
\label{fig:picsar_cases_1_2}
\end{figure*}
\clearpage
\subsection{Case 3}
\begin{figure*}[h]
\centering
\begin{tcolorbox}[colback=gray!10,colframe=gray!50,boxrule=0.5pt]
{\small\ttfamily

\textbf{Case 3: Impact of Answer Confidence}

\medskip
\textbf{Question:} In the first week, Judy read for 15 minutes each night before going to sleep. In the second week, she read a total of 100 pages. If she can read 2 pages per 1.5 minutes, how many pages did she read in two weeks?

\medskip
\textbf{Before Answer Confidence (Incorrect Generation)}\\
Step 1: Total reading time in the first week: 15 $\times$ 7 = 105 minutes.\\
Step 2: Pages read in the first week: 105 / 1.5 = 70 pages.\\
Step 3: Total pages in two weeks: 70 + 100 = 170 pages.\\
The final answer is: 170

\medskip
\textbf{After Answer Confidence (Correct Generation)}\\
Step 1: Total reading time in the first week: 15 $\times$ 7 = 105 minutes.\\
Step 2: Pages read in the first week: 105 / 1.5 $\times$ 2 = 140 pages.\\
Step 3: Total pages in two weeks: 140 + 100 = 240 pages.\\
The final answer is: 240

\medskip
\textbf{Correct Answer:} 240

\medskip
\rmfamily\itshape
Both generations share similar reasoning structure and length. The incorrect generation omits the multiplication by 2 in Step 2, yielding a lower answer confidence that allows PiCSAR to discriminate between the two.
}
\end{tcolorbox}
\caption{Example illustrating the role of answer confidence in PiCSAR scoring.}
\label{fig:picsar_case_3}
\end{figure*}

\twocolumn

\begin{figure}[tp]
  \centering

  \begin{subfigure}[b]{0.4\textwidth}
    \centering
    \includegraphics[width=\textwidth]{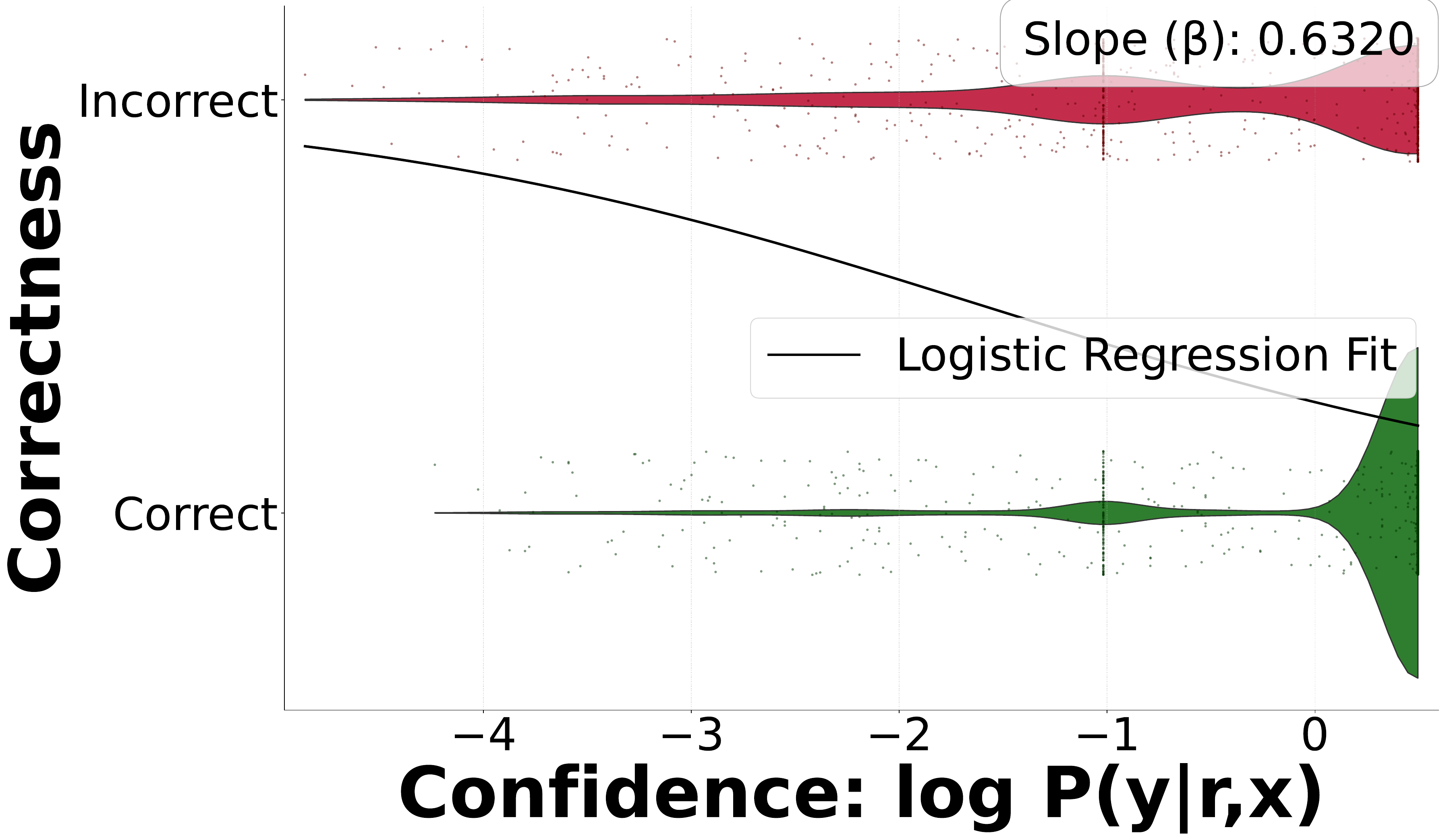}
    \caption{Qwen3 8B}
    \label{fig:conf_cal_qwen8b_logi}
  \end{subfigure}
  \hfill
  \begin{subfigure}[b]{0.4\textwidth}
    \centering
    \includegraphics[width=\textwidth]{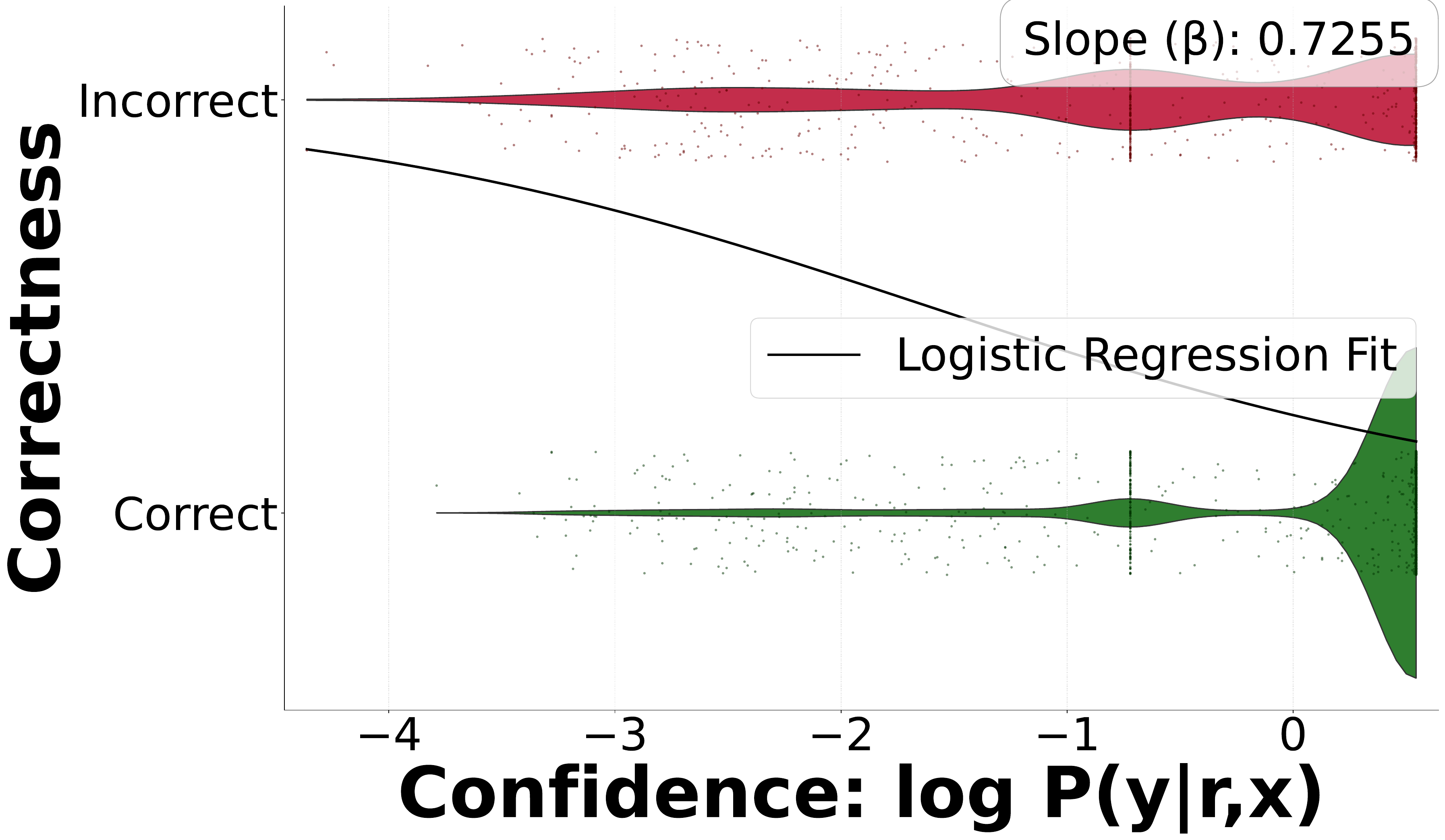}
    \caption{Qwen3 14B}
    \label{fig:conf_cal_qwen14b_logi}
  \end{subfigure}
  \hfill
  \begin{subfigure}[b]{0.4\textwidth}
    \centering
    \includegraphics[width=\textwidth]{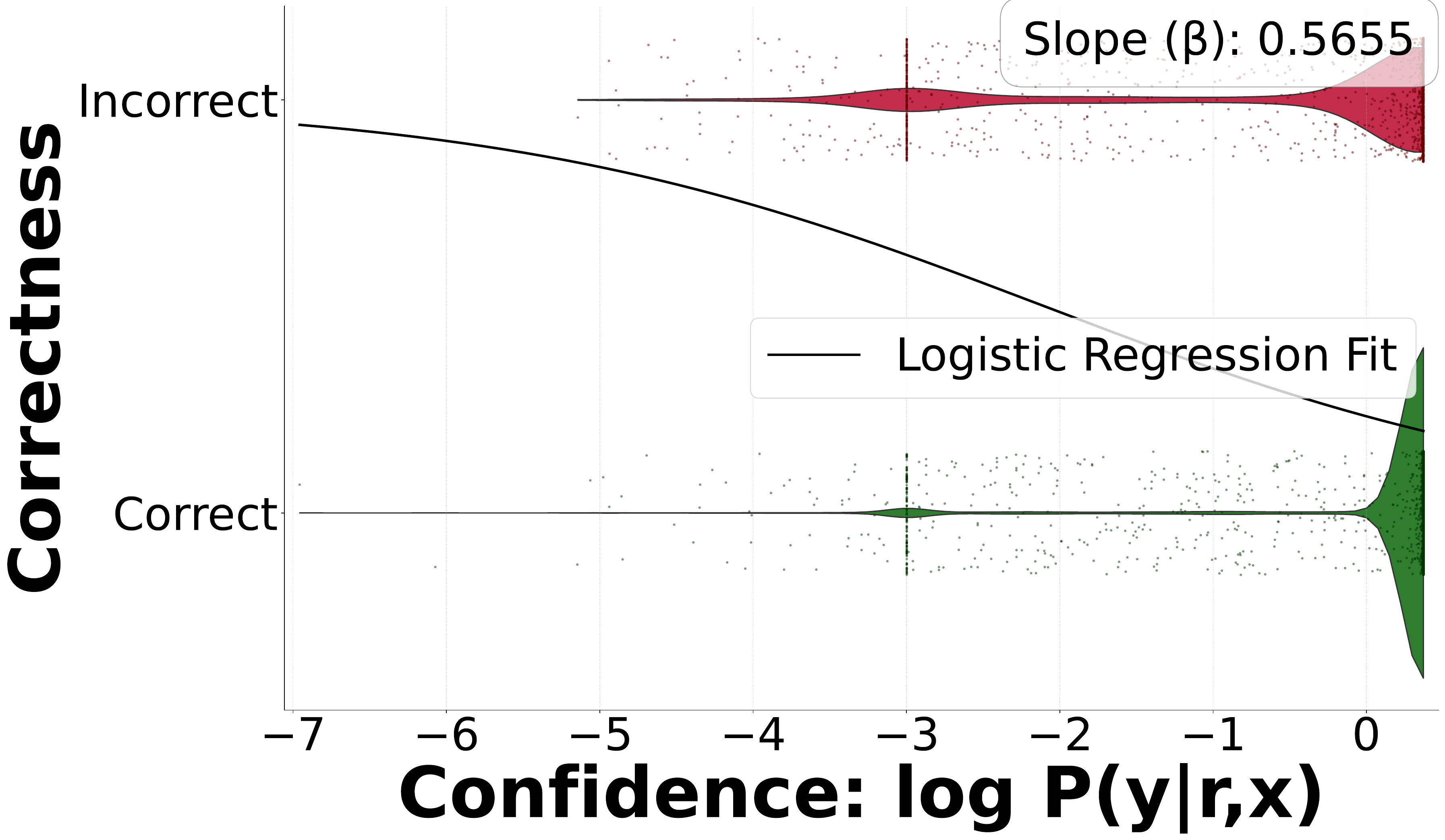}
    \caption{Qwen3 32B}
    \label{fig:conf_cal_qwen32b_logi}
  \end{subfigure}

  \caption{A detailed visualisation on the correct/incorrect densities based on logistic regression plot.
  } 
  \label{fig:conf_cal_all_qwen_logi}
\end{figure}

\section{Intra-model Reliability}

To support the intra-model results in \cref{intra-inter-confidence}, we analyse the calibration of \methodabbrev's confidence signal using the evaluation traces collected for the Qwen3 family.
For every sample we pair the answer log-probability $\log p(y \mid r, x)$ with its correctness label and fit a separate model per backbone.
The resulting calibration curves in \Cref{fig:qwen-calibration-wrap} exhibit a consistent monotonic trend: the logistic slopes are $0.63$, $0.73$, and $0.57$ for Qwen3-8B, 14B, and 32B respectively, and the corresponding point-biserial coefficients ($r \approx 0.31$, $0.35$, $0.29$) show a positive correlation between higher confidence and the probability of a correct answer.
\cref{fig:conf_cal_all_qwen_logi} also shows how this effect manifests in the raw score distribution.
Correct solutions concentrate around higher confidence values (closer to zero log-probability), whereas incorrect ones remain several nats lower, leaving limited overlap in the high-confidence region.
%

\label{sec:detailed-intra-model}
\subsection{Logistic Regression Experimental Training}



We model the relationship between confidence and correctness using logistic regression, similar to \citet{gema2024decore}. The binary outcome variable encodes whether the final answer is correct ($y \in {0,1}$), while the predictor is the model’s confidence score expressed as the log-probability of the final answer:
\begin{equation*}
    Pr(y=1 \mid \textit{Conf}) = \sigma(\alpha + \beta \cdot \textit{Conf})
\end{equation*}
where $\sigma$ is the sigmoid function. The regression coefficient $\beta$ quantifies the change in log-odds of correctness per unit change in confidence. A positive $\beta$ indicates that higher confidence increases the likelihood of correctness. For instance, as shown in Figure~\ref{fig:conf_cal_qwen14b_logi}, in Qwen3-14B, $\beta=0.7255$ corresponds to more than doubling the odds of correctness ($e^{0.7255} \approx 2.07$).

\begin{figure}[tp]
  \centering
  \includegraphics[width=0.43\textwidth]{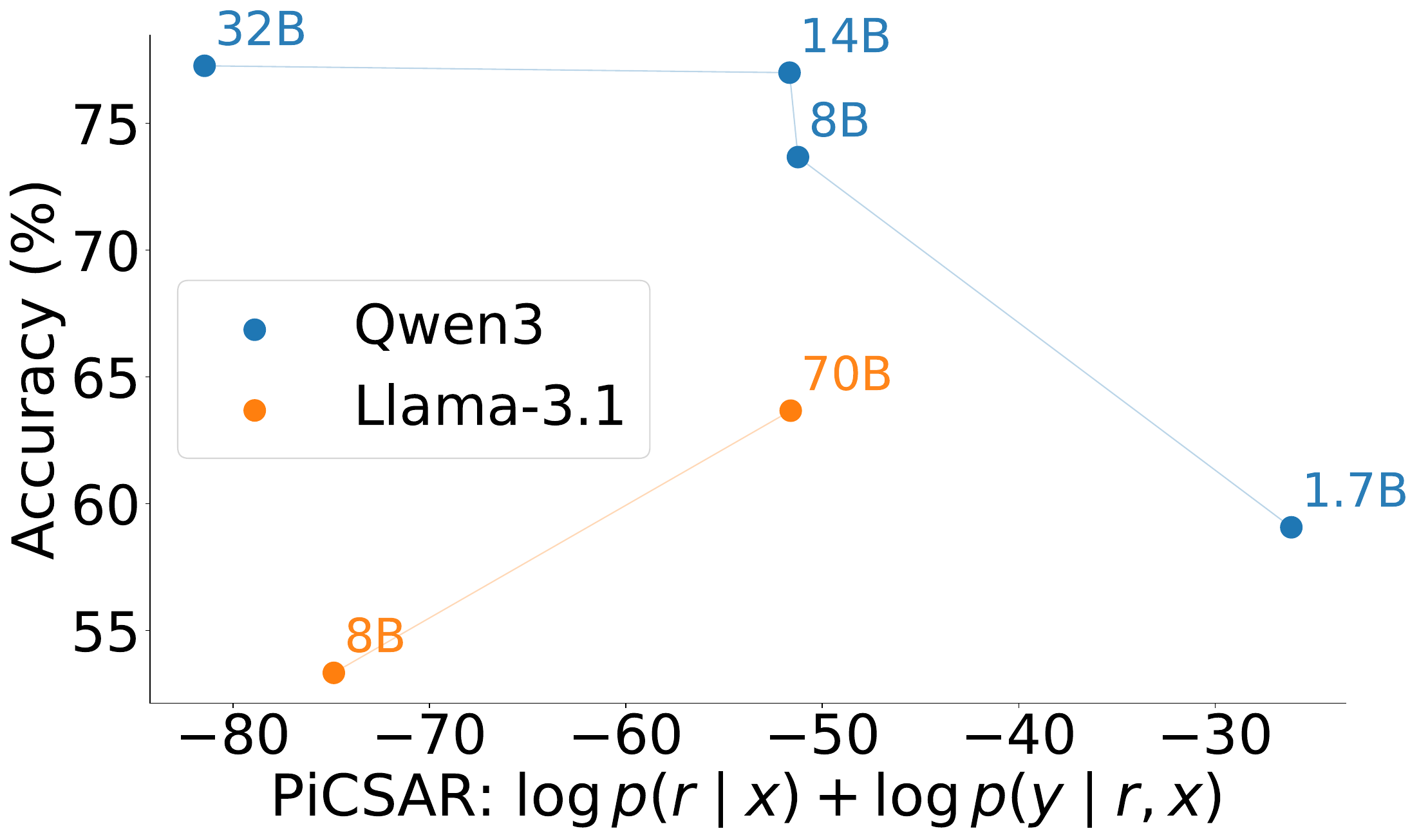}
  \caption{Comparison of $\%$ and \methodabbrev score.}
  \label{fig:models_comparison}
\end{figure}

\subsection{Point-biserial Correlation Coefficient}  
As a complementary measure to logistic regression, we compute the point-biserial correlation coefficient between confidence scores (continuous) and correctness (binary). This statistic, mathematically equivalent to Pearson’s correlation with a dichotomous variable, directly quantifies the strength of association between the two. It is defined as  
\begin{equation*}
r = \frac{\bar{x}_1 - \bar{x}_0}{s_x} \sqrt{\frac{n_1 n_0}{n^2}},
\end{equation*}
where $\bar{x}_1$ and $\bar{x}_0$ denote the mean confidence scores for correct and incorrect samples, $s_x$ is the pooled standard deviation, and $n_1, n_0$ are the respective sample counts. The coefficient is bounded in $[-1,1]$, with positive values indicating alignment between confidence and correctness. For instance, an $r$ of 0.35 for Qwen3-14B indicates a moderate positive association. Together with logistic regression, this provides a scale-free validation that confidence is a consistent predictor of correctness within a given model.  

\section{Inter-model Variance}
\label{sec:intermodel}
\emph{Inter-model variance analysis} challenges the assumption that confidence scores represent universal correctness measures across different models. While intra-model reliability remains stable across different model sizes and architectures, confidence scores cannot be compared across models of different parameter sizes and architectures. As shown in
\cref{fig:models_comparison}, 
the Llama family exhibits predictable trend: both accuracy and confidence increase with model size. 
In contrast, the Qwen family shows a non-monotonic relationship; Qwen3-1.7B achieves the highest confidence while showing the lowest accuracy.
\emph{This difference implies that while there is a general trend that confidence is a useful proxy for selecting an accurate reasoning path from a set of candidates within models, but
its actual value is model-specific and incomparable across different models.}

\end{document}